\documentclass[10pt,twocolumn,letterpaper]{article}

\usepackage[pagenumbers]{cvpr} % To force page numbers, e.g. for an arXiv version

\definecolor{cvprblue}{rgb}{0.21,0.49,0.74}
\usepackage[pagebackref,breaklinks,colorlinks,allcolors=cvprblue]{hyperref}

\usepackage{multirow}
\usepackage{algorithm}
\usepackage{algorithmic}
\usepackage{listings}
\usepackage[table]{xcolor}
\usepackage[most]{tcolorbox}

\def\paperID{} % *** Enter the Paper ID here 37144
\def\confName{CVPR}
\def\confYear{2026}

\title{LAMMI-Pathology: A Tool-Centric Bottom-Up LVLM-Agent Framework for Molecularly Informed Medical Intelligence in Pathology}

\author{Haoyang Su\\
\textit{Fudan University} \\
\textit{Shanghai Innovation Institute} \\
\textit{Shanghai Artificial Intelligence Laboratory}
\and
Shaoting Zhang\\
\textit{Shanghai Jiao Tong University} \\
\textit{Shanghai Artificial Intelligence Laboratory}
\and
Xiaosong Wang\\
\textit{Shanghai Innovation Institute} \\
\textit{Shanghai Artificial Intelligence Laboratory}
}
\begin{document}
\maketitle
\begin{abstract}
The emergence of tool-calling-based agent systems introduces a more evidence-driven paradigm for pathology image analysis in contrast to the coarse-grained text-image diagnostic approaches. With the recent large-scale experimental adoption of spatial transcriptomics technologies, molecularly validated pathological diagnosis is becoming increasingly open and accessible.
In this work, we propose LAMMI-Pathology (LVLM-Agent System for Molecularly Informed Medical Intelligence in Pathology), a scalable agent framework for domain-specific agent tool-calling. LAMMI-Pathology adopts a tool-centric, bottom-up architecture in which customized domain-adaptive tools serve as the foundation. These tools are clustered by domain style to form component agents, which are then coordinated through a top-level planner hierarchically, avoiding excessively long context lengths that could induce task drift.
Based on that, we introduce a novel trajectory construction mechanism based on Atomic Execution Nodes (AENs), which serve as reliable and composable units for building semi-simulated reasoning trajectories that capture credible agent-tool interactions. Building on this foundation, we develop a trajectory-aware fine-tuning strategy that aligns the planner’s decision-making process with these multi-step reasoning trajectories, thereby enhancing inference robustness in pathology understanding and its adaptive use of the customized toolset. Our
code is available at \href{https://github.com/Hoyant-Su/LAMMI}{https://github.com/Hoyant-Su/LAMMI}.
\end{abstract}
\section{Introduction}
\label{sec:intro}
Pathology image understanding is pivotal for advancing diagnostic precision and streamlining clinical workflows in computational pathology. The emergence of Large Vision-Language Models (LVLMs) has catalyzed a paradigm shift, with pioneering work~\cite{sun2025pathgenm, Sun_Zhu_Zheng_Zhang_Sun_Shui_Zhang_Li_Yang_2024PathAsst,Lu2024PathChat,Seyfioglu2024QuiltLLaVA} demonstrating that generative, text-based foundation models can render pathological image analysis more intuitive. However, the prevailing research landscape exhibits two critical limitations. 

First, existing studies~\cite{zhang2020evaluating,sun2025pathgenm,Seyfioglu2024QuiltLLaVA,Che2024WsiCaption,Lu2023Visual,Lu2024VisualLanguageCP} overemphasize textual representations, which fundamentally misaligns with the visual-centric nature of pathology, where diagnostic reasoning primarily relies on the direct morphological interpretation of histopathological images rather than textual descriptions. This textual bias creates an excessive dependency on expert-authored morphological reports, diverting attention from the primary visual evidence that pathologists inherently utilize. 
Second, although methods such as Chain-of-Thought~\cite{Wei2022CoT} (CoT) reasoning and Retrieval-Augmented Generation~\cite{Lewis2020RAG} (RAG) attempt to capture intermediate reasoning steps, they still fall short of fully supporting evidence-grounded medical decision-making in pathology~\cite{JEON2025comparative}, as these approaches often rely on textual intermediaries but overlook the molecular evidence-gathering modalities that are widely employed in clinical practice, such as immunohistochemistry (IHC) and RNA sequencing (RNA-seq), which provide quantitative, objective biological signals that can serve as direct evidence for pathological interpretation. Although recent works~\cite{chen2025evidencebaseddiagnosticreasoningmultiagent,sun2025pathgenm,Wang2025MedAgentProTM} have attempted to address these limitations by employing multi-agent strategies, these methods remain limited in realizing fully evidence-grounded reasoning in medical domains, as their reasoning processes largely operate on internally simulated outputs rather than direct observations, and their increased framework complexity may exacerbate hallucination issues common in generative models~\cite{Chen2025EvidencebasedDR}.

As an exploration of pathological vision-driven molecular mechanisms and evidence-based reasoning, gene expression profiles from techniques like spatial transcriptomics provide a rich source of quantitative data that captures the tissue's intrinsic state~\cite{Wang2024TMONet,Lazar2025Spatial,li2025multimodalfoundationmodelscomputational}. While existing datasets~\cite{jaume2024hest,Chen2024STimage} offer extensive molecular annotations paired with histopathology images, applications have largely been limited to image-level labeling or predictive tasks focused on specific organs or diseases~\cite{shi2024highresolutionspatialtranscriptomicshistology}. Consequently, integrating these molecular signals with generative vision-language models presents a critical opportunity to enable reliable, evidence-based reasoning, bridging the gap between morphological presentation and textual interpretation.

\begin{figure*}[htbp]
    \centering
    \includegraphics[width=0.9\linewidth]{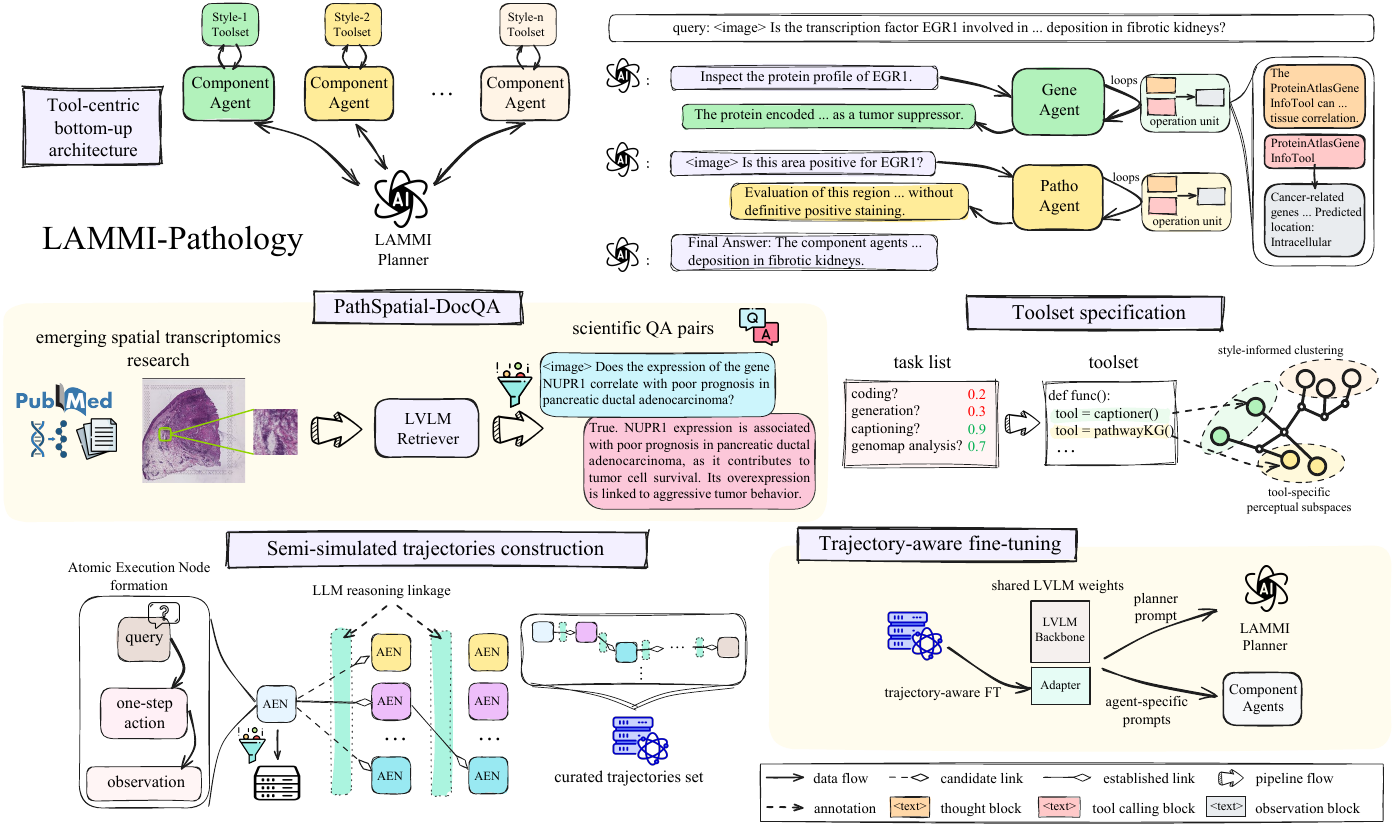}
    \caption{Overview of the LAMMI-Pathology framework. The top row illustrates the hierarchical reasoning architecture where component agents orchestrate style-specific tools and aggregate contextual evidence for the LAMMI planner. The middle row depicts the data and tool construction pipeline, where spatial transcriptomics literature paired with histopathology images are processed through LVLM retriever to extract QA pairs, and tools are bottom-up clustered according to their sequential co-occurrence frequency during AEN construction, with adaptive cluster numbers. The bottom row presents the AEN-driven semi-simulated trajectory generation and trajectory-aware fine-tuning methodology, with both the planner and component agents operating on shared fine-tuned model parameters.}
    \label{fig:LAMMI_arch}
\end{figure*}

Building on this opportunity, we introduce LAMMI-Pathology, as detailed in \cref{fig:LAMMI_arch}, a domain-specific agent framework designed to effectively learn from and leverage customized tools. Unlike traditional multi-agent systems, LAMMI-Pathology employs a bottom-up architecture in which customized, domain-adaptive tools are first specifically designed and then grouped into clusters, each managed by a component agent, which in turn is coordinated hierarchically under a visual-driven LVLM planner.

 To support robust reasoning, we develop a scalable strategy to generate semi-simulated agent trajectories, in which the returns of real tools are incorporated and enriched with LLM reasoning. Correspondingly, we design a trajectory-aware fine-tuning strategy to align the planner with these trajectories, ensuring effective adaptation and utilization of downstream tools within LAMMI-Pathology.

  Our contributions in the proposed framework are threefold. First, we develop a scalable planner-driven agent framework that supports hierarchical tool-calling, where both tools and component agents can be locally integrated rather than constrained by external APIs. We demonstrate that a single set of fine-tuned weights can adapt to both the planner and component agents, which significantly reduces the memory allocation required for constructing extensible agent systems. Second, we design our framework with pathology-specific toolsets and domain knowledge, addressing the limitations of conventional VLM in pathology, which often suffers from coarse-grained image-to-text mappings and the absence of evidence-grounded traces. Third, we introduce Atomic Execution Nodes (AENs) for tool-execution-grounded trajectory generation and a trajectory-aware fine-tuning strategy that leverages these nodes, enabling the planner to coordinate multiple component agents while improving their tool-calling decisions.
      
\section{Related Work}
\label{sec:relatedwork}

\subsection{Tool-Calling Agent Systems and Trajectory Learning}
Tool-calling based agent system design has become prevalent in both general-purpose and domain-specific applications~\cite{Xu2025LLMAgentSurvey,Li2025Prominent}, driven by the capabilities of large language models (LLMs) to perform complex tasks through tool utilization~\cite{masterman2024landscapeemergingaiagent, Georg2025LLM, liu2025toolace}. Early studies such as TALM~\cite{Parisi2022TALMTA} and Toolformer~\cite{schick2023toolformerlanguagemodelsteach}  first demonstrated that LLMs could self-learn tool invocation patterns, inspiring structured reasoning frameworks. ReACT~\cite{Shunyu2023REACT} proposed a standardized trajectory comprising Thought, Action, and Observation, providing a foundational framework for subsequent developments but exhibiting a simple, linear structure that limits its flexibility. 
Following works such as ToolACE~\cite{liu2025toolace}, which synthesizes pipeline heavily relies on the general capabilities of pretrained LLMs, confining it to the text modality. The complex exploration challenge inherent in such knowledge distillation is not adequately addressed by its sequential three-stage logic of speciation, adaptation, and evolution.
Yet, text-only tool-calling paradigms are not suitable for image-oriented tasks since visual tokens are absent from the reasoning process. To address these shortcomings, several multimodal agent systems have been proposed,
yet they still exhibit distinct limitations. Works such as MAT-Agent~\cite{gao2025multi} have explored trajectory simulation, but their focus remains constrained to specific benchmarks, such as GAIA~\cite{mialon2023gaiabenchmarkgeneralai} and GTA~\cite{wang2024gta}, thereby limiting their adaptability in novel or emerging domains. 
Other systems, like MLLM-Tool~\cite{wang2024mllmtool}, are limited by controllers that validate tool invocations only from a formal perspective rather than through outcome-based correctness.

\subsection{LVLM-Driven Agent Systems in Pathology}
Extending these paradigms to pathology, several works have investigated LVLM-driven agent systems tailored for medical image understanding. For instance, SlideSeek~\cite{chen2025evidencebaseddiagnosticreasoningmultiagent} introduced a multi-agent system for evaluating gigapixel whole-slide images, but it encodes a fixed supervisor-explorer-report workflow and is tightly coupled to the PathChat+, which limits composability with alternative toolsets and evidence modules. PathGen-1.6M~\cite{sun2025pathgenm} leveraged multi-agent collaboration to generate a large-scale dataset, yet its iterative revision process remains text-centric and not fully evidence-based. Moreover, PathFinder~\cite{Ghezloo2025PathFinder} proposes a multi-modal, multi-agent system for histopathology that emulates expert workflows across gigapixel WSIs, but it relies on fixed agent roles and lacks flexible tool integration and trajectory-aware reasoning. These works collectively highlight the need for a flexible agent framework capable of dynamic tool integration and generalizable, evidence-grounded reasoning through trajectories, motivating the design of LAMMI-Pathology.

\section{Method}
\label{sec:method}

\subsection{Trajectory Generation via Tool Execution}
\label{subsec:AEN}
Conventional agent tool-calling approaches~\cite{Tang2023ToolAlpacaGT,Parisi2022TALMTA} either impose structured constraints through predefined schemas or rely on carefully designed prompts with large numbers of simulated tool-calling examples to teach agents how to use tools. However, these methods are not well suited for downstream applications that involve custom tools, and agents cannot easily learn the usage protocols of new tools in zero-shot scenarios. This limitation restricts the extension of tools for downstream tasks.

In this work, to facilitate the construction of a semi-simulated trajectory dataset for LVLM agents, we introduce a novel concept termed Atomic Execution Nodes (AENs). An AEN represents the minimal verifiable interaction primitive between an agent and an external tool, encapsulating a query, the corresponding tool input, and the observed output. Formally, each AEN is defined as a triplet $(Q, A, O) \in \mathcal{Q} \times \mathcal{T} \times \mathcal{O}$, where $Q$ is a query sampled from a systematically constructed set of pathology-genomics questions $\mathcal{Q}$, $A$ is the input instruction for a potentially relevant tool selected from the custom toolset $\mathcal{T}$, and $O \in \mathcal{O}$ is the resulting observation returned by the tool execution. Unlike purely synthetic or imagined reasoning traces, AENs directly encode grounded tool interactions, providing reliable atomic units for constructing higher-level trajectories.

These atomic nodes serve as building blocks for generating coherent meta-trajectories. To this end, a large language model (LLM) performs causal concatenation over candidate AENs, identifying semantically and temporally compatible successor nodes and inserting intermediate Thoughts that explicitly capture reasoning transitions. Formally, a meta-trajectory $\tau$ is constructed as a sequence of alternating reasoning and execution steps, as specified in the expression below:
\begin{equation}
\label{eq:meta_trajectory}
\tau = \{ (T_i, A_i, O_i) \}_{i=1}^{K},
\end{equation}
where $T_i$ denotes the $i$-th Thought segment capturing reasoning transitions, $A_i \in \mathcal{T}$ represents the $i$-th Action (tool selection), and $O_i \in \mathcal{O}$ is the corresponding Observation. Each inserted Thought $T_i$ ensures that the agent's reasoning chain remains causally consistent and that tool executions are interpreted and applied correctly.

Furthermore, the LLM filters out semantically invalid or inconsistent nodes, maintaining high-quality trajectories suitable for downstream fine-tuning. This semi-simulated design balances grounded authenticity through real tool outputs and scalable synthesis via LLM-driven concatenation, enabling the construction of large-scale, structured training datasets that capture both operational logic and reasoning processes for LVLM agents.

While AEN-based trajectory construction provides rich training data, traditional ReAct-based systems suffer from context length explosion when interacting with entire toolsets~\cite{li2025deepagentgeneralreasoningagent}, thus degrading reasoning quality. To address this limitation, we propose a Tool-centric Bottom-Up Clustering Architecture that fundamentally reimagines how agents interact with tool repertoires. Our architecture clusters tools based on their style and utility into domain-specific component agents. These component agents, each equipped with a curated subset of specialized tools, interact with their respective tool styles to generate outputs that collectively inform the central planner's resolution of user queries, significantly reducing context complexity while maintaining specialized problem-solving capabilities.

\subsection{Trajectory-aware Adapter Fine-tuning}
\subsubsection{Adapter Architecture}
\begin{figure*}
        \centering
        \includegraphics[width=\linewidth]{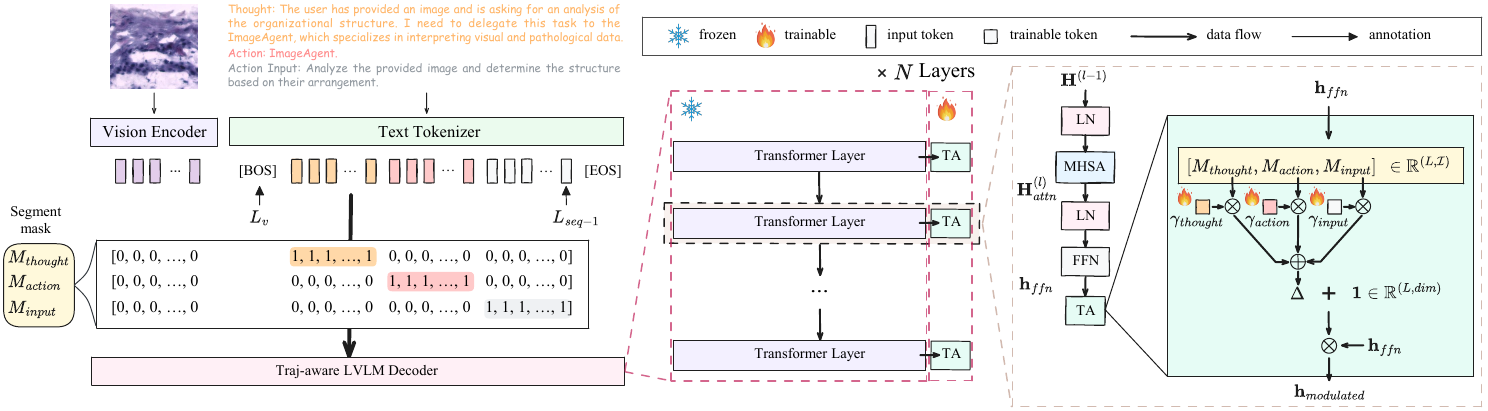}
        \caption{Trajectory-aware Adapter architecture. The adapter is injected after the FFN in each Transformer decoder layer. Segment masks are dynamically generated from input sequences to identify \texttt{Thought}, \texttt{Action}, and \texttt{Action Input} segments. Three learnable per-channel scaling vectors are applied, producing a modulation term that is multiplicatively applied to the FFN output.}
        \label{fig:traj_adapter_arch}
    \end{figure*}

Rather than fully fine-tuning the base VLM, we introduce a Trajectory-aware Adapter (TA) that operates as a structural alignment module in \cref{fig:traj_adapter_arch}. The adapter constrains gradient updates to low rank parameter subspaces associated with output trajectory formatting. This design aligns with observations that structural compliance such as producing syntactically valid tool calls tends to emerge earlier during fine-tuning and facilitates subsequent content level learning, with models typically acquiring format and procedural patterns before developing deeper semantic understanding~\cite{gong2025disentanglingfeaturestructuremathematically, Sinclair2022Structural}.

Following standard VLM architectures, visual and textual tokens are concatenated into a unified sequence $\mathbf{H}^{(0)} \in \mathbb{R}^{B \times (L_v + L_t) \times d}$ processed through $L$ Transformer layers, where $B$ is the batch size, $L_v$ and $L_t$ denote visual and textual token counts, respectively, $d$ is the hidden dimension, and $L$ is the number of layers.

Each Transformer layer $l \in \{1, \ldots, L\}$ transforms hidden states $\mathbf{H}^{(l-1)} \in \mathbb{R}^{B \times (L_v + L_t) \times d}$ through the standard pre-norm architecture. We inject the trajectory adapter $\text{TA}^{(l)}$ by dynamically patching the feed-forward network (FFN) block's forward method, which intercepts FFN computation to apply adapter modulation before the residual connection. The complete layer transformation is
\begin{equation}
\label{eq:layer_transformation}
\mathbf{H}^{(l)} = \mathbf{H}_{\text{attn}}^{(l)} + \text{TA}^{(l)}\left(\text{FFN}^{(l)}(\text{LayerNorm}(\mathbf{H}_{\text{attn}}^{(l)}))\right),
\end{equation}
where $\mathbf{H}_{\text{attn}}^{(l)} = \mathbf{H}^{(l-1)} + \text{MHSA}^{(l)}(\text{LayerNorm}(\mathbf{H}^{(l-1)}))$ denotes the output after attention and the first residual connection in Eq.~\eqref{eq:layer_transformation}. We denote full-layer hidden state tensors as $\mathbf{H}$ and intermediate adapter outputs as $\mathbf{h}$.

\subsubsection{Segment Mask-Guided Modulation Mechanism}
The trajectory adapter $\text{TA}^{(l)}$ receives FFN output $\mathbf{h}_{\text{ffn}}^{(l)} = \text{FFN}^{(l)}(\text{LayerNorm}(\mathbf{H}_{\text{attn}}^{(l)})) \in \mathbb{R}^{B \times (L_v + L_t) \times d}$ and applies segment mask-guided modulation. The adapter employs three learnable per-channel scaling vectors $\boldsymbol{\gamma}_{\text{thought}}$, $\boldsymbol{\gamma}_{\text{action}}$, $\boldsymbol{\gamma}_{\text{input}} \in \mathbb{R}^d$, each zero-initialized to preserve pre-trained behavior. The modulation mechanism adopts IA3's core principle~\cite{NEURIPS2022IA3} of per-channel scaling via element-wise multiplication, applied to FFN outputs rather than attention keys/values, and extended with segment-aware selection among multiple scaling vectors.

During training, we dynamically generate a segment mask $\mathbf{M} \in \{0,1\}^{B \times L_{\text{seq}} \times 3}$ where $L_{\text{seq}} = L_v + L_t$ denotes the total sequence length. Let $\mathcal{I} = \{0, 1, 2\}$ index segment types corresponding to \texttt{Thought}, \texttt{Action}, and \texttt{Action Input}, respectively. For visual tokens ($t \leq L_v$), $\mathbf{M}[b, t, i] = 0$ for all $i \in \mathcal{I}$. For text tokens ($t > L_v$), $\mathbf{M}[b, t, i] = \mathbf{1}[t \in \mathcal{S}_i]$ where $\mathcal{S}_0$, $\mathcal{S}_1$, $\mathcal{S}_2$ denote \texttt{Thought}, \texttt{Action}, and \texttt{Action Input} position sets respectively, and $\mathbf{1}[\cdot]$ denotes the indicator function. Mask generation performs on-the-fly substring matching against tokenized trajectory markers, eliminating the need for pre-annotated data.

The modulation term $\boldsymbol{\Delta} \in \mathbb{R}^{B \times L_{\text{seq}} \times d}$ is computed as
\begin{equation}
\label{eq:modulation_term}
\boldsymbol{\Delta}[b, t, :] = \sum_{i \in \mathcal{I}} \mathbf{M}_b[b, t, i] \odot \boldsymbol{\gamma}_i,
\end{equation}
where $\boldsymbol{\gamma}_0 = \boldsymbol{\gamma}_{\text{thought}}$, $\boldsymbol{\gamma}_1 = \boldsymbol{\gamma}_{\text{action}}$, and $\boldsymbol{\gamma}_2 = \boldsymbol{\gamma}_{\text{input}}$ denote the scaling vectors $\boldsymbol{\gamma}_i \in \mathbb{R}^d$ for the three segment types, $\mathbf{M}_b[b, t, i]$ denotes the broadcast of the scalar mask $\mathbf{M}[b, t, i] \in \{0,1\}$ to all $d$ dimensions when multiplied with $\boldsymbol{\gamma}_i$, and $\odot$ denotes element-wise multiplication. The modulation is applied multiplicatively to the FFN output,

\begin{equation}
\label{eq:modulated_output}
\mathbf{h}_{\text{modulated}}^{(l)} = \mathbf{h}_{\text{ffn}}^{(l)} \odot (\mathbf{1} + \boldsymbol{\Delta}),
\end{equation}
where $\mathbf{1}$ denotes a tensor of ones matching $\mathbf{h}_{\text{ffn}}^{(l)}$ in shape. This modulated output is added to $\mathbf{H}_{\text{attn}}^{(l)}$ via residual connection, producing $\mathbf{H}^{(l)}$ as specified in Eq.~\eqref{eq:layer_transformation}.

The segment-aware modulation mechanism enables learning distinct transformation patterns for different trajectory components. By applying segment-specific scaling vectors, the adapter captures tool-agnostic structural regularities including formatting requirements, enabling generalization to unseen tools while preserving the base model's general reasoning capabilities. Zero-initialized scaling vectors ensure minimal disruption to pre-trained representations, allowing gradual adaptation of structure-aware patterns without catastrophic forgetting of visual understanding and general language abilities.

\section{Experiments}
\label{sec:experiment}
\subsection{Dataset Curation and Experimental Settings}
Dataset construction, information extraction, and results evaluation were performed and verified using Qwen3-VL-235B-A22B-Instruct~\cite{qwen3vl_blog_2025} and Qwen3-235B-A22B~\cite{yang2025qwen3technicalreport}.

Following the procedure described in \cref{subsec:AEN}, we construct ST-Traj, a trajectory-level corpus derived from 10,684 successfully executed AENs in spatial transcriptomics. Through semantic filtering, length control, and limiting each node's usage to a maximum of three times, 6,818 high-quality meta-trajectories are obtained, each spanning 2-8 AENs and split into training, validation, and test subsets at an 85:5:10 ratio. This dataset serves as the foundation for analyzing multi-step tool-calling behaviors and reasoning dependencies.

To complement trajectory supervision with natural-language reasoning, we assemble PathSpatial-DocQA, a pathology-grounded question-answering dataset curated from HEST~\cite{jaume2024hest} and STimage-1K4M~\cite{Chen2024STimage}. The curation process is guided by 158 high-impact studies from Nature sub-journals, Cell, and other top-tier venues in spatial transcriptomics, emphasizing modal reasoning, pathology-driven visual interpretation, and the incorporation of molecular gene expression information.

In addition, we adopt PathMMU~\cite{sun2024pathmmumassivemultimodalexpertlevel}, one of the public QA datasets of clinical origin, as a gold-standard benchmark for pathological image understanding. Its test-tiny split is used for pure inference evaluation under all agent frameworks in comparison.

Tool incorporation policy includes molecular-related APIs~\cite{Kundra2021Oncotree,Putman2023Monarch,Wu2013BioGPS,Xin2016MyGene,Lelong2022BioThings,Yates2019Ensembl,Robert2016GDC,Mathias2015ProteinAtlas,KUNDU2023GeDiPNet}, with captioning models fine-tuned on Quilt-Instruct~\cite{Seyfioglu2024QuiltLLaVA}, PathGen~\cite{sun2025pathgenm}, and PathCap~\cite{Sun2024pathcap} datasets, and pathology-gene expression alignment tools~\cite{li2022blipbootstrappinglanguageimagepretraining} trained using ROI and RNA-Seq labels from HEST~\cite{jaume2024hest} and STimage-1K4M~\cite{Chen2024STimage}.

All experiments use a fixed random seed of 37. LVLM fine-tuning is implemented using DeepSpeed ZeRO-3~\cite{rajbhandari2020zeromemoryoptimizationstraining, Rasley2020DeepSpeed} with four H200 GPUs, while inference is carried out on four RTX 4090 GPUs. Hyperparameters are standardized across all frameworks, with maximum iterations, execution timeout, and generation length set to 8, 300 seconds, and 2048 tokens, respectively. For trajectory quality assessment, we employ Tool Consistency F1 (TCF1) and Trajectory Success Score (TSS) to evaluate ST-Traj quality against standard trajectories. Across all datasets, we measure the Tool Redundancy Rate (TRR), Answer Consistency Score (ACS), and Hallucination Rate (HR) to assess trajectory quality. For closed-ended questions, we additionally compute F1 scores.

For framework and model selection, we compare our framework against leading LVLM-based tool-calling systems, including MLLM-Tools~\cite{wang2024mllmtool}, MAT-Agent~\cite{gao2025multi}, and OpenAI-Agent-SDK, as well as the canonical ReAct~\cite{Shunyu2023REACT} framework. For the base LVLMs used to construct the planner and component agents, we adopt several state-of-the-art (SOTA) models, including Qwen3-VL-8B~\cite{qwen3vl_blog_2025}, InternVL3.5-8B~\cite{wang2025internvl35advancingopensourcemultimodal}, MiniCPM-V-4.5~\cite{yao2024minicpm}, and the proprietary GPT-5~\cite{openai2025gpt5systemcard}. For MLLM-Tools, which decouples autoregressive reasoning from visual understanding, we employ Vicuna-8B~\cite{vicuna2023}, Qwen3-8B~\cite{yang2025qwen3technicalreport}, and ToolACE-8B~\cite{liu2025toolace} as the base LLMs coupled with the ImageBind~\cite{Girdhar2023ImageBindOE} vision tower.

\subsection{Overall Performance}
\begin{table*}[htbp]
  \centering
  \caption{Performance comparison on PathSpatial-DocQA dataset. We report ACS, HR, and TRR across different frameworks and base models.}
    \begin{tabular}{cclccc}
    \toprule
    Framework & Adaptation & Planner & ACS $\uparrow$  & HR $\downarrow$ & TRR $\downarrow$ \\
    \midrule
    \rowcolor{blue!5}
        OpenAI-Agents-SDK & PE    & GPT-5 & 0.739  & \textbf{0.164}  & 0.033  \\
    \multirow{3}[1]{*}{ReACT} & \multirow{3}[1]{*}{PE} & InternVL3.5 8B & \underline{0.798}  & 0.301  & \textbf{0.000} \\
          &       & MiniCPM-V 4.5 & 0.715  & 0.353  & \underline{0.018} \\
          &       & Qwen3 VL 8B Instruct & 0.705  & 0.255  & 0.036  \\
          \rowcolor{blue!5}
    &  & InternVL3.5 8B & 0.462  & 0.662  & 0.049  \\
    \rowcolor{blue!5}
       MAT-Agent   &    LoRA+PE   & MiniCPM-V 4.5 & 0.373  & 0.660  & 0.238  \\
    \rowcolor{blue!5}
          &       & Qwen3 VL 8B Instruct & 0.404  & 0.713  & 0.096  \\
    \multirow{3}[0]{*}{MLLM-Tools} & \multirow{3}[0]{*}{LoRA+PE}   & ToolACE 8B + ImageBind & 0.448  & 0.540  & 0.374  \\
          &       & Vicuna v1.5 7B + ImageBind & 0.578  & 0.387  & 0.060  \\
          &       & Qwen3 8B + ImageBind & 0.498  & 0.231  & 0.603  \\
    \rowcolor{blue!5}
     &  & InternVL3.5 8B & \textbf{0.809}  & 0.338  & \textbf{0.000} \\
    \rowcolor{blue!5}
      LAMMI (Ours)    &   TA+PE    & MiniCPM-V 4.5 & 0.676  & \underline{0.205}  & 0.120  \\
    \rowcolor{blue!5}
          &       & Qwen3 VL 8B Instruct & 0.777  & 0.238  & 0.056  \\
    \bottomrule
    \end{tabular}%
  \label{tab:patho_spatial}%
\end{table*}%

When confronting frontier discovery questions incorporated in PathSpatial-DocQA, as \cref{tab:patho_spatial} shows, LAMMI with InternVL3.5 8B reaches 0.809 ACS, exceeding the SOTA OpenAI-Agents-SDK with GPT-5 by 0.070. Instruction-tuned models like Qwen3-VL-8B Instruct and MiniCPM-V-4.5 show higher adherence to PE under LAMMI, with slightly elevated TRR but reduced HR compared to direct parametric responses, as tool-calling introduces factual verification that mitigates hallucination. In contrast, LoRA-fine-tuned LVLMs in MAT-Agent and MLLM Tools exhibit superficial tool-calling that mimics rather than internalizes scientific reasoning, resulting in elevated TRR without ACS gains, suggesting that elevated tool invocation frequency fails to compensate for underdeveloped reasoning mechanisms.

  \begin{table*}[htbp]
     \centering
     \caption{Performance comparison on ST-Traj dataset. We report TRR, TCF1, TSS, ACS, and HR across different frameworks and base models.}
    \begin{tabular}{cclccccc}
    \toprule
    Framework & Adaptation & Planner & TRR $\downarrow$   & TCF1 $\uparrow$ & TSS $\uparrow$  & ACS $\uparrow$  & HR $\downarrow$\\
    \midrule
    \rowcolor{blue!5}
      OpenAI-Agents-SDK & PE    & GPT-5 & \underline{0.025} & 0.370 & 0.861 & 0.492 & \textbf{0.147} \\
    \multirow{3}[1]{*}{ReACT} & \multirow{3}[1]{*}{PE} & InternVL3.5 8B & \textbf{0.008} & 0.031 & 0.381 & 0.525 & \underline{0.446} \\
          &       & MiniCPM-V 4.5 & 0.048 & 0.362 & \underline{0.879} & \underline{0.615} & 0.579 \\
          &       & Qwen3 VL 8B Instruct & 0.092 & 0.377 & 0.792 & 0.587 & 0.506 \\
    \rowcolor{blue!5}
     &   & InternVL3.5 8B & 0.054 & 0.154 & 0.655 & 0.534 & 0.733 \\
     \rowcolor{blue!5}
      MAT-Agent    &   LoRA+PE    & MiniCPM-V 4.5 & 0.171 & 0.153 & 0.743 & 0.475 & 0.757 \\
    \rowcolor{blue!5}
    &       & Qwen3 VL 8B Instruct & 0.081 & 0.085 & 0.699 & 0.480  & 0.771 \\
    \multirow{3}[0]{*}{MLLM-Tools} & \multirow{3}[0]{*}{LoRA+PE}  & Vicuna v1.5 7B + ImageBind & 0.052 & 0.215 & 0.633 & 0.566 & 0.674 \\
          &       & ToolACE 8B + ImageBind &   0.355	&  \textbf{0.436}	& 0.423	& 0.482	& 0.686  \\
          &       & Qwen3 8B + ImageBind &   0.614    &   0.322    &  0.457     & 0.555      &  0.562\\
    \rowcolor{blue!5}
        &       & InternVL3.5 8B & 0.030  & 0.275 & 0.817 & \textbf{0.639} & 0.593 \\
    \rowcolor{blue!5}
       LAMMI (Ours)   &    TA+PE   & MiniCPM-V 4.5 & 0.143 & 0.397 & 0.868 & 0.533 & 0.562 \\
    \rowcolor{blue!5}
        &       & Qwen3 VL 8B Instruct & 0.036 & \underline{0.427} & \textbf{0.901} & 0.592 & {0.472} \\
    \bottomrule
      \end{tabular}%
    \label{tab:st_traj}%
  \end{table*}%
On ST-Traj, TA fine-tuning yields greater TSS gains than LoRA while achieving better TCF1 alignment with lower HR, as shown in \cref{tab:st_traj}. LAMMI employing InternVL3.5-8B as the planner with TA+PE reaches 0.275 TCF1, which is substantially higher than ReACT with PE alone at 0.031, while maintaining a comparatively low TRR. Notably, the TA+PE strategy enables open-source models to surpass GPT-5 in TCF1, TSS, and ACS, with the latter exhibiting notably lower ACS of 0.176, indicating limited generalization on structurally novel out-of-domain data like ST-Traj.

\begin{table*}[htbp]
    \centering
    \caption{Performance comparison on PathMMU dataset. We report ACS and F1 scores across four sub-tasks (Atlas, EduContent, PathCLS, PubMed) and trajectory quality metrics (HR, TRR) across different frameworks and base models.}
    \resizebox{\textwidth}{!}{
    \begin{tabular}{clccccccc}
    \toprule
    \multicolumn{1}{c}{\multirow{2}[2]{*}{Framework}} & \multicolumn{1}{l}{\multirow{2}[2]{*}{Planner}} & \multirow{2}[2]{*}{ACS $\uparrow$} & \multicolumn{4}{c}{F1 $\uparrow$}        & \multirow{2}[2]{*}{HR $\downarrow$} & \multirow{2}[2]{*}{TRR $\downarrow$} \\
          &       &       & Atlas & EduContent & PathCLS & Pubmed &  &  \\
    \midrule
    \rowcolor{blue!5}
    OpenAI-Agents-SDK & GPT-5 & \textbf{0.709}  & \textbf{0.755}  & \textbf{0.773}  & \textbf{0.480}  & \textbf{0.719}  & \underline{0.288}  & \textbf{0.000} \\
    \multirow{3}[1]{*}{ReACT} & InternVL3.5 8B & 0.500  & 0.341  & 0.369  & 0.198  & 0.443  & 0.691  & \textbf{0.000} \\
          & MiniCPM-V 4.5 & 0.554  & 0.438  & 0.494  & 0.232  & 0.512  & 0.538  & 0.092  \\
          & Qwen3 VL 8B Instruct & 0.572  & 0.433  & 0.541  & 0.288  & 0.518  & 0.483  & \underline{0.001} \\
    \rowcolor{blue!5}
      & InternVL3.5 8B & 0.506  & 0.349  & 0.437  & 0.257  & 0.440  & 0.726  & 0.041  \\
    \rowcolor{blue!5}
       MAT-Agent   & MiniCPM-V 4.5 & 0.403  & 0.231  & 0.261  & 0.141  & 0.315  & 0.742  & 0.173  \\
    \rowcolor{blue!5}
          & Qwen3 VL 8B Instruct & 0.469  & 0.322  & 0.376  & 0.226  & 0.395  & 0.694  & 0.128  \\
    \multirow{3}[0]{*}{MLLM-Tools} 
          & ToolACE 8B + ImageBind & 0.405  & 0.256  & 0.259  & 0.210  & 0.249  & 0.778  & 0.413  \\
          & Vicuna v1.5 7B + ImageBind & 0.426  & 0.298  & 0.292  & 0.186  & 0.275  & 0.727  & 0.045  \\
          & Qwen3 8B + ImageBind & 0.571  & 0.277  & 0.310  & 0.206  & 0.275  & 0.579  & 0.750  \\
    \rowcolor{blue!5}
     & InternVL3.5 8B & 0.528  & 0.382  & 0.390  & 0.181  & 0.512  & 0.671  & \textbf{0.000} \\
    \rowcolor{blue!5}
       LAMMI (Ours)   & MiniCPM-V 4.5 & 0.541  & 0.490  & 0.469  & 0.299  & 0.498  & \textbf{0.281}  & 0.139  \\
    \rowcolor{blue!5}
          & Qwen3 VL 8B Instruct & \underline{0.582}  & \underline{0.546}  & \underline{0.561}  & \underline{0.384}  & \underline{0.521}  & 0.455  & 0.017  \\
    \bottomrule
    \end{tabular}%
    }
    \label{tab:pathmmu}%
  \end{table*}%

As \cref{tab:pathmmu} shows, across the open-source benchmark PathMMU, although OpenAI-Agents-SDK with GPT-5 maintains superior comprehension with minimal tool reliance, LAMMI consistently achieves the highest performance among open-source frameworks when paired with the same base LVLM, with Qwen3-VL-8B-Instruct achieving the highest performance, reaching 0.582 ACS and 0.503 F1 on average, with an average F1 improvement of 0.058 and 0.173 on multiple choice questions compared to ReACT and MAT-Agent, respectively. The same trend is observed with InternVL3.5-8B and MiniCPM-V-4.5 incorporated into LAMMI.

% We observe that the zero-shot planner exhibits two types of inefficient behaviors when encountering previously unseen tools, namely unguided exploratory tool invocation and redundant repeated tool usage. Both behaviors persist even when tool descriptions are provided, indicating that mere description injection is insufficient to mitigate these issues.

% %MLLM-Tools 
% During the fine-tuning stage, MLLM-Tools demonstrates a rapid convergence toward the limited set of tool options, indicating that the model efficiently memorizes the small-scale action space. However, this convergence does not translate into a genuine ability to generalize tool usage beyond the training distribution. This limitation is reflected in the model’s failure to adapt when encountering novel or compositionally varied tool configurations.

\subsection{Ablation Studies}
Ablation studies are conducted across two dimensions. To assess generalization capability when incorporating new tools, adaptation methods are compared under varying Newly Incorporated Tools Ratio (NITR) on the PathSpatial-DocQA dataset to encourage free exploration trajectories. Newly incorporated tools form a fixed set, while existing tools are sampled per sample from a fixed seed according to the NITR ratio, ensuring varying tool combinations across samples for fairness. Additionally, the impact of maximum iteration limits on agent execution and GPU memory allocation between LAMMI and standard multi-agent systems (MAS) are analyzed on the ST-Traj dataset.

\subsubsection{Adaptation Method Comparison}

We compare four adaptation strategies on the PathSpatial-DocQA dataset under varying NITR. This dataset contains only QA tasks and encourages free exploration trajectories, making it suitable for evaluating adaptation methods under different tool incorporation scenarios. The baseline uses Prompt Engineering (PE) alone. We evaluate PE with TA, PE with LoRA, and PE with full fine-tuning to assess parameter efficiency and performance trade-offs.

\begin{table*}[htbp]
    \centering
    \caption{Ablation study on adaptation methods under varying NITR on PathSpatial-DocQA dataset. We compare PE, TA+PE, LoRA+PE, and Full+PE across different base models and report TRR and TSS for each NITR value.}
    \begin{tabular}{llcccccc}
    \toprule
    \multirow{2}[4]{*}{Adaption} & \multirow{2}[4]{*}{Planner} & \multicolumn{2}{c}{NITR=0.2} & \multicolumn{2}{c}{NITR=0.5} & \multicolumn{2}{c}{NITR=1.0} \\
\cmidrule{3-8}          &       & TRR $\downarrow$  & TSS $\uparrow$  & TRR  $\downarrow$   & TSS $\uparrow$  & TRR  $\downarrow$   & TSS $\uparrow$\\
    \midrule
    \rowcolor{blue!5}
     & InternVL3.5 8B & \textbf{0.000}  & 0.504  & \textbf{0.000}  & 0.500  & \textbf{0.000}  & 0.500  \\
    \rowcolor{blue!5}
      PE    & MiniCPM-V 4.5 & 0.027  & 0.814  & 0.050  & 0.780  & 0.085  & 0.801  \\
    \rowcolor{blue!5}
          & Qwen3 VL 8B Instruct & 0.022  & 0.720  & 0.049  & 0.839  & 0.022  & 0.820  \\
    \multirow{3}[0]{*}{LoRA+PE} & InternVL3.5 8B & 0.230  & 0.435  & 0.312  & 0.468  & 0.411  & 0.441  \\
          & MiniCPM-V 4.5 & 0.282  & 0.557  & 0.327  & 0.716  & 0.386  & 0.784  \\
          & Qwen3 VL 8B Instruct & 0.332  & 0.407  & 0.383  & 0.384  & 0.484  & 0.393  \\
    \rowcolor{blue!5}
    & InternVL3.5 8B & 0.128  & 0.412  & 0.429  & 0.200  & 0.547  & 0.154  \\
    \rowcolor{blue!5}
       Full+PE   & MiniCPM-V 4.5 & 0.152  & 0.399  & 0.456  & 0.189  & 0.564  & 0.137  \\
    \rowcolor{blue!5}
          & Qwen3 VL 8B Instruct & 0.142  & 0.418  & 0.483  & 0.152  & 0.634  & 0.088  \\
    \multirow{3}[1]{*}{TA+PE (Ours)} & InternVL3.5 8B & \textbf{0.000}  & 0.504  & \textbf{0.000}  & 0.504  & \textbf{0.000}  & 0.504  \\
          & MiniCPM-V 4.5 & 0.085  & 0.783  & 0.025  & 0.611  & 0.081  & 0.752  \\
          & Qwen3 VL 8B Instruct & 0.034  & \textbf{0.858}  & 0.047  & \textbf{0.861}  & 0.022  & \textbf{0.868}  \\
    \bottomrule
    \end{tabular}%
    
    \label{tab:ablation_adaptation}
\end{table*}
Distinct patterns across adaptation methods are observed in \cref{tab:ablation_adaptation}. Full fine-tuning shows pronounced TSS decline as NITR increases, indicating overfitting to the training toolset. LoRA maintains stable TSS but suffers elevated TRR and fails to achieve substantial gains over PE, suggesting it primarily fits structural patterns rather than enhancing tool-calling capabilities. TA preserves PE capabilities with minimal TRR and demonstrates stable or improving TSS as NITR increases, indicating robust generalization.

\subsubsection{Maximum Iteration and Memory Efficiency Analysis}

 We analyze the relationship between the maximum iteration limit $l_\tau$ and agent execution TSS on the ST-Traj dataset in \cref{tab:ablation_iterations} to observe the trend. Results are averaged across all samples through the LVLMs in experiments. Additionally, we compare memory consumption between our LAMMI framework and direct multi-agent approaches to assess computational efficiency.
  
  \begin{table}[htbp]
      \centering
      \caption{TSS as a function of maximum iteration limit $l_{\tau}$ on ST-Traj dataset.}
     \resizebox{0.47\textwidth}{!}{ \begin{tabular}{ccccccc}
      \toprule
    {Adaption} & $l_{\tau}=3$ & $l_{\tau}=4$ & $l_{\tau}=5$ & $l_{\tau}=6$ & $l_{\tau}=7$ & $l_{\tau}=8$ \\
    \midrule
    w/o TA & 0.602  & 0.640  & 0.660  & 0.675  & 0.683  & 0.691\\
    w/ TA & 0.715  & 0.769  & 0.796  & 0.817  & 0.830  & 0.862\\
    \bottomrule
    \end{tabular}%
      }
      \label{tab:ablation_iterations}
  \end{table}

\begin{figure}[htbp]
    \centering
    \includegraphics[width=0.9\linewidth]{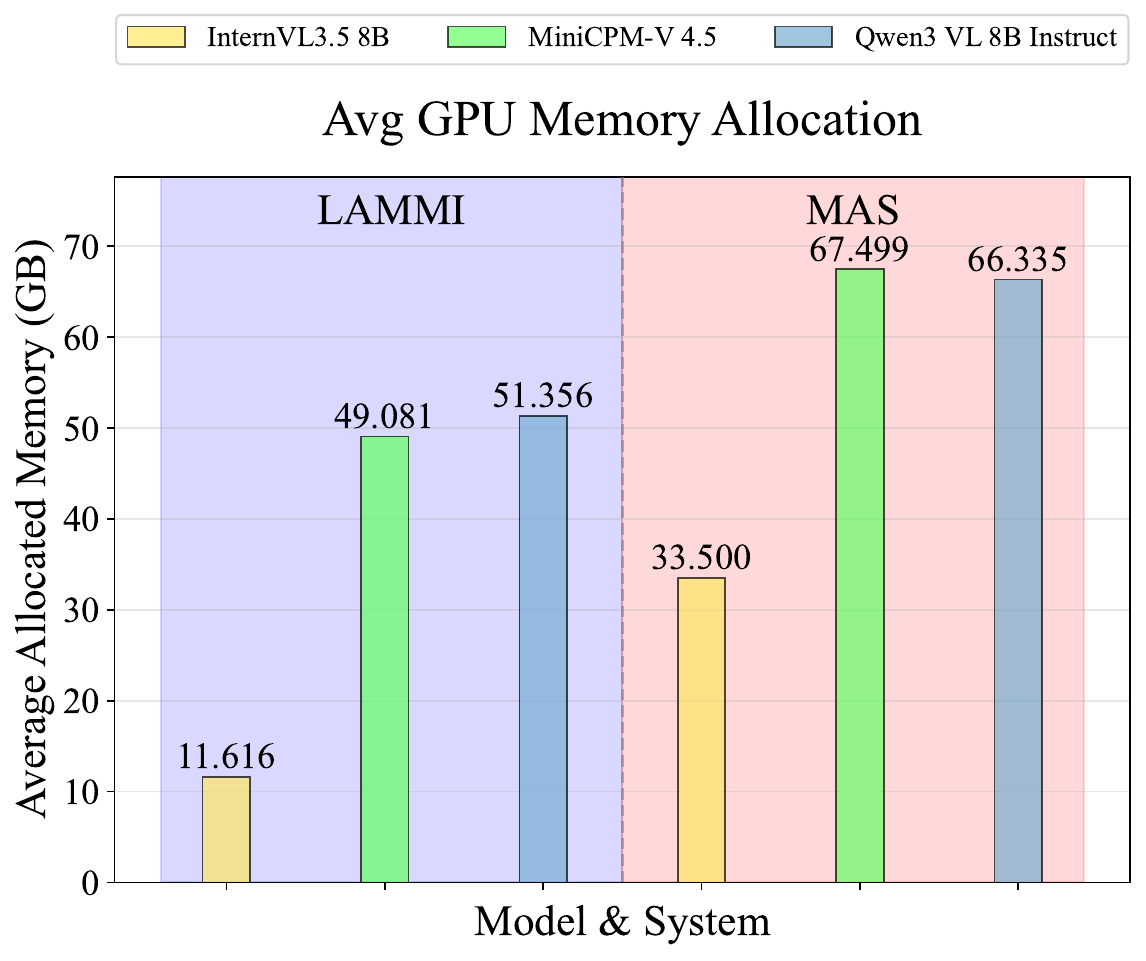}
      \caption{Memory consumption comparison between LAMMI framework and direct multi-agent approaches. The plot shows GPU memory usage differences between LAMMI and standard MAS.}
    \label{fig:ablation_memory}
\end{figure}

\begin{figure*}
    \centering
    \includegraphics[width=1\linewidth]{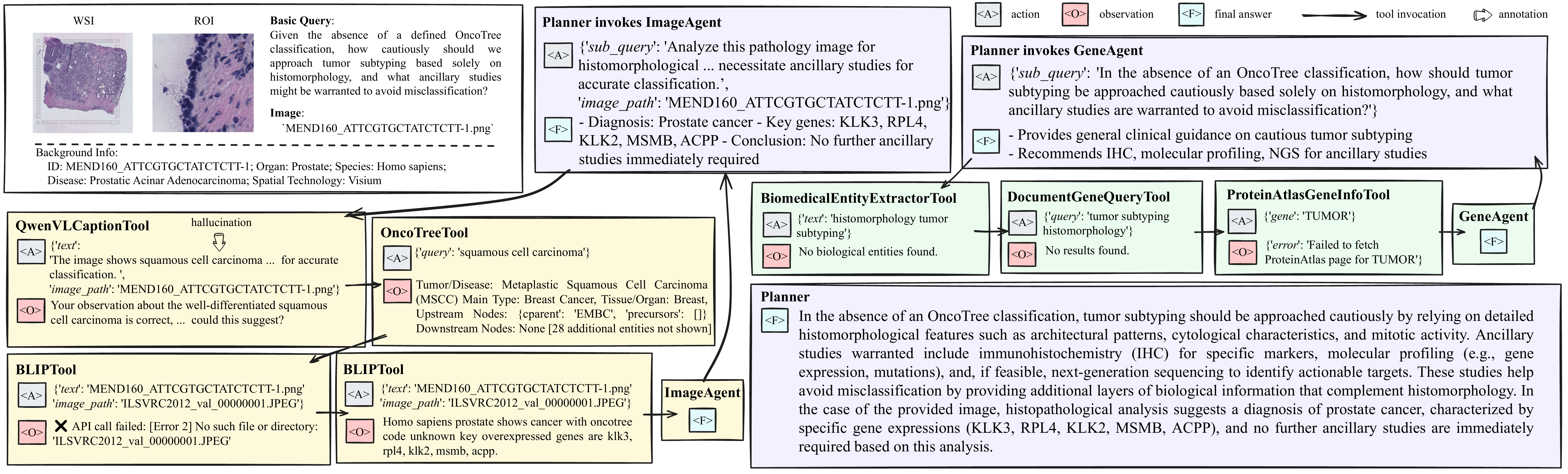}
    \caption{Trajectory visualization of an open-ended research-oriented query showing a semantically coherent solution path that incorporates hallucination occurrence and evidential reasoning through planner-guided exploratory tool invocation by ImageAgent and GeneAgent.}
    \label{fig:case_study}
\end{figure*}

As shown in \cref{tab:ablation_iterations}, the TA fine-tuning approach consistently outperforms the baseline across all $l_\tau$ settings. A rapid growth in TSS is observed as $l_{\tau}$ increases from 3 to 6, with the TA-enhanced method achieving a 0.102 improvement  over this range, compared to a 0.073 gain  for the baseline. This trend reflects the multi-step nature of tasks in ST-Traj, where questions require multiple tool-calling iterations, as evidenced by the ground truth trajectories that span 2 to 8 AENs per sample.

As \cref{fig:ablation_memory} shows, reduced average GPU memory allocation is observed in LAMMI compared with a conventional MAS. With InternVL3.5-8B, the average usage is 11.6 GB, equal to 34.7 percent of the MAS baseline, while MiniCPM-V 4.5 and Qwen3-VL-8B-Instruct consume 72.7 percent and 77.4 percent, respectively. The memory reduction stems from dynamic LLM chain instantiation and a shared weight architecture in LAMMI, where component chains are instantiated on demand and released after task completion. The variation across models reflects their tool-calling tendencies. Instruction-tuned LVLMs such as Qwen-VL and MiniCPM-V exhibit stronger tool-calling behavior, invoking multiple component chains with longer aggregate lifecycles similar to MAS, whereas models favoring direct visual answering, such as InternVL, show higher proportions of direct responses, resulting in shorter chain lifecycles and lower memory consumption.

\subsection{Case Study}
Beyond guided question-answer pairs, we illustrate a representative ST-Traj case drawn from a long trajectory engaging eight successive AENs under a research-oriented questioning paradigm. Iterative tool interactions synthesize visual and molecular information even when the query offers minimal guidance, demonstrating how LAMMI with TA fine-tuning enhances the base LVLM’s exploratory capabilities. The histopathology image is sourced from a Nature study~\cite{erickson2022spatially} on spatially resolved prostate tissue, with a 224-pixel region of interest centered at coordinate 11023, 8088 within a 15821 × 15821 whole-slide image.

As \cref{fig:case_study} shows, the initial research-oriented query, offering minimal guidance and no explicit image-related cues, was routed by the planner to the GeneAgent, whose three exploratory calls failed within the gene API space. This occurred on a compounded hallucination where the Planner's invocation of the ImageAgent and the captioner's observation jointly diverted inference from the prostate ground truth. OncoTree validation, influenced by TA-induced evidential reasoning, revealed the anatomical implausibility of squamous cell carcinoma and prompted molecular cross-checking. Consistent with TA’s objective of minimizing tool redundancy rate, the GeneAgent terminated further exploration rather than persist exhaustively. The mismatch between organ-level context and membrane-level gene expression ultimately corrected the initial hypothesis while maintaining coherent system-level execution, illustrating how coordinated evidence synthesis enables robust recovery from cascading errors.

\section{Conclusion}
\label{sec:conclusion}
We introduce LAMMI-Pathology, a tool-centric agent framework that bridges morphological interpretation with molecular validation for evidence-driven pathology diagnosis. Our TA fine-tuning strategy enables robust structural learning while preserving base model capabilities, achieving superior performance on frontier discovery questions and outperforming existing MAS frameworks. The framework demonstrates multi-agent coordination that synthesizes visual and molecular evidence, establishing a foundation for scalable domain-specific agent systems. Future directions include generalizing this framework to broader medical imaging applications and incorporating diverse molecular data sources to enhance diagnostic reasoning.
{
    \small
    \bibliographystyle{ieeenat_fullname}
    \bibliography{main}
}

\clearpage
\setcounter{page}{1}
\maketitlesupplementary

% \section{Rationale}
% \label{sec:rationale}
% % 
% Having the supplementary compiled together with the main paper means that:
% % 
% \begin{itemize}
% \item The supplementary can back-reference sections of the main paper, for example, we can refer to \cref{sec:intro};
% \item The main paper can forward reference sub-sections within the supplementary explicitly (e.g. referring to a particular experiment); 
% \item When submitted to arXiv, the supplementary will already included at the end of the paper.
% \end{itemize}
% % 
% To split the supplementary pages from the main paper, you can use \href{https://support.apple.com/en-ca/guide/preview/prvw11793/mac#:~:text=Delete%20a%20page%20from%20a,or%20choose%20Edit%20%3E%20Delete).}{Preview (on macOS)}, \href{https://www.adobe.com/acrobat/how-to/delete-pages-from-pdf.html#:~:text=Choose%20%E2%80%9CTools%E2%80%9D%20%3E%20%E2%80%9COrganize,or%20pages%20from%20the%20file.}{Adobe Acrobat} (on all OSs), as well as \href{https://superuser.com/questions/517986/is-it-possible-to-delete-some-pages-of-a-pdf-document}{command line tools}.

\section{AEN Construction and Trajectory Generation}
\label{sec:appendix_aen}

\subsection{Atomic Execution Node Generation}

The AEN generation process creates atomic execution nodes by generating tool actions from queries and executing them, as detailed in \cref{alg:aen_generation}.

\begin{algorithm}[!htbp]
\caption{AEN Generation}
\label{alg:aen_generation}
\begin{algorithmic}[1]
\REQUIRE Query $Q \in \mathcal{Q}$, Tool set $\mathcal{T}$, LLM $f_{\text{LLM}}$
\ENSURE AEN node $N = (Q, A, O)$
\STATE Generate action and input: $(A, \text{input}) \leftarrow f_{\text{LLM}}(Q)$
\STATE Execute tool: $O \leftarrow \text{Execute}(A, \text{input})$
\STATE \textbf{return} $(Q, A, O)$
\end{algorithmic}
\end{algorithm}

\subsection{Trajectory Connection Algorithm}

The trajectory connection algorithm discovers valid connections between AEN nodes by sampling node pairs and evaluating their compatibility, as shown in \cref{alg:node_connection,alg:traj_construct}.

\begin{algorithm}[!htbp]
\caption{Node Connection Discovery}
\label{alg:node_connection}
\begin{algorithmic}[1]
\REQUIRE Node set $\mathcal{N} = \{N_1, \ldots, N_n\}$, Threshold $\theta$, Max pairs $P$
\ENSURE Connection list $\mathcal{C} = \{(i, j, s_{ij}, R_k)\}$
\STATE $\mathcal{C} \leftarrow \emptyset$, $\mathcal{P} \leftarrow \emptyset$, $\text{attempts} \leftarrow 0$
\WHILE{$|\mathcal{P}| < P$ \AND $\text{attempts} < 10 \cdot P$}
    \STATE Sample $(i, j) \sim \text{Uniform}(\mathcal{N} \times \mathcal{N})$, $i \neq j$
    \STATE $\text{attempts} \leftarrow \text{attempts} + 1$
    \IF{$(i, j) \notin \mathcal{P}$ \AND ImageCompatible$(N_i, N_j)$}
        \STATE $\mathcal{P} \leftarrow \mathcal{P} \cup \{(i, j)\}$
        \STATE $s_{ij}, R_k \leftarrow f_{\text{LLM}}(\text{ConnectionPrompt}(N_i, N_j))$
        \IF{$s_{ij} \geq \theta$}
            \STATE $\mathcal{C} \leftarrow \mathcal{C} \cup \{(i, j, s_{ij}, R_k)\}$
        \ENDIF
    \ENDIF
\ENDWHILE
\STATE Sort $\mathcal{C}$ by $s_{ij}$ descending
\STATE \textbf{return} $\mathcal{C}$
\COMMENT{The multiplier 10 is an implementation safeguard; theoretical analysis treats the number of evaluated pairs as $m(n)$ and lets $m(n)$ grow as specified in (A2).}
\end{algorithmic}
\end{algorithm}

\begin{algorithm}[!htbp]
\caption{Trajectory Construction}
\label{alg:traj_construct}
\begin{algorithmic}[1]
\REQUIRE Nodes $\mathcal{N}$, Connections $\mathcal{C}$, Max length $K$, Max usage $M$, Max trajectories $T_{\max}$
\ENSURE Trajectories $\mathcal{T} = \{\tau_1, \ldots, \tau_t\}$
\STATE $\mathcal{T} \leftarrow \emptyset$, $\text{used} \leftarrow \emptyset$, $\text{count} \leftarrow \text{zeros}(|\mathcal{N}|)$
\FOR{$(i, j, s_{ij}, R_k) \in \mathcal{C}$ in descending score order}
    \IF{$|\mathcal{T}| \geq T_{\max}$}
        \STATE \textbf{break}
    \ENDIF
    \IF{$(i, j) \in \text{used}$ \OR $\text{count}[i] \geq M$ \OR $\text{count}[j] \geq M$}
        \STATE \textbf{continue}
    \ENDIF
    \STATE $\tau \leftarrow [N_i, N_j]$ \COMMENT{Initialize with source and target}
    \STATE $\tau[0].\text{step} \leftarrow 0$, $\tau[1].\text{step} \leftarrow 1$, $\tau[1].\text{reasoning} \leftarrow R_k$
    \STATE $\text{current} \leftarrow j$, $\ell \leftarrow 2$
    \WHILE{$\ell < K$}
        \STATE $(j^*, s^*, R^*) \leftarrow \arg\max_{(j, s, R) \in \text{Candidates}(\text{current}, \mathcal{C})} s$
        \IF{$s^* = 0$}
            \STATE \textbf{break}
        \ENDIF
        \STATE $N^* \leftarrow N_{j^*}$, $N^*.\text{step} \leftarrow \ell$, $N^*.\text{reasoning} \leftarrow R^*$
        \STATE $\tau \leftarrow \tau \cup [N^*]$, $\text{current} \leftarrow j^*$, $\ell \leftarrow \ell + 1$
    \ENDWHILE
    \STATE $\text{final\_answer} \leftarrow f_{\text{LLM}}(\text{FinalAnswerPrompt}(\tau))$
    \STATE $\tau \leftarrow \tau \cup [\text{final\_answer}]$
    \STATE $\mathcal{T} \leftarrow \mathcal{T} \cup \{\tau\}$, $\text{used} \leftarrow \text{used} \cup \{(i, j)\}$
    \STATE Update $\text{count}$ for all nodes in $\tau$
\ENDFOR
\STATE \textbf{return} $\mathcal{T}$
\COMMENT{Overall time complexity: $O(|\mathcal{C}| \cdot K)$ under greedy extension.}
\end{algorithmic}
\end{algorithm}

\subsection{Scalability Analysis}

We analyze the scalability properties of the AEN construction method under idealized assumptions. The following propositions provide conditional guarantees that hold under specific assumptions about the connection score distribution, sampling strategy, and system constraints.

Let $n = |\mathcal{N}|$ denote the number of nodes.

\textbf{Assumptions.} The analysis assumes: (A1) connection scores $s_{ij}$ are i.i.d. with cumulative distribution function $F$ (or satisfy weak-dependence conditions such that the asymptotic distribution of extremes remains valid); (A2) the number of evaluated candidate pairs $m(n)$ satisfies $m(n) \to \infty$ as $n \to \infty$, and in particular $m(n) = \Omega(n^\alpha)$ for some $\alpha > 0$ (typical analysis uses $\alpha = 2$); (A3) image compatibility constraints do not reduce the effective candidate pool below $\Omega(n^2)$; (A4) node usage limits $M$ are constant or grow sublinearly with $n$ (i.e., $M = o(n)$).

\textbf{Proposition 1: Connection Quality Monotonicity (Conditional).} Under assumptions (A1)--(A2), let $\mathcal{C}_n$ denote the connection set discovered from $n$ nodes. As $n$ increases, the expected maximum connection score $\mathbb{E}[\max_{(i,j) \in \mathcal{C}_n} s_{ij}]$ increases.

\textbf{Analysis.} The number of candidate pairs grows as $\Theta(n^2)$, while the number of valid connections $|\mathcal{C}_n|$ scales as $\Theta(n^2 \cdot p)$ where $p$ is the probability that a random pair satisfies $s_{ij} \geq \theta$ and image compatibility. Under (A1), assuming $F$ admits a density $f$ (otherwise use the general order-statistic expression), the maximum of $|\mathcal{C}_n|$ independent samples has expected value:
\begin{equation}
\mathbb{E}[\max_{(i,j) \in \mathcal{C}_n} s_{ij}] = \int_{-\infty}^{\infty} x \cdot f_{\max}(x) \, dx
\end{equation}
where $f_{\max}(x) = |\mathcal{C}_n| \cdot F(x)^{|\mathcal{C}_n|-1} \cdot f(x)$ is the density of the maximum order statistic. Since $f_{\max}$ stochastically dominates $f$ for $|\mathcal{C}_n| > 1$, and $|\mathcal{C}_n| \to \infty$ under (A2), the expected maximum increases with $n$.

\textbf{Practical Considerations.} In practice, LLM outputs may exhibit correlation for similar prompts, and image compatibility may reduce effective candidate pairs. If the sampling budget $P$ is fixed rather than scaling with $n$, the above conclusion may not hold. Empirical validation is recommended to verify these conditions in specific deployment scenarios.

\textbf{Proposition 2: Step Distribution Diversity (Conditional).} Under assumptions (A1)--(A4), let $D_k(n)$ denote the number of trajectories of length $k$ generated from $n$ nodes. As $n$ increases, the step distribution becomes more uniform in the sense that longer trajectories become more feasible.

\textbf{Analysis.} We model the connection graph as $G(n, p_n)$ where edges exist independently with probability $p_n = \Pr(s_{ij} \geq \theta \text{ and image compatibility})$. The extension probability at step $\ell$ is $P(\ell) = \mathbb{P}(\exists j: (i_\ell, j, s_{ij}, R_k) \in \mathcal{C}_n \text{ s.t. }\allowbreak s_{ij} \geq \theta \text{ and } j \text{ not in trajectory})$, where $i_\ell$ is the current trajectory endpoint. For fixed $n$, $P(\ell)$ decreases as $\ell$ increases due to node usage constraints (each node may appear at most $M$ times). A sufficient condition for $P(\ell) \to 1$ is $p_n n \to \infty$ as $n \to \infty$. If moreover $M = o(n)$ (as in A4), then for each fixed $\ell < K$ we have $P(\ell) \to 1$ as $n \to \infty$, making longer trajectories feasible and leading to a more uniform step distribution.

\textbf{Practical Considerations.} The greedy extension strategy introduces dependencies between steps, and the actual distribution depends on the specific connection graph topology. The variance bound in the original formulation requires additional assumptions about the connection graph structure and extension probabilities. In practice, the step distribution improvement should be validated through empirical measurements.

\textbf{Corollary 3: Trajectory Quality (Conditional).} Under assumptions (A1)--(A4), and the additional condition that greedy local decisions do not systematically induce adverse global conflicts, the average trajectory quality, measured by connection score sum $Q(\tau) = \sum_{(i,j) \in \tau} s_{ij}$, tends to increase with $n$.

\textbf{Analysis.} By Proposition 1, each connection in a trajectory has higher expected score as $n$ increases. By Proposition 2, trajectories can achieve greater lengths under the idealized model. Since trajectories are constructed greedily by selecting maximum-score connections at each step, $\mathbb{E}[Q(\tau)] = \sum_{k=1}^{|\tau|} \mathbb{E}[s_{i_k j_k}]$ tends to increase with $n$. Note that without the additional condition on greedy decisions, only an expected or upper-bound improvement can be guaranteed.

\textbf{Practical Considerations.} Greedy selection does not guarantee global optimality, and larger $n$ may introduce conflicting high-score connections that lead to local optima. The actual trajectory quality improvement should be validated empirically, especially when budget constraints or image compatibility significantly restrict the candidate pool.

\textbf{Empirical Validation.} The theoretical analysis provides conditional guarantees under idealized assumptions. In practice, we recommend empirical validation across different node set sizes, sampling budgets, image compatibility regimes, and LLM configurations to verify the scalability properties in specific deployment scenarios. The conditional nature of these results underscores the need to align theoretical assumptions with empirical realities when deploying large-scale AEN-based systems.

\section{Tool Specifications}
\label{sec:tool_specs}

All open-source tools and APIs deployments used in this work adhere to standards suitable for integration in reproducible and verifiable scientific research. Table~\ref{tab:tools} provides a comprehensive overview of all tools integrated into the LAMMI framework, including their sources, training data, and descriptions provided in the agent templates.

\paragraph{BLIP-based Patch-to-Gene Tool and CLIP Image-Text Matching Tool.}  
Within our agent tool-calling framework, we implement two complementary vision-language tools for pathology image analysis. The BLIP-based Patch-to-Gene Tool maps pathology ROI patches to corresponding gene expression representations using the BLIP architecture~\cite{li2022blipbootstrappinglanguageimagepretraining}, which is widely adopted as a reference standard in spatial pathology studies. The visual embedding is processed by BLIP's text decoder, reformulated to generate a continuous vector representing the predicted gene expression profile for that patch. This approach allows the agent to retrieve molecular information directly from histological patterns without requiring pre-defined gene lists during inference stage. The tool is trained on spatial transcriptomics datasets, including HEST and STimage-1K4M. To account for variability in sequencing depth across patches, raw counts are first normalized to a consistent scale, and these normalized counts are then used to supervise the generation of patch-level gene embeddings. This normalization ensures that differences in total transcript counts do not confound the learning of morphological-to-molecular associations.

The CLIP Image-Text Matching Tool complements the BLIP tool by providing image-text similarity matching capabilities. It utilizes OpenCLIP ViT-H-14 model~\cite{Radford2021LearningTV} fine-tuned on pathogenomics datasets with training data consistent with BLIPTool. The tool encodes input pathology images and compares them against a predefined set of text candidates describing pathological ROI characteristics, including species, tissue types, and high-expression gene lists. The model computes similarity scores between image embeddings and text embeddings, ranking candidates by probability. This enables the agent to retrieve relevant pathological descriptions and gene expression patterns for given images, supporting image-based retrieval and classification tasks in spatial pathology analysis.

\paragraph{Qwen-VL Caption Tool.}
The Qwen-VL Caption Tool is integrated into our agent framework to provide vision-language question answering capabilities for pathology images. It leverages the Qwen-VL-2.5 vision-language model~\cite{Bai2025Qwen25VL}, which is fine-tuned on large-scale pathology image-text caption pairs including PathCap with 0.17 million samples, Quilt-Instruct with 0.096 million samples, and PathGen with 0.14 million samples. The tool accepts a text instruction or question along with an image path, and returns the model's textual output describing the image content or answering the question. This enables the agent to perform natural language-based image understanding tasks, including image captioning, visual question answering, and detailed image description generation for pathology images.

\paragraph{OncoTree Tool.}
The OncoTree tool~\cite{Kundra2021Oncotree} is integrated into our agent framework to provide standardized cancer type taxonomy and hierarchical disease mapping. It queries the MSKCC OncoTree API to retrieve structured information about tumor types, diseases, and tissues. For each matched result, the tool returns information including tumor name, OncoTree code, main type classification, tissue and organ mapping, upstream nodes consisting of parent nodes and precursors, and downstream nodes representing children in the hierarchy. This enables the agent to interpret or convert free-text tumor descriptions into structured OncoTree codes, facilitating consistent cancer subtype identification and hierarchical navigation across datasets.

\paragraph{Pathway Knowledge Graph Tool.}
The Pathway Knowledge Graph Tool provides gene pair relationship analysis through local subgraph overlap computation. It operates on a local pathway knowledge graph implemented as a NetworkX DiGraph containing 18,588 nodes and 874,888 edges with 6 relation types: controls-expression-of, in-complex-with, controls-phosphorylation-of, controls-state-change-of, catalysis-precedes, and controls-transport-of. For a given pair of genes, the tool extracts ego subgraphs representing neighborhoods of specified radius around each gene node, considering edge directionality. It then computes weighted Jaccard similarity between these subgraphs, where edge weights are incorporated into the similarity calculation. The tool supports both outgoing and incoming edge directions, allowing flexible analysis of gene regulatory relationships. This enables the agent to assess pathway relationships and functional associations between gene pairs based on their local network topology, supporting gene interaction analysis and pathway-based reasoning.

\paragraph{Ensembl to Database Tool.}
The Ensembl to Database Tool enables cross-reference mapping between Ensembl gene identifiers and external database entries. It utilizes the Ensembl REST API~\cite{Yates2019Ensembl} to fetch related database information for a given Ensembl ID.  For the specified Ensembl ID, the tool queries the xrefs/id endpoint to retrieve all cross-references, then filters results by database name with GeneCards as the default. The returned information includes synonyms, display ID, info type, info text, database name, primary ID, database display name, version, and description. This enables the agent to map gene identifiers across multiple authoritative databases, facilitating data integration and cross-database queries in biomedical research.

\paragraph{Protein Atlas Gene Information Tool.}
The Protein Atlas Gene Information Tool retrieves comprehensive gene and protein information from the Human Protein Atlas~\cite{Mathias2015ProteinAtlas}. The tool fetches the gene summary page from Protein Atlas and parses structured information from HTML tables, extracting three main information blocks: General Information including gene name, synonyms, gene description, protein class, and predicted location; Gene Information including chromosome location, cytoband, Ensembl ID, Entrez gene, HGNC, UniProt, neXtProt, GeneCards, and Antibodypedia; and Protein Function including protein function, molecular function, biological process, ligand, and gene summary. This enables the agent to retrieve detailed protein function, localization, and expression information for genes.

\paragraph{Document Gene Query Tool.}
The Document Gene Query Tool provides gene information retrieval with article-correlated summaries through the MyGene API~\cite{Xin2016MyGene}. It returns the top-ranked gene entries with correlation scores, including Entrez ID, gene name, and detailed summaries extracted from scientific literature. This enables the agent to retrieve literature-backed gene summaries and descriptions.

\paragraph{Biomedical Entity Extractor.}
The Biomedical Entity Extractor is integrated into our agent framework to provide standardized biological entity recognition and mapping from natural language text to structured biomedical knowledge. It leverages the Monarch Initiative v3 API~\cite{Putman2023Monarch,Putman2024Monarch}, which integrates data from multiple authoritative biomedical databases and ontologies including the Monarch Disease Ontology, the Human Phenotype Ontology, Uberon, and other standardized biomedical ontologies. The tool performs a two-step process. First, it calls the Monarch API's entities endpoint to extract entity mentions and their IDs from the input text. Then, for each unique entity ID, it queries the Monarch API's entity endpoint to retrieve comprehensive entity information including standardized ID, name, category, description, symbol, and synonyms.

\begin{table*}[htbp!]
    \centering
    \caption{Tool Specifications and Sources}
    \label{tab:tools}
    {\small
    \begin{tabular}{p{4cm}p{4.8cm}p{7cm}}
    \toprule
    \textbf{Tool Name} & \textbf{Source} & \textbf{Description} \\
    \midrule
    \multicolumn{3}{c}{\textbf{ImageAgent}} \\
    \midrule
    BLIPTool & Model Training: 3.68M HEST ROI, 15k Visium ROI, 10K ST ROI with corresponding high-expression genes (based on RNA-seq data) & Provides genetic answers for pathology images with questions related to gene expressions. Input: text instruction/question and image name. Output: basic genetic information including high-expression genes, species, and tissue. \\
    \midrule
    CLIPTool & Model Training: OpenCLIP ViT-H-14 fine-tuned on pathogenomics dataset (training data consistent with BLIPTool) & OpenCLIP image-text matching tool. Strictly accepts image file paths as input. Do not provide image descriptions or text; only valid paths allowed. \\
    \midrule
    QwenVLCaptionTool & Model Training: Image-text caption pairs including PathCap 0.17M, Quilt-Instruct 0.096M, and PathGen 0.14M samples; Base model: Qwen-VL-2.5 vision-language model  & Provides answers for pathology images with associated questions. Input: text instruction/question and image path. Returns model's textual output. \\
    \midrule
    OncoTreeTool & API: MSKCC OncoTree (oncotree\allowbreak.mskcc\allowbreak.org) & Query OncoTree knowledge graph. Input: tumor, disease, or tissue keyword. Returns upstream/downstream nodes and tissue mapping. \\
    \midrule
    \multicolumn{3}{c}{\textbf{GeneAgent}} \\
    \midrule
    PathwayKGTool & Knowledge Graph: Local pathway knowledge graph (DiGraph with 18,588 nodes and 874,888 edges) with 6 relation types (controls-expression-of, in-complex-with, controls-phosphorylation-of, controls-state-change-of, catalysis-precedes, controls-transport-of); NetworkX-based weighted Jaccard similarity & Query Pathway Commons neighborhood for a given gene and return network edges as plain text. \\
    \midrule
    EnsemblToDatabaseTool & API: Ensembl REST API (rest\allowbreak.ensembl\allowbreak.org) & Fetches related database info for the Ensembl ID using Ensembl REST API. Returns synonyms, display ID, info type, info text, database name, primary ID, and version. \\
    \midrule
    ProteinAtlasGeneInfoTool & API: Human Protein Atlas (proteinatlas.org) & Fetches gene information from Protein Atlas using gene name. Returns protein function, chromosome location, UniProt, Antibodypedia, protein function, and gene summary. \\
    \midrule
    DocumentGeneQueryTool & API: MyGene.info v3 (mygene.info) & Query gene information from MyGene.info using gene symbol or keyword. Useful for retrieving article-correlated gene summaries. Encouraged to summarize information most correlated to query. \\
    \midrule
    BiomedicalEntityExtractorTool & API: Monarch Initiative API v3 (mon\allowbreak archinitiative.org) & Extract biological entities (genes, diseases, proteins, etc.) from text using Monarch Initiative API. Returns formatted markdown with entity details including ID, category, description, and synonyms. \\
    \bottomrule
    \end{tabular}
    }
\end{table*}

\section{Parameter Efficiency and Computational Complexity}

\subsection{Parameter Efficiency Analysis}
The adapter introduces $3d$ trainable parameters per layer, corresponding to the three segment-specific scaling vectors $\boldsymbol{\gamma}_{\text{thought}}$, $\boldsymbol{\gamma}_{\text{action}}$, and $\boldsymbol{\gamma}_{\text{input}}$. For a model with $L$ layers, the total adapter parameter count is $\Theta(Ld)$, compared to $\Theta(Ld^2)$ for standard FFN weights (assuming feed-forward dimension $d_{\text{ff}} \approx 4d$). The parameter efficiency ratio quantifies the reduction,
\begin{equation}
\label{eq:parameter_efficiency}
\rho = \frac{|\Theta_{\text{adapter}}|}{|\Theta_{\text{full}}|} = \frac{3Ld}{12Ld^2} = \frac{1}{4d}.
\end{equation}
For $d = 4096$, this yields $\rho \approx 0.0061\%$, demonstrating substantial parameter reduction. Compared to LoRA~\cite{hu2022lora}, which introduces $2rd$ parameters per layer (typically $r = 8$), our method achieves a $5.3\times$ reduction: $3d = 12,288$ versus $2rd = 65,536$ parameters per layer, while providing explicit structural awareness through segment masks.

\subsection{Computational Complexity}
The modulation computation requires $O(BL_{\text{seq}}d)$ operations for element-wise multiplications, where $L_{\text{seq}} = L_v + L_t$ includes both visual and textual tokens. Segment mask generation involves $O(BL_t)$ token matching operations on text tokens only, performed by scanning input sequences to locate trajectory markers (\texttt{Thought}, \texttt{Action}, \texttt{Action Input}). Compared to the $O(BL_{\text{seq}}d^2)$ FFN forward pass, the relative overhead is
\begin{equation}
\label{eq:computational_overhead}
\frac{O(BL_{\text{seq}}d) + O(BL_t)}{O(BL_{\text{seq}}d^2)} = O\left(\frac{1}{d}\right) \approx 0.024\%,
\end{equation}
where $d$ equals to $4096$. The negligible computational overhead, combined with parameter efficiency, makes the adapter suitable for resource-constrained deployment scenarios.

\section{Evaluation Metrics}
\label{sec:evaluation_metrics}

We establish a comprehensive evaluation framework with six core metrics to assess trajectory quality, tool usage efficiency, and answer correctness. The following formulations define each metric's computation.

\subsection{Trajectory Success Score}

For trajectory data, we measure execution success by combining output validity and tool call success,

\begin{equation}
S(\tau) = 0.5 \cdot \mathbf{1}[\text{valid\_output}] + 0.5 \cdot \frac{N_{\text{success}}}{N_{\text{total}}},
\end{equation}
where $N_{\text{success}}$ and $N_{\text{total}}$ denote successful and total tool calls, respectively. A trajectory is considered successful if $S(\tau) \geq 1.0$.

\subsection{Tool Redundancy Rate}

We quantify redundant tool usage by measuring the proportion of tool calls with similar inputs,

\begin{equation}
R = \frac{|\{(i,j): \text{sim}(u_i, u_j) > \theta, t_i = t_j\}|}{|\{(i,j): i < j\}|},
\end{equation}
where $t_i$ and $u_i$ denote the tool name and input content for call $i$, $\text{sim}(\cdot, \cdot)$ is a Jaccard similarity function, and $\theta$ is a threshold equals to 0.7. The redundancy rate aggregates across all samples: $\bar{R} = \frac{1}{N}\sum_{k=1}^{N} R_k$.

\subsection{Tool Consistency F1 Score}

For trajectory data, we compare expected tools $\mathcal{T}_{\text{exp}}$ from ground truth with actual tools $\mathcal{T}_{\text{act}}$ from model output,

\begin{align}
\text{Precision} &= \frac{|\mathcal{T}_{\text{exp}} \cap \mathcal{T}_{\text{act}}|}{|\mathcal{T}_{\text{act}}|}, \\
\text{Recall} &= \frac{|\mathcal{T}_{\text{exp}} \cap \mathcal{T}_{\text{act}}|}{|\mathcal{T}_{\text{exp}}|}, \\
\text{F1} &= \frac{2 \cdot \text{Precision} \cdot \text{Recall}}{\text{Precision} + \text{Recall}}.
\end{align}
The average F1 score across trajectory samples is: $\overline{\text{F1}} = \frac{1}{N}\sum_{k=1}^{N} \text{F1}_k$.

\subsection{Answer Consistency Score}

We employ LLM-based evaluation to measure semantic consistency between reference answer $a_{\text{ref}}$ and model answer $a_{\text{model}}$:

\begin{equation}
C(a_{\text{ref}}, a_{\text{model}}) = f_{\text{LLM}}(\text{Consistency}(a_{\text{ref}}, a_{\text{model}})),
\end{equation}
where $f_{\text{LLM}}$ returns a score in $[0, 1]$. The average consistency score is: $\bar{C} = \frac{1}{N}\sum_{k=1}^{N} C_k$.

\subsection{Hallucination Rate}

We detect hallucinations in model responses using LLM evaluation:

\begin{equation}
H(q, a_{\text{model}}, \tau) = f_{\text{LLM}}(\text{HallucinationPrompt}(q, a_{\text{model}}, \tau)),
\end{equation}
where $q$ is the question, $a_{\text{model}}$ is the model answer, $\tau$ is the full reasoning trajectory, and $H \in [0, 1]$ indicates hallucination severity. The hallucination rate is: $\bar{H} = \frac{1}{N}\sum_{k=1}^{N} \mathbf{1}[H_k > 0.5]$.

\subsection{Multiple Choice F1 Score}

For multiple-choice questions with options $\mathcal{O} = \{o_1, \ldots, o_K\}$, we compute:

\begin{equation}
\text{Acc} = \begin{cases}
1.0 & \text{if } a_{\text{pred}} \text{ exactly matches } a_{\text{correct}} \\
\text{sim}(a_{\text{pred}}, a_{\text{correct}}) & \text{otherwise}
\end{cases},
\end{equation}
where $\text{sim}(\cdot, \cdot)$ is LLM-based semantic similarity evaluation. The F1 score equals accuracy for multiple-choice questions: $\text{F1} = \text{Acc}$.

%%%%%%%%%%%%%%%%%%%%%%%%%%%%%%%%%%%%%%%%%%%%%%%%%%%%%%%%%%%%
\section{LAMMI Architecture Specification}
\label{sec:lammi_architecture}

\subsection{Template Specification}
The LAMMI architecture employs a hierarchical agent system with specialized prompt templates, including the Planner Agent template, the Component Agent base template (inherited by GeneAgent and ImageAgent), and their domain-specific extensions. This section provides detailed specifications of the prompt templates used in our experiments.

\subsubsection{Planner Agent Template}

The Planner Agent serves as the master coordinator that analyzes user queries and delegates tasks to specialized agents. The prompt template adopts a ReACT-based PE design with \texttt{Thought}, \texttt{Action}, and \texttt{Action Input} loops. The template is shown in ~\cref{fig:planner_agent_template}.

\begin{figure}
    \centering
\begin{tcolorbox}[
  colback=gray!5,
  colframe=black!30,
]

\textless Context Begin\textgreater: \\
You are a master planner. Analyze the user's query and decide whether to answer directly or delegate to a specialized agent. \\

Please strictly follow the ReACT format, and give dialogues in Thought/Action/Action Input loop. \\

Available Specialist Agents: \\
- ImageAgent: [description] \\
- GeneAgent: [description] \\

If you can answer the query directly, respond with: \\
\textbf{Final Answer}: [your direct answer] \\

If the query requires a specialist, strictly follow the ReACT format: \\
\textbf{Thought}: The information obtained so far regarding the user query, along with the sub-agents that need to be invoked next and the aspects that require further discussion. \\
\textbf{Action}: [agent\_name] \\
\textbf{Action Input}: [detailed sub query] \\

Example: \\
User query: "What is the role of TP53 in cancer?" \\

\textbf{Thought}: To answer the effects of TP53, I need to call sub-agent xx to investigate yy. \\
\textbf{Action}: GeneAgent \\
\textbf{Action Input}: Explain the role and function of TP53 gene in cancer development. \\

Once you believe the summaries from the sub-agents are sufficient to answer the user's question, please output \textbf{Final Answer} and provide the summarized answer to the question. Repeatedly calling the same sub-agent and asking the same query is meaningless. \\

Begin! \\

User Query: \textless Question\textgreater: \texttt{\{query\}} \\
\textless Image\textgreater: \texttt{\{img\}} \\
Extra parameters: \\
\texttt{\{extra\_params\_str\}} \\
\textbf{Thought}: \texttt{\{agent\_scratchpad\}}

\end{tcolorbox}
\caption{Planner Agent prompt template.}
\label{fig:planner_agent_template}
\end{figure}

\subsubsection{Component Agent Template}

Component agents, including GeneAgent and ImageAgent, inherit from a base template that enforces the ReACT format structure. The base template provides the common framework for tool-calling interactions as follows. The template is shown in ~\cref{fig:component_agent_template}.

\begin{figure}
    \centering
\begin{tcolorbox}[
  colback=gray!5,
  colframe=black!30,
]

\textless Context Begin\textgreater: \\
Please follow strict ReACT format (Thought/Action/Action Input) to solve tasks. \\

\textbf{Thought:} what should I do? \\
\textbf{Action:} the subagent or tool to use, return only the subagent or tool name without any redundance \\
\textbf{Action Input:} the input to the subagent or tool \\

(Observation will be provided after tool execution) \\

Launch Thought/Action/Action Loop as needed, repeatedly generating the same action and action input is not suggested. \\

If you think you have enough information or you cannot solve the query anymore, please provide: \\
\textbf{Final Answer:} your final answer to the question \\

You MUST NOT output "Final Answer" in the same step as "Action" and "Action Input". If you do so, your output will be considered INVALID and ignored. \\

Specific requirements are given below:

\texttt{\{extra\_instruction\}} \\

\texttt{\{agent\_scratchpad\}}

\end{tcolorbox}
\caption{Component Agent base prompt template.}
\label{fig:component_agent_template}
\end{figure}

\subsubsection{GeneAgent Template}

The GeneAgent specializes in gene-related queries and uses a ReACT-based tool-calling interface. It inherits from the base template and adds domain-specific instructions. The template is shown in ~\cref{fig:geneagent_template}.

\begin{figure}
    \centering
\begin{tcolorbox}[
  colback=gray!5,
  colframe=black!30,
]

[Component Agent Heading Template] 

You are a gene expert. Answer gene-related queries by tool calling. \\
You have access to the following tools:

[Tool descriptions with names and descriptions] \\

\textless Question\textgreater: \texttt{\{query\}} \\
\textless Image\textgreater: \texttt{\{img\}} \\

Extra parameters: \\
\texttt{\{extra\_params\_str\}} \\

Use the following format to solve a given question: \\
\textbf{Thought:} what should I do? \\
\textbf{Action:} must be one of [tool\_names] \\
\textbf{Action Input:} the input to the tool \\

(Observation will be provided after tool execution)

\end{tcolorbox}
\caption{GeneAgent prompt template.}
\label{fig:geneagent_template}
\end{figure}

\subsubsection{ImageAgent Template}

The ImageAgent focuses on histopathology image analysis and encourages hypothesis generation based on visual evidence. It follows the same ReACT structure as the GeneAgent but with pathology-specific instructions. The template is shown in ~\cref{fig:imageagent_template}.

\begin{figure}
    \centering
\begin{tcolorbox}[
  colback=gray!5,
  colframe=black!30,
]
[Component Agent Heading Template] 

You are an image expert. Analyze histopathology slides by tool calling. 
You are encouraged to give hypotheses based on the pathology image you see.

You have access to the following tools:

[Tool descriptions with names and descriptions]\\

\textless Question\textgreater: \texttt{\{query\}} \\
\textless Image\textgreater: \texttt{\{img\}} \\

Extra parameters: 

\texttt{\{extra\_params\_str\}}\\

Use the following format to solve a given question:

\textbf{Thought}: what should I do? \\
\textbf{Action}: must be one of [tool\_names] \\
\textbf{Action Input}: the input to the tool \\

(Observation will be provided after tool execution)

\end{tcolorbox}
\caption{ImageAgent prompt template.}
\label{fig:imageagent_template}
\end{figure}

\subsubsection{Execution Flow and Text Parsing Template}
The agent execution framework employs a two-stage parsing mechanism to convert free-form LLM text outputs into structured data. First, regular expressions extract \texttt{Action} and \texttt{Action Input} segments from the raw text. Then, for component agents, the \texttt{Action Input} text is parsed into structured JSON format conforming to tool-specific Pydantic schemas using the Outlines library~\cite{willard2023efficient}, which leverages a dedicated LLM (Qwen2.5-7B) to ensure schema compliance. The parsing LLM is prompted with the template shown in ~\cref{fig:parsing_template}.

\begin{figure}
    \centering
\begin{tcolorbox}[
  colback=gray!5,
  colframe=black!30,
]

Do not interpret, or infer new information from the input. You are not allowed to add extra information. \\
You must only fill the fields according to the schema below. \\

=== SCHEMA === \\
\texttt{\{schema\_json\}} \\

=== INPUT === \\
\texttt{\{raw\_input\}} \\

Now return the JSON object:

\end{tcolorbox}
\caption{Parsing LLM prompt template for schema compliance.}
\label{fig:parsing_template}
\end{figure}
% This prompt template enforces strict adherence to the provided schema fields, prohibiting interpretation or inference of new information beyond the input text, and requiring output of only the JSON object that fills schema fields according to the input.

%%%%%%%%%%%%%%%%%%%%%%%%%%%%%%%%%%%%%%%%%%%%%%%%%%%%%%%%%%%%
\section{Curated Dataset Details}
\label{sec:curated_dataset_details}

This section presents detailed examples from our two curated datasets: PathSpatial-DocQA and ST-Traj. We showcase typical samples and ground truth examples to illustrate the dataset characteristics, query complexity, and expected trajectory structures.

\subsection{PathSpatial-DocQA Dataset}

The PathSpatial-DocQA dataset addresses frontier questions in pathology-molecular diagnosis integration. Below we present typical samples from the dataset, including the questions and reference answers. The samples are organized into categories, including causal mechanism reasoning, multi-entity relation reasoning, hypothesis verification, and temporal and sequential reasoning.

\subsubsection{Causal Mechanism Reasoning}

\begin{figure}
    \centering
\begin{tcolorbox}[
  colback=gray!5,
  colframe=black!30,
]

\textbf{Question}: Can TGF$\beta$ signaling loss in cancer cells convert the tumor microenvironment into a fibro-macrophage state and exclude T cells? \\
{image\_path}: GSE193460\_GSM5808056\_61x69-12\allowbreak x20.png
\begin{center}
\includegraphics[height=0.3\textwidth]{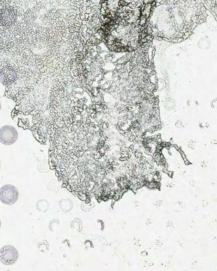}
\end{center}
\textbf{Answer}: True. When TGF$\beta$ signaling is lost in cancer cells, the tumor microenvironment undergoes remodeling characterized by activation of fibroblasts and accumulation of macrophages, while T cells are excluded. This shift is associated with increased fibrotic stroma and reduced anti-tumor immunity, indicating that TGF$\beta$ responsiveness in cancer cells directly shapes the immune landscape of the tumor.
\end{tcolorbox}
\caption{Typical PathSpatial-DocQA dataset example for case STimage-1K4M\_35290801\_page\_2\_2.}
\label{fig:STimage-1K4M_35290801_page_2_2}
\end{figure}

\begin{figure}
    \centering
\begin{tcolorbox}[
  colback=gray!5,
  colframe=black!30,
]

\textbf{Question}: Is the presence of TLSs (tertiary lymphoid structures) in the liver associated with chronic inflammation and fibrosis in patients with HS (hepatic steatosis)? \\
{image\_path}: GSE249729\_GSM7962129\_38x45-47\allowbreak x54.png
\begin{center}
\includegraphics[height=0.3\textwidth]{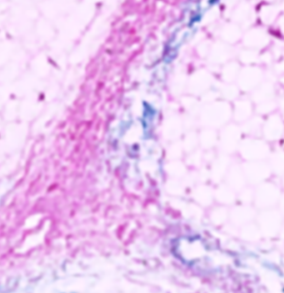}
\end{center}

\textbf{Answer}: True. TLSs are organized aggregates of immune cells that form in nonlymphoid tissues, including the liver, during chronic inflammatory conditions such as HS. Their presence correlates with disease severity, including fibrosis, and they are recognized as a histological feature of HS lesions, suggesting their role in sustaining local immune responses that drive tissue damage.

\end{tcolorbox}
\caption{Typical PathSpatial-DocQA dataset example for case STimage-1K4M\_38113104\_page\_9\_0.}
\label{fig:STimage-1K4M_38113104_page_9_0}
\end{figure}

Typical samples are shown in ~\cref{fig:STimage-1K4M_35290801_page_2_2,fig:STimage-1K4M_36283023_page_17_2,fig:STimage-1K4M_38113104_page_9_0}.

\begin{figure}
    \centering
\begin{tcolorbox}[
  colback=gray!5,
  colframe=black!30,
]

\textbf{Question}: Can triple-negative breast cancer (TNBC) be targeted therapeutically by inhibiting the NDRG1 gene? \\
\textbf{Answer}: False. NDRG1 functions as a tumor suppressor in TNBC; therefore, inhibiting it would likely promote tumor progression rather than suppress it. Therapeutic strategies should aim to restore or enhance NDRG1 function, not inhibit it, as shown in studies of TNBC progression and metastasis.

\end{tcolorbox}
\caption{Typical PathSpatial-DocQA dataset example for case STimage-1K4M\_36283023\_page\_17\_2.}
\label{fig:STimage-1K4M_36283023_page_17_2}
\end{figure}

\subsubsection{Multi-entity Relation Reasoning}

\begin{figure}
    \centering
\begin{tcolorbox}[
  colback=gray!5,
  colframe=black!30,
]

\textbf{Question}: Does the presence of osteopontin correlate with macrophage recruitment in renal papillary tissue? \\
{image\_path}: GSE206306\_GSM6250307\_56x57-61\allowbreak x62.png
\begin{center}
\includegraphics[height=0.3\textwidth]{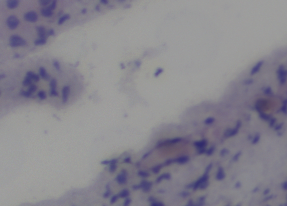}
\end{center}
\textbf{Answer}: True. Osteopontin interacts with macrophages to enhance their chemotaxis and recruitment, and it is co-expressed with macrophage markers such as CD68 and CD45 in renal papillary tissue, indicating a direct association between osteopontin and macrophage infiltration in stone-forming regions. 
\end{tcolorbox}
\caption{Typical PathSpatial-DocQA dataset example for case STimage-1K4M\_37468493\_page\_10\_1.}
\label{fig:STimage-1K4M_37468493_page_10_1}
\end{figure}

\begin{figure}
    \centering
\begin{tcolorbox}[
  colback=gray!5,
  colframe=black!30,
]

\textbf{Question}: Are CD34+ and PDGFR$\alpha$+ fibroblasts found in both the lining and sublining layers of rheumatoid synovium? \\
{image\_path}: GSE249729\_GSM7962129\_44x51-39\allowbreak x46.png
\begin{center}
\includegraphics[height=0.3\textwidth]{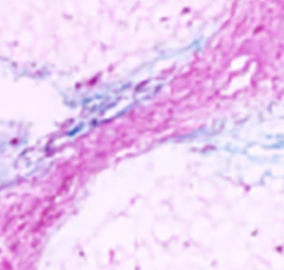}
\end{center}
\textbf{Answer}: True. CD34+ and PDGFR$\alpha$+ fibroblasts are present in both the lining layer (where they are most abundant) and the sublining layer of rheumatoid synovium. Their presence in these regions suggests a role in both synovial expansion and stromal remodeling during chronic inflammation. 
\end{tcolorbox}
\caption{Typical PathSpatial-DocQA dataset example for case STimage-1K4M\_38113104\_page\_4\_1.}
\label{fig:STimage-1K4M_38113104_page_4_1}
\end{figure}

Typical samples are shown in ~\cref{fig:STimage-1K4M_38282415_page_3_0,fig:STimage-1K4M_37468493_page_10_1,fig:STimage-1K4M_38113104_page_4_1}.

\begin{figure}
    \centering
\begin{tcolorbox}[
  colback=gray!5,
  colframe=black!30,
]

\textbf{Question}: Is it true that ER+ breast cancers exhibit less proliferation and more differentiation compared to ER- breast cancers? \\
\textbf{Answer}: True. ER+ breast cancers are characterized by lower proliferation rates and higher differentiation, which contrasts with the more aggressive, poorly differentiated phenotype typically seen in ER- tumors. This distinction is clinically relevant and forms the basis for prognostic classification and therapeutic strategies targeting estrogen receptor signaling.

\end{tcolorbox}
\caption{Typical PathSpatial-DocQA dataset example for case STimage-1K4M\_38282415\_page\_3\_0.}
\label{fig:STimage-1K4M_38282415_page_3_0}
\end{figure}

\subsubsection{Hypothesis Verification}

Typical samples are shown in ~\cref{fig:STimage-1K4M_38067175_page_2_2,fig:HEST_page_25_0}.

\begin{figure}
    \centering
\begin{tcolorbox}[
  colback=gray!5,
  colframe=black!30,
]

\textbf{Question}: Do conventional corticosteroid therapies effectively treat muscular sarcoidosis, and what is the current therapeutic strategy based on? \\
\textbf{Answer}: False. Conventional corticosteroid therapy alone is not sufficient for treating muscular sarcoidosis, as the disease is based on systemic immune dysregulation. Current therapeutic strategies are derived from the treatment of systemic sarcoidosis and include immunosuppressive agents, since corticosteroids alone provide only a first-line approach without addressing the underlying immune-mediated pathology.

\end{tcolorbox}
\caption{Typical PathSpatial-DocQA dataset example for case STimage-1K4M\_38067175\_page\_2\_2.}
\label{fig:STimage-1K4M_38067175_page_2_2}
\end{figure}

\begin{figure}
    \centering
\begin{tcolorbox}[
  colback=gray!5,
  colframe=black!30,
]

\textbf{Question}: Is T cell infiltration into colorectal tumors associated with favorable clinical outcomes? \\
\textbf{Answer}: True. T cell infiltration into colorectal tumors has long been associated with favorable outcomes, suggesting a possible role for immunomodulation in controlling tumor growth. This association implies that the presence of T cells in the tumor microenvironment may contribute to better patient prognosis.

\end{tcolorbox}
\caption{Typical PathSpatial-DocQA dataset example for case HEST\_page\_25\_0.}
\label{fig:HEST_page_25_0}
\end{figure}

\subsubsection{Temporal and Sequential Reasoning}

\begin{figure}
    \centering
\begin{tcolorbox}[
  colback=gray!5,
  colframe=black!30,
]

\textbf{Question}: Is the transcription factor EGR1 involved in the regulation of extracellular matrix (ECM) deposition in fibrotic kidneys? \\
{image\_path}: GSE183456\_GSM6047785\_50x58-49\allowbreak x57.png
\begin{center}
\includegraphics[height=0.3\textwidth]{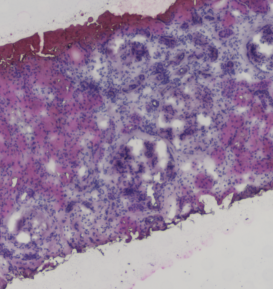}
\end{center}
\textbf{Answer}: True. EGR1 is activated in adaptive fibrosis and contributes to the progression of tubulointerstitial fibrosis through gene expression changes that regulate ECM deposition. This transcription factor plays a fundamental role in fibrosis mechanisms across multiple organ systems. 
\end{tcolorbox}

\caption{Typical PathSpatial-DocQA dataset example for case STimage-1K4M\_37468583\_page\_7\_2.}
\label{fig:STimage-1K4M_37468583_page_7_2}
\end{figure}

Typical samples are shown in ~\cref{fig:STimage-1K4M_36741401_page_2_2,fig:STimage-1K4M_37468583_page_7_2}.

\begin{figure}
    \centering
\begin{tcolorbox}[
  colback=gray!5,
  colframe=black!30,
]

\textbf{Question}: Is it true that the thymus undergoes significant involution after birth, leading to a decline in thymocyte output and reduced thymic function in adulthood? \\
\textbf{Answer}: True. The thymus is most active during infancy and childhood, with high thymocyte output. After birth, it undergoes progressive involution, characterized by a decline in thymocyte production and replacement of lymphoid tissue with adipose tissue. This results in diminished capacity to generate new T cells in adults.

\end{tcolorbox}
\caption{Typical PathSpatial-DocQA dataset example for case STimage-1K4M\_36741401\_page\_2\_2.}
\label{fig:STimage-1K4M_36741401_page_2_2}
\end{figure}

\subsection{ST-Traj Dataset}

\begin{figure*}
    \centering
    \resizebox{\textwidth}{!}{
\begin{tcolorbox}[
  colback=gray!5,
  colframe=black!30,
  fontupper=\fontsize{7pt}{8.4pt}\selectfont,
]
\textbf{Sample ID:} 1384\_8588\_9733\_10239

image\_path: TENX125\_027x090.png

\begin{center}
\includegraphics[height=0.16\textwidth]{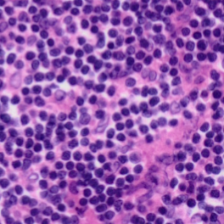}
\end{center}

\textbf{Step 1:} \\
\textbf{Query:} What histological features and cellular morphology in this lymphoid tissue sample are indicative of disease pathology, and how might AI-driven image analysis improve diagnostic precision? \\
\textbf{Thought:} To analyze the histological features and cellular morphology in the lymphoid tissue sample and understand how AI-driven image analysis can improve diagnostic precision, I need to first gather detailed information about the image content. \\
\textbf{Action:} QwenVLCaptionTool \\
\textbf{Action Input:} {\{"text": "Describe the histological features and cellular morphology in this lymphoid tissue sample.", "image\_path": "TENX125\_027x090.png"\}} \\
\textbf{Observation:} The histological features of this lymphoid tissue sample show a dense, diffuse infiltrate of cells with scant cytoplasm, round nuclei, and condensed chromatin. The nucleoli are not prominent, and the cells are tightly packed together. The lack of significant pleomorphism and the absence of mitotic figures suggest a monomorphic population of cells, which is characteristic of a lymphoid neoplasm. The overall architecture of the tissue is effaced by the infiltrate, indicating a disruption of the normal tissue structure. The presence of these features is consistent with a diagnosis of a lymphoid neoplasm, such as a lymphoma.

\textbf{Step 2:} \\
\textbf{Reasoning:} The histological findings reveal a monomorphic, densely packed lymphoid infiltrate with nuclear uniformity and effaced tissue architecture, highly suggestive of a malignant lymphoma. Given the morphological suspicion of a high-grade or aggressive lymphoid neoplasm, understanding the potential underlying genetic aberrations becomes critical for accurate classification, prognostication, and therapeutic planning. Among the most significant molecular alterations in lymphoid malignancies is mutation of the TP53 tumor suppressor gene, which is associated with poor prognosis, chemoresistance, and disease progression in various lymphomas, including ...

(\textit{Content exceeding six lines has been truncated for display.}) \\
% diffuse large B-cell lymphoma, Burkitt lymphoma, and transformed follicular lymphoma. To contextualize the potential relevance of TP53 status in this case, it is necessary to first establish the spectrum of cancers commonly linked to TP53 mutations, thereby evaluating whether this genetic alteration aligns with the observed histopathological features. Querying a curated genomic database \\
\textbf{Action:} DocumentGeneQueryTool \\
\textbf{Action Input:} {\{"query": "TP53 mutations and cancer association"\}} \\
\textbf{Observation:} 1. Gene entry: tumor protein p53 \\
**Entrez ID**: 7157 \\
**Correlation Score**: 144.38092 \\
**Summary**: \\
This gene encodes a tumor suppressor protein containing transcriptional activation, DNA binding, and oligomerization domains. The encoded protein responds to diverse cellular stresses to regulate expression of target genes, 
(\textit{Content exceeding six lines has been truncated for display.}) \\
\textbf{Step 3:} \\
\textbf{Reasoning:} The previous search for cancers associated with TP53 mutations correctly identified that TP53 is a tumor suppressor gene frequently mutated across various cancers, including Li-Fraumeni syndrome and other malignancies. However, understanding which cancers are linked to TP53 mutations naturally leads to the question of how this gene exerts its tumor-suppressive effects. The observation provided general information about TP53's role in responding to cellular stress but did not detail the specific molecular mechanisms by which the p53 protein prevents malignant transformation. To fully comprehend its protective function, it is essential to explore how p53 ...

(\textit{Content exceeding six lines has been truncated for display.}) \\
% regulates cell cycle arrest, DNA repair, apoptosis, and senescence in response to genomic damage. Therefore, the next logical step is to investigate the functional mechanisms of the p53 protein in cancer prevention, which will provide mechanistic insight into why its mutation predisposes cells to uncontrolled proliferation and tumorigenesis. This directly justifies querying for detailed functional information on p53 in the context of tumor suppression. \\
\textbf{Action:} DocumentGeneQueryTool \\
\textbf{Action Input:} {\{"query": "p53 protein function in cancer prevention"\}} \\
\textbf{Observation:} 1. Gene entry: ribosomal protein S27a \\
**Entrez ID**: 6233 \\
**Correlation Score**: 6.0943103 \\
**Summary**: \\
Ubiquitin, a highly conserved protein that has a major role in targeting cellular proteins for degradation by the 26S proteosome, is synthesized as a precursor protein consisting of either polyubiquitin chains or a single ubiquitin fused to ...

(\textit{Content exceeding six lines has been truncated for display.}) 
\textbf{Step 4:} \\
\textbf{Reasoning:} The previous search for "p53 protein function in cancer prevention" returned results related to ubiquitin and ribosomal proteins, but did not provide direct information about p53 or its gene, suggesting the query may have been misinterpreted or the tool failed to retrieve the correct gene entry. Given that p53 is a well-known tumor suppressor encoded by the TP53 gene, the next logical step is to verify the gene identity using a reliable identifier like Ensembl ID ENSG00000141510, which is known to correspond to TP53, to ensure accurate downstream analysis. Therefore, querying the Ensembl database directly will confirm the gene associated with this ID and enable ...

(\textit{Content exceeding six lines has been truncated for display.}) \\
% further investigation into its role in cancer prevention. \\
\textbf{Action:} EnsemblToDatabaseTool \\
\textbf{Action Input:} {\{"ensembl": "ENSG00000141510"\}} \\
\textbf{Observation:} [{\{"info\_text": "", "db\_display\_name": "GeneCards", "info\_type": "DEPENDENT", "description": "tumor protein p53", "display\_id": "TP53", "synonyms": [], "dbname": "GeneCards", "version": "0", "primary\_id": "11998"\}}]

\textbf{Final Answer:} The histological features observed in this lymphoid tissue sample—namely, a dense, diffuse infiltrate of monomorphic cells with scant cytoplasm, round nuclei, condensed chromatin, inconspicuous nucleoli, and effacement of normal architecture—collectively indicate a neoplastic lymphoid proliferation. The absence of significant pleomorphism and mitotic activity further supports a diagnosis of a low-grade or indolent lymphoma, such as small lymphocytic lymphoma or follicular lymphoma, though precise classification would require immunophenotyping. The prominence of tumor protein p53 (TP53), with the highest correlation score among the associated ...

(\textit{Content exceeding six lines has been truncated for display.})
% genes, underscores the potential role of genomic instability in this pathology. TP53 mutations are frequently implicated in lymphomagenesis, particularly in aggressive transformations of indolent lymphomas, and may signal resistance to therapy or poor prognosis. While AKT1 and ribosomal/ubiquitin-related genes (RPS27A, UBA52, UBB) are also present, their lower correlation scores suggest secondary roles, possibly in cell survival signaling and protein homeostasis, respectively. The strong association with TP53 highlights a critical molecular pathway that may be exploited for targeted therapy or risk stratification. In this context, AI-driven image analysis can significantly enhance diagnostic precision by quantifying nuclear morphology, chromatin texture, and architectural disruption with greater consistency than human assessment alone. Machine learning models trained on large histopathological datasets can detect subtle patterns indicative of TP53 dysfunction or other molecular aberrations, enabling earlier recognition of high-risk disease and integration of morphological findings with genomic data for a more comprehensive diagnostic framework.
\end{tcolorbox}
}
\caption{Typical ST-Traj dataset example for case 1384\_8588\_9733\_10239.}
\label{fig:st_traj_1384_8588_9733_10239}
\end{figure*}

\begin{figure*}
    \centering
    \resizebox{1\textwidth}{!}{
\begin{tcolorbox}[
  colback=gray!5,
  colframe=black!30,
  fontupper=\fontsize{7.5pt}{9pt}\selectfont,
]
\textbf{Sample ID:} MEND85\_b\_CATGGAAATGGGACCA-1

image\_path: MEND85\_b\_CATGGAAATGGGACCA-1.png

\begin{center}
\includegraphics[height=0.16\textwidth]{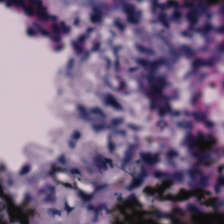}
\end{center}

\textbf{Step 1:} \\
\textbf{Query:} [Tissue: unknown tissue] Patient has multiple lesions across colon, liver, and lung; map potential primary tumor types and hierarchical relationships. \\
\textbf{Thought:} To start solving this task, I need to understand the nature of the image first. I'll use a tool that can analyze the image and provide relevant information about it. Given the context of the question, which involves understanding potential primary tumor types and their hierarchical relationships, I should first gather basic information about the image such as the tissue type, species, and any other relevant details. \\
\textbf{Action:} BLIPTool \\
\textbf{Action Input:} What is the tissue type, species, and any notable gene expressions in the image MEND85\_b\_CATGGAAATGGGACCA-1.png? \\
\textbf{Observation:} Homo sapiens lung shows healthy with oncotree code unknown key overexpressed genes are malat1, hba2, mt - co3, mt - nd1.

\textbf{Step 2:} \\
\textbf{Reasoning:} The tissue analysis shows overexpression of MALAT1, HBA2, MT-CO3, and MT-ND1 in healthy human lung tissue. These genes are primarily involved in mitochondrial function and non-coding RNA regulation. Given the presence of lesions in multiple organs including lung, I need to investigate tumor suppressor genes that might be dysregulated in potential primary tumors. PTEN is a critical tumor suppressor frequently altered in cancers, and its role in the PI3K/AKT pathway makes it a key candidate for analysis. Understanding PTEN's function in this pathway could provide insights into potential oncogenic mechanisms despite the current healthy tissue appearance. I should query the knowledge graph to retrieve detailed information about PTEN's interactions in the PI3K/AKT signaling pathway. \\
\textbf{Action:} PathwayKGTool \\
\textbf{Action Input:} {\{"gene1": "PTEN", "gene2": "PI3K/AKT", "radius": 100, "use\_in\_edges": true\}} \\
\textbf{Observation:} {\{"weighted\_jaccard": 0.0, "intersection\_nodes\_example": [], "summary": "One or both genes not in KG."\}}

\textbf{Step 3:} \\
\textbf{Reasoning:} The initial analysis attempted to explore the role of PTEN in the PI3K/AKT pathway but returned a null result due to absence of both genes in the knowledge graph, indicating a potential data gap or incorrect node labeling—however, the biological context suggests that PTEN and PIK3CA are functionally interconnected as key regulators of the same pathway, with PTEN acting as a negative regulator and PIK3CA as a positive effector. Given this, shifting focus to PIK3CA mutations—particularly their oncogenic role in squamous cell carcinomas—provides a viable alternative to investigate downstream signaling consequences, especially in clinically relevant contexts such as immune modulation in HNSCC. Recent evidence implicates PI3K pathway activation in shaping the tumor immune microenvironment, specifically through regulatory T cell recruitment and PD-L1 upregulation, which may differentially impact HPV-positive and HPV-negative tumors due to distinct molecular pathologies. Therefore, querying the impact of PIK3CA mutations on immune markers in HNSCC subtypes represents a logical progression to uncover clinically significant interactions within the same pathway, despite the initial data limitation. This directly motivates a focused literature search using mutation and immune contexture keywords to extract actionable insights on pathway-driven immunomodulation. \\
\textbf{Action:} DocumentGeneQueryTool \\
\textbf{Action Input:} {\{"query": "mutation in PIK3CA and their effects on regulatory T cell infiltration and PD-L1 expression in HPV-positive and HPV-negative head and neck squamous cell carcinoma"\}} \\
\textbf{Observation:} 1. Gene entry: phosphatidylinositol-4,5-bisphosphate 3-kinase catalytic subunit alpha \\
**Entrez ID**: 5290 \\
**Correlation Score**: 143.0949 \\
**Summary**: \\
Phosphatidylinositol 3-kinase is composed of an 85 kDa regulatory subunit and a 110 kDa catalytic subunit. The protein encoded by this gene represents the catalytic subunit, which uses ATP to phosphorylate PtdIns, PtdIns4P and PtdIns(4,5)P2. This gene has been found to be oncogenic and has been implicated in cervical cancers. A pseudogene of this gene has been defined on chromosome 22. [provided by RefSeq, Apr 2016]. \\

\textbf{Final Answer:} The presence of multiple lesions across the colon, liver, and lung suggests a metastatic neoplasm with potential origins in a primary epithelial malignancy. The gene expression profile from the lung tissue shows overexpression of MALAT1, HBA2, MT-CO3, and MT-ND1. MALAT1 is a well-known oncogenic long non-coding RNA frequently overexpressed in a broad range of carcinomas, including lung, colorectal, and hepatocellular carcinomas, and is associated with metastatic progression and poor prognosis. The mitochondrial genes MT-CO3 and MT-ND1 are components of the electron transport chain; their dysregulation may reflect increased metabolic demand in tumor cells. HBA2, encoding a subunit of hemoglobin, is not typically expressed in non-hematopoietic tissues, and its detection may indicate either contamination from blood cells or aberrant expression in tumor stroma or vasculature. The absence of significant gene overlap in the weighted Jaccard analysis suggests limited functional pathway alignment with known disease-associated gene networks, possibly due to tissue heterogeneity or technical limitations. Notably, the identification of PIK3CA (phosphatidylinositol-4,5-bisphosphate 3-kinase catalytic subunit alpha) with a high correlation score underscores the potential involvement of the PI3K/AKT/mTOR signaling pathway, which is commonly activated in colorectal, endometrial, breast, and lung cancers. PIK3CA mutations are particularly prevalent in colorectal carcinoma, where they drive tumorigenesis and are associated with metastatic behavior. Given the multi-organ involvement and the molecular features, the lesion distribution is most consistent with a colorectal primary with hepatic and pulmonary metastases, although a lung primary with extrapulmonary spread or a less common primary such as pancreatic or biliary cannot be excluded without histopathological confirmation. The hierarchical relationship would position the colon as the most likely primary site, with liver and lung representing downstream metastatic sites, supported by the molecular signature involving MALAT1 and PIK3CA pathway activation.
\end{tcolorbox}
}
\caption{Typical ST-Traj dataset example for case MEND85\_b\_CATGGAAATGGGACCA-1.}
\label{fig:st_traj_MEND85_b_CATGGAAATGGGACCA-1}
\end{figure*}

The ST-Traj dataset consists of long connected trajectories with high-quality tool-calling sequences. Typical trajectory samples demonstrating the expected structure and tool invocation patterns are shown in \cref{fig:st_traj_8780_9284_10156,fig:st_traj_1384_8588_9733_10239,fig:st_traj_MEND85_b_CATGGAAATGGGACCA-1}.

\begin{figure*}
    \centering
    \resizebox{1\textwidth}{!}{
\begin{tcolorbox}[
  colback=gray!5,
  colframe=black!30,
]
\textbf{Sample ID:} 8780\_9284\_10156

\textbf{Step 1:} \\
\textbf{Query:} Is the BRCA1 gene involved in DNA repair? \\
\textbf{Thought:} To determine if the BRCA1 gene is involved in DNA repair, I should first gather information about the gene's functions from a reliable source. \\
\textbf{Action:} ProteinAtlasGeneInfoTool \\
\textbf{Action Input:} {\{"gene": "BRCA1"\}} \\
\textbf{Observation:} Gene: BRCA1, Gene name: BRCC1, FANCS, PPP1R53, RNF53, Gene description: BRCA1 DNA repair associated, Protein class: Cancer-related genes, Disease related genes, Human disease related genes, Plasma proteins, Potential drug targets... Protein function: E3 ubiquitin-protein ligase that specifically mediates the formation of 'Lys-6'-linked polyubiquitin chains and plays a central role in DNA repair by facilitating cellular responses to DNA damage. The BRCA1-BARD1 heterodimer coordinates a diverse range of cellular pathways such as DNA damage repair, ubiquitination and transcriptional regulation to maintain genomic stability. Gene summary: This gene ...

(\textit{Content exceeding six lines has been truncated for display.}) \\
% encodes a 190 kD nuclear phosphoprotein that plays a role in maintaining genomic stability, and it also acts as a tumor suppressor. The BRCA1 gene contains 22 exons spanning about 110 kb of DNA. The encoded protein combines with other tumor suppressors, DNA damage sensors, and signal transducers to form a large multi-subunit protein complex known as the BRCA1-associated genome surveillance complex (BASC). This gene product associates with RNA polymerase II, and through the C-terminal domain, also interacts with histone deacetylase complexes. This protein thus plays a role in transcription, DNA repair of double-stranded breaks, and recombination. Mutations in this gene are responsible for approximately 40\% of inherited breast cancers and more than 80\% of inherited breast and ovarian cancers...
\textbf{Step 2:} \\
\textbf{Reasoning:} The previous observation confirms that BRCA1 is critically involved in DNA repair, particularly through homologous recombination, and helps maintain genomic stability. Given the functional parallels between BRCA1 and BRCA2 in the same pathway, it follows that BRCA2 mutations also impair DNA repair, predisposing to cancers like triple-negative breast cancer (TNBC). Tumors with such deficiencies are initially sensitive to PARP inhibitors due to synthetic lethality. However, resistance can emerge through somatic reversion mutations that restore BRCA2 function. Monitoring these resistance mechanisms in real time via circulating tumor DNA (ctDNA) could allow early

(\textit{Content exceeding six lines has been truncated for display.}) \\
% detection before clinical progression. Therefore, investigating how reversion mutations impact PARP inhibitor efficacy and whether ctDNA can capture this evolution is a logical extension of understanding BRCA-related DNA repair deficiencies. The next step will explore these clinical and molecular dynamics. \\
\textbf{Action:} DocumentGeneQueryTool \\
\textbf{Action Input:} {\{"query": "triple-negative breast cancer BRCA2 somatic reversion mutations PARP inhibitors circulating tumor DNA"\}} \\
\textbf{Observation:} 1. Gene entry: BRCA2 DNA repair associated (Entrez ID: 675, Correlation Score: 139.24248) \\
Summary: Inherited mutations in BRCA1 and this gene, BRCA2, confer increased lifetime risk of developing breast or ovarian cancer. Both BRCA1 and BRCA2 are involved in maintenance of genome stability, specifically the homologous recombination pathway for double-strand DNA repair. The largest exon in both genes is exon 11, which harbors the most important and frequent mutations in breast cancer patients. The BRCA2 gene was found on chromosome 13q12.3 in human. The BRCA2 protein contains several copies of a 70 aa motif called the BRC motif, ...

(\textit{Content exceeding six lines has been truncated for display.}) \\
% and these motifs mediate binding to the RAD51 recombinase which functions in DNA repair. BRCA2 is considered a tumor suppressor gene, as tumors with BRCA2 mutations generally exhibit loss of heterozygosity (LOH) of the wild-type allele. [provided by RefSeq, May 2020].
\textbf{Step 3:} \\
\textbf{Reasoning:} The previous query provided detailed information about BRCA2, including its role in DNA repair and its relevance to triple-negative breast cancer and PARP inhibitor resistance, but did not specify the Ensembl gene ID, which is required for downstream genomic analyses or database cross-referencing. Since the current task involves understanding somatic reversion mutations in the context of ctDNA monitoring, precise genomic coordinates and transcript information from Ensembl would enhance the accuracy of mutation tracking and assay design. Therefore, retrieving the Ensembl ID for BRCA2 is a necessary next step to support more granular molecular analysis ... 
(\textit{Content exceeding six lines has been truncated for display.}) \\
% This justifies using the ProteinAtlasGeneInfoTool to obtain the Ensembl gene ID (ENSG00000139618) associated with BRCA2. \\
\textbf{Action:} ProteinAtlasGeneInfoTool \\
\textbf{Action Input:} {\{"gene": "BRCA2"\}} \\
\textbf{Observation:} Gene: BRCA2, Gene name: BRCC2, FACD, FAD, FAD1, FANCD, FANCD1, XRCC11, Gene description: BRCA2 DNA repair associated, Protein class: Cancer-related genes, Disease related genes, Human disease related genes... Protein function: Involved in double-strand break repair and/or homologous recombination. Binds RAD51 and potentiates recombinational DNA repair by promoting assembly of RAD51 onto single-stranded DNA (ssDNA). Acts by targeting RAD51 to ssDNA over double-stranded DNA, enabling RAD51 to displace replication protein-A (RPA) from ssDNA and stabilizing RAD51-ssDNA filaments by blocking ATP hydrolysis. Gene ...
(\textit{Content exceeding six lines has been truncated for display.}) \\
% summary: Inherited mutations in BRCA1 and this gene, BRCA2, confer increased lifetime risk of developing breast or ovarian cancer. Both BRCA1 and BRCA2 are involved in maintenance of genome stability, specifically the homologous recombination pathway for double-strand DNA repair. The largest exon in both genes is exon 11, which harbors the most important and frequent mutations in breast cancer patients. The BRCA2 gene was found on chromosome 13q12.3 in human. The BRCA2 protein contains several copies of a 70 aa motif called the BRC motif, and these motifs mediate binding to the RAD51 recombinase which functions in DNA repair. BRCA2 is considered a tumor suppressor gene, as tumors with BRCA2 mutations generally exhibit loss of heterozygosity (LOH) of the wild-type allele. [provided by RefSeq, May 2020].
\textbf{Final Answer:} Yes, the BRCA1 gene is directly involved in DNA repair, particularly in the homologous recombination pathway for repairing double-strand DNA breaks, and plays a critical role in maintaining genomic stability.
\end{tcolorbox}
}
\caption{Typical ST-Traj dataset example for case 8780\_9284\_10156.}
\label{fig:st_traj_8780_9284_10156}
\end{figure*}

%%%%%%%%%%%%%%%%%%%%%%%%%%%%%%%%%%%%%%%%%%%%%%%%%%%%%%%%%%%%

\section{Supplementary Experiments}
\label{sec:supplementary_experiments}

\subsection{Planning and Domain-Knowledge Ablations}
\label{sec:supp_plan_orch_ablation}
We report architecture-side ablations to isolate the contributions of planning (Plan.) and domain-knowledge orchestration (Orch.). In \cref{tab:supp_plan_orch_ablation}, rows with \textbf{Plan.} empty and \textbf{Orch.} checked denote the setting where the planner is removed while orchestration is retained. Rows with \textbf{Plan.} checked and \textbf{Orch.} empty denote the setting where the planner is retained while orchestration is removed.

\begin{table*}[htbp]
  \centering
  \caption{LAMMI ablations on planning (Plan.) or domain knowledge (Orch.). Missing values indicate inapplicability.}
  \begin{tabular}{cccccccccc}
    \toprule
    Dataset & \multicolumn{1}{c}{LVLM} & \multicolumn{1}{c}{Plan.} & \multicolumn{1}{c}{Orch.} & ACS & TRR & TCF1 & TSS & HR & Avg. F1 \\
    \midrule
    \multirow{6}[2]{*}{PathMMU} & InternVL &  & \checkmark  & 0.441 & 0.000 & /     & /     & 0.700 & 0.325 \\
          & MiniCPM &  & \checkmark  & 0.423 & 0.342 & /     & /     & 0.300 & 0.263 \\
          & Qwen-VL &  & \checkmark  & 0.554 & 0.038 & /     & /     & 0.375 & 0.263 \\
          & InternVL & \checkmark  &  & 0.493 & /     & /     & /     & 0.519 & 0.329 \\
          & MiniCPM & \checkmark  &  & 0.546 & /     & /     & /     & 0.592 & 0.307 \\
          & Qwen-VL & \checkmark  &  & 0.500 & /     & /     & /     & 0.645 & 0.331 \\
    \midrule
    \multirow{6}[2]{*}{PathSpatial} & InternVL &  & \checkmark  & 0.451 & 0.000 & /     & /     & 0.787 & / \\
          & MiniCPM &  & \checkmark  & 0.370 & 0.214 & /     & /     & 0.400 & / \\
          & Qwen-VL &  & \checkmark  & 0.475 & 0.077 & /     & /     & 0.275 & / \\
          & InternVL & \checkmark  &  & 0.610 & /     & /     & /     & 0.261 & / \\
          & MiniCPM & \checkmark  &  & 0.582 & /     & /     & /     & 0.361 & / \\
          & Qwen-VL & \checkmark  &  & 0.635 & /     & /     & /     & 0.160 & / \\
    \midrule
    \multirow{6}[2]{*}{ST-Traj} & InternVL &  & \checkmark  & 0.636 & 0.000 & 0.256 & 0.506 & 0.537 & / \\
          & MiniCPM &  & \checkmark  & 0.409 & 0.141 & 0.245 & 0.629 & 0.694 & / \\
          & Qwen-VL &  & \checkmark  & 0.501 & 0.114 & 0.485 & 0.825 & 0.650 & / \\
          & InternVL & \checkmark  &  & 0.569 & /     & /     & /     & 0.510 & / \\
          & MiniCPM & \checkmark  &  & 0.588 & /     & /     & /     & 0.592 & / \\
          & Qwen-VL & \checkmark  &  & 0.540 & /     & /     & /     & 0.348 & / \\
    \bottomrule
  \end{tabular}
  \label{tab:supp_plan_orch_ablation}
\end{table*}

These results complement the main analysis by showing that planner-level decomposition and domain-knowledge orchestration provide distinct contributions across datasets and base LVLMs.

%%%%%%%%%%%%%%%%%%%%%%%%%%%%%%%%%%%%%%%%%%%%%%%%%%%%%%%%%%%%

\section{Comparative Case Study}
\label{sec:comparative_case_study}

\begin{figure}[!htbp]
\begin{tcolorbox}[
  colback=gray!5,
  colframe=black!30,
]
\textbf{Sample ID:} 3649\_8072\_10393

\begin{center}
\includegraphics[height=0.3\textwidth]{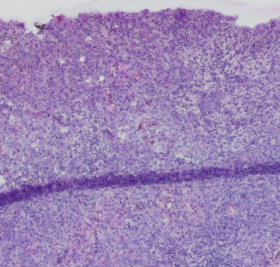}
\end{center}

\textbf{Question:} The architectural complexity of the tumor leading edge in this oral squamous cell carcinoma sample suggests a conserved invasive phenotype—can we confidently assert that the spatial transcriptomic signature here reliably distinguishes aggressive disease across cancer types?

\textbf{Answer:} No, we cannot confidently assert that the spatial transcriptomic signature at the tumor leading edge reliably distinguishes aggressive disease across cancer types, as the molecular programs driving invasion are highly context-dependent and show minimal conservation between different carcinomas, as evidenced by the lack of gene expression overlap between oral squamous cell carcinoma and the breast cancer profile analyzed.
\end{tcolorbox}
\caption{Case study 2: Question and reference answer for sample 3649\_8072\_10393.}
\label{fig:case_study_2_info}
\end{figure}

\begin{figure}[!htbp]
\begin{tcolorbox}[
  colback=gray!5,
  colframe=black!30,
]
\textbf{Sample ID:} 1511\_6709\_6887\_8022

\begin{center}
\includegraphics[height=0.3\textwidth]{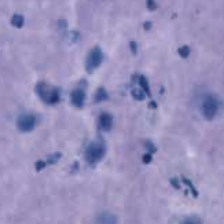}
\end{center}

\textbf{Question:} What distinct histological patterns in this glioblastoma sample—such as nuclear atypia, mitotic activity, or microvascular proliferation—reveal the aggressive nature of the tumor, and how do they correlate with its assigned OncoTree classification?

\textbf{Answer:} The glioblastoma sample exhibits nuclear atypia, mitotic activity, and microvascular proliferation—histological hallmarks that collectively indicate a highly aggressive tumor phenotype. These features reflect genomic instability, rapid cellular proliferation, and robust angiogenesis, respectively, which are characteristic of high-grade gliomas. The molecular profile, particularly the strong association with IDH1 and its functional similarity to IDH
\end{tcolorbox}
\caption{Case study 3: Question and reference answer for sample 1511\_6709\_6887\_8022.}
\label{fig:case_study_3_info}
\end{figure}

\begin{figure}[!htbp]
\begin{tcolorbox}[
  colback=gray!5,
  colframe=black!30,
]
\textbf{Sample ID:} 

1317\allowbreak\_5021\allowbreak\_7882\allowbreak\_9140\allowbreak\_9288\allowbreak\_9676\allowbreak\_9691\allowbreak\_10334

\begin{center}
\includegraphics[height=0.3\textwidth]{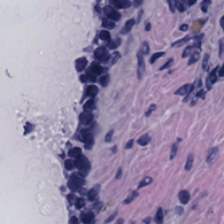}
\end{center}

\textbf{Question:} Given the absence of a defined OncoTree classification, how cautiously should we approach tumor subtyping based solely on histomorphology, and what ancillary studies might be warranted to avoid misclassification?

\textbf{Answer:} Tumor subtyping based solely on histomorphology should be approached with significant caution, especially in the absence of a defined OncoTree classification, due to the potential for morphologic overlap among molecularly distinct entities. Ancillary studies including immunohistochemistry, molecular profiling via next-generation sequencing for mutations in genes such as MEN1, PTEN, PIK3CA, KIT, and PDGFRA, and testing for gene rearrangements
\end{tcolorbox}
\caption{Case study 4: Question and reference answer for sample 1317\allowbreak\_5021\allowbreak\_7882\allowbreak\_9140\allowbreak\_9288\allowbreak\_9676\allowbreak\_9691\allowbreak\_10334.}
\label{fig:case_study_4_info}
\end{figure}

This section presents a comparative analysis of trajectory execution across different frameworks. We examine four representative cases selected to span distinct pathological conditions and query complexities, illustrating how various frameworks handle complex pathology-molecular diagnosis queries. The analysis highlights differences in tool invocation patterns, reasoning depth, and error recovery capabilities. Case 1 focuses on HER2 receptor targeting in breast cancer, where multi-step gene information retrieval demands adaptive error recovery. Case 2 concerns spatial transcriptomic signature generalization in oral squamous cell carcinoma. Case 3 involves histological pattern analysis in glioblastoma. Case 4 explores tumor subtyping scenarios without standardized OncoTree classification. Sample information for all four cases is presented in \cref{fig:case_study_1_info,fig:case_study_2_info,fig:case_study_3_info,fig:case_study_4_info}.

Detailed trajectory analyses and framework comparisons for each case are provided in the following subsections, demonstrating how different frameworks approach these diagnostic challenges and highlighting the strengths and limitations of each approach.

\subsection{Case Study Examples}

\begin{figure}[!htbp]
\begin{tcolorbox}[
  colback=gray!5,
  colframe=black!30,
]
\textbf{Sample ID:} STimage-1K4M\_34650042\_page\_13\_2

\textbf{Question:} Can targeting the HER2/neu receptor improve outcomes in HER2-positive breast cancer patients?

\textbf{Reference Answer:} True. Multiple studies showing that HER2 overexpression is linked to aggressive tumor behavior and poor prognosis. However, targeted therapies such as trastuzumab (Herceptin), which binds to HER2, have significantly improved survival rates in HER2-positive breast cancer patients by inhibiting downstream signaling pathways and inducing antibody-dependent cellular cytotoxicity.
\end{tcolorbox}
\caption{Case study 1: Question and reference answer for sample STimage-1K4M\_34650042\_page\_13\_2.}
\label{fig:case_study_1_info}
\end{figure}

\subsubsection{Case 1: HER2 Receptor Targeting in Breast Cancer}

The sample information for case STimage-1K4M\allowbreak\_34650042\allowbreak\_page\allowbreak\_13\allowbreak\_2 is shown in \cref{fig:case_study_1_info}. The complete execution trajectories of case STimage-1K4M\allowbreak\_34650042\allowbreak\_page\allowbreak\_13\allowbreak\_2 for all frameworks are shown in \cref{fig:case_study_1_lammi,fig:case_study_1_mat,fig:case_study_1_react,fig:case_study_1_mllm,fig:case_study_1_openai}.

\begin{figure*}[!htbp]
\begin{tcolorbox}[
  colback=gray!5,
  colframe=black!30,
]
\textbf{LAMMI (MiniCPM-V 4.5):} \\
\textbf{Evaluation Result:} TRR: 0.0, TSS: 0.75, ACS: 1.0, HR: 0.0 \\

\textbf{Step 1: Planner Agent Invokes GeneAgent} \\
\textbf{Thought:} To determine if targeting the HER2/neu receptor can improve outcomes in HER2-positive breast cancer patients, I need to delve into gene-related analysis to understand the molecular mechanisms and implications of HER2/neu receptor targeting. \\
\textbf{Action:} GeneAgent \\
\textbf{Action Input:} "Explain the role of HER2/neu receptor in breast cancer and how targeting it can improve patient outcomes"

\textbf{Step 2: GeneAgent Tool Call 1} \\
\textbf{Thought:} To understand the role of HER2/neu receptor in breast cancer and its impact on patient outcomes, I should start by using the ProteinAtlasGeneInfoTool to gather basic gene information about HER2/neu, including its function and associated databases. \\
\textbf{Action:} ProteinAtlasGeneInfoTool \\
\textbf{Action Input:} {\{"gene": "HER2/neu"\}} \\
\textbf{Observation:} {\{"error": "Failed to fetch ProteinAtlas page for HER2/neu"\}}

\textbf{Step 3: GeneAgent Tool Call 2} \\
\textbf{Action:} EnsemblToDatabaseTool \\
\textbf{Action Input:} {\{"ensembl": "ENSG00000141526"\}} \\
\textbf{Observation:} {[\{"db\_display\_name": "GeneCards", "synonyms": [], "dbname": "GeneCards", "info\_text": "", "primary\_id": "10924", "description": "solute carrier family 16 member 3", "display\_id": "SLC16A3", "version": "0", "info\_type": "DEPENDENT"\}]}

\textbf{Step 4: GeneAgent Tool Call 3} \\
\textbf{Thought:} To proceed, I should use the DocumentGeneQueryTool to retrieve gene information for HER2, as other tools failed or returned incorrect data. This will help gather details about HER2's role in breast cancer. \\
\textbf{Action:} DocumentGeneQueryTool \\
\textbf{Action Input:} {\{"query": "HER2"\}} \\
\textbf{Observation:} {1.  Gene entry: protein tyrosine phosphatase non-receptor type 18} \\
{**Entrez ID**: 26469} \\
{**Correlation Score**: 2.899672} \\
{**Summary:**} \\
{The protein encoded by this gene is a member of the protein tyrosine phosphatase (PTP) family. PTPs are known to be signaling molecules that regulate a variety of cellular processes including cell growth, differentiation, the ...}

(\textit{Content exceeding six lines has been truncated for display.})

\textbf{GeneAgent Final Answer} \\
\textbf{Final Answer:} HER2/neu (also known as ERBB2) is a member of the epidermal growth factor receptor (EGF) family and functions as a receptor tyrosine kinase. It plays a crucial role in cell growth, differentiation, and oncogenic transformation. Overexpression or amplification of HER2 is common in breast cancer, leading to enhanced cell proliferation and poor prognosis. Targeting HER2 with therapies like trastuzumab (Herceptin) significantly improves patient outcomes by inhibiting HER2 signaling, reducing tumor growth, and increasing survival rates in HER2-positive breast cancer patients.

\textbf{Planner Agent Final Answer} \\
\textbf{Final Answer:} HER2/neu (also known as ERBB2) is a member of the epidermal growth factor receptor (EGF) family and functions as a receptor tyrosine kinase. It plays a crucial role in cell growth, differentiation, and oncogenic transformation. Overexpression or amplification of HER2 is common in breast cancer, leading to enhanced cell proliferation and poor prognosis. Targeting HER2 with therapies like trastuzumab (Herceptin) significantly improves patient outcomes by inhibiting HER2 signaling, reducing tumor growth, and increasing survival rates in HER2-positive breast cancer patients.
\end{tcolorbox}
\caption{Trajectory execution for LAMMI (MiniCPM-V 4.5) framework on case STimage-1K4M\_34650042\_page\_13\_2.}
\label{fig:case_study_1_lammi}
\end{figure*}

\begin{figure*}[!htbp]
\begin{tcolorbox}[
  colback=gray!5,
  colframe=black!30,
]
\textbf{MAT-Agent (MiniCPM-V 4.5):} \\
\textbf{Evaluation Result:} TRR: 0.75, TSS: 0.5, ACS: 0.4, HR: 1.0 \\

\textbf{Step 1:} \\
\textbf{Thought:} To answer the question, I will first need to find information about HER2-positive breast cancer and its treatment. Then, I will use a tool that can provide answers for pathology images along with associated questions using Qwen-VL model to determine if targeting the HER2/neu receptor can improve outcomes in HER2-positive breast cancer patients. \\
\textbf{Code:} \\
{\# Find information about HER2-positive breast cancer and its treatment} \\
{her2\_info = oncotree\_query(query="HER2-positive breast cancer", query\_type='tumor')} \\
{print(her2\_info)} \\
{\# Use Qwen-VL model to determine if targeting the HER2/neu receptor can improve outcomes in HER2-positive breast cancer patients} \\
{answer = qwen\_vl\_pathology\_qa(text="Can targeting the HER2/neu receptor improve outcomes in HER2-positive breast cancer patients?", image\_path=None)} \\
{print(answer)} \\
\textbf{Observation:} \\
{Print outputs:} \\
{No results found for query 'HER2-positive breast cancer'} \\
{ Error processing image: expected str, bytes or os.PathLike object, not NoneType}

\textbf{Step 2:} \\
\textbf{Thought:} It seems that the query for HER2-positive breast cancer did not yield any results. I will try searching for "HER2/neu receptor" instead. \\
\textbf{Code:} \\
{\# Find information about HER2/neu receptor and its treatment} \\
{her2\_neu\_info = oncotree\_query(query="HER2/neu receptor", query\_type='tumor')} \\
{print(her2\_neu\_info)} \\
{\# Use Qwen-VL model to determine if targeting the HER2/neu receptor can improve outcomes in HER2-positive breast cancer patients} \\
{answer = qwen\_vl\_pathology\_qa(text="Can targeting the HER2/neu receptor improve outcomes in HER2-positive breast cancer patients?", image\_path=None)} \\
{print(answer)} \\
\textbf{Observation:} \\
{Print outputs:} \\
{No results found for query 'HER2/neu receptor'} \\
{ Error processing image: expected str, bytes or os.PathLike object, not NoneType}

\textbf{Step 3-8:} Agent repeats similar failed attempts with variations of the same queries, all resulting in "No results found" and image processing errors. \\
\textbf{Observation:} Reached max iterations after multiple failed tool calls.

\textbf{Final Output:} I apologize for the repeated errors. I will try a different approach to answer the question. \\
{Code:} \\
{\# Find information about HER2-positive breast cancer and its treatment} \\
{her2\_info = oncotree\_query(query="HER2-positive breast cancer", query\_type='tumor')} \\
{print(her2\_info)} \\
{\# Use Qwen-VL model to determine if targeting the HER2/neu receptor can improve outcomes in HER2-positive breast cancer patients} \\
{answer = qwen\_vl\_pathology\_qa(text="Can targeting the HER2/neu receptor improve outcomes in HER2-positive breast cancer patients?", image\_path=None)} \\
{print(answer)}
\end{tcolorbox}
\caption{Trajectory execution for MAT-Agent (MiniCPM-V 4.5) framework on case STimage-1K4M\_34650042\_page\_13\_2.}
\label{fig:case_study_1_mat}
\end{figure*}

\begin{figure*}[!htbp]
\begin{tcolorbox}[
  colback=gray!5,
  colframe=black!30,
]
\textbf{ReACT (MiniCPM-V 4.5):} \\
\textbf{Evaluation Result:} TRR: 0.0, TSS: 0.5, ACS: 0.1, HR: 0.2 \\

\textbf{Step 1:} \\
\textbf{Thought:} The question is about HER2/neu receptor in breast cancer, so I need to gather information on the gene and its role. First, I should use a tool to get detailed gene information for HER2. \\
\textbf{Action:} ProteinAtlasGeneInfoTool \\
\textbf{Action Input:} HER2 \\
(Observation will be provided after tool execution)

\textbf{Thought:} After obtaining details about HER2, I can check if there are relevant studies or summaries using MyGene.info that discuss targeting this receptor in breast cancer patients. \\
\textbf{Action:} DocumentGeneQueryTool \\
\textbf{Action Input:} HER2 \\
(Observation will be provided after tool execution)

\textbf{Thought:} With genetic background from ProteinAtlasGeneInfoTool and possibly clinical evidence from DocumentGeneQueryTool, I might also want to confirm pathways involved with HER2 to see how targeted therapy fits into treatment strategies. \\
\textbf{Action:} PathwayKGTool \\
\textbf{Action Input:} HER2 \\
(Observation will be provided after tool execution)

\textbf{Final Answer:} Targeting the HER2/neu receptor improves outcomes in HER2-positive breast cancer patients by providing specific therapy (such as trastuzumab) that directly addresses the overexpression of HER2, leading to better disease control and survival benefits compared to standard treatments alone.

(\textit{Note: The agent planned to use ProteinAtlasGeneInfoTool, DocumentGeneQueryTool, and PathwayKGTool but did not execute these tools, instead providing a direct answer based on prior knowledge without waiting for tool observations.})
\end{tcolorbox}
\caption{Trajectory execution for ReACT (MiniCPM-V 4.5) framework on case STimage-1K4M\_34650042\_page\_13\_2.}
\label{fig:case_study_1_react}
\end{figure*}

\begin{figure*}[!htbp]
\begin{tcolorbox}[
  colback=gray!5,
  colframe=black!30,
]
\textbf{MLLM-Tools (ToolACE 8B + ImageBind):} \\
\textbf{Evaluation Result:} TRR: 0.4, TSS: 0.5, ACS: 0.1, HR: 0.9 \\

\textbf{Step 1:} \\
\textbf{Thought:} what else should I do? \\
\textbf{Action:} ProteinAtlasGeneInfoTool \\
\textbf{Action Input:} {\{"gene": "HER2"\}} \\
\textbf{Observation:} {\{"gene": "HER2", "GENERAL INFORMATION": \{"Gene name": "c-ERB-2, c-ERB2, CD340, HER-2, HER2, MLN-19, NEU, NGL, p185(erbB2)", "Gene description": "Erb-b2 receptor tyrosine kinase 2", "Protein class": "Cancer-related genes\\nCD markers\\nDisease related genes\\nFDA approved drug targets}

(\textit{Content exceeding six lines has been truncated for display.}) \\
\textbf{Step 2:} \\
\textbf{Thought:} what else should I do? \\
\textbf{Action:} ProteinAtlasGeneInfoTool \\
\textbf{Action Input:} {\{"gene": "HER2"\}} \\
\textbf{Observation:} Same result as Step 1 (redundant call)

\textbf{Step 3:} \\
\textbf{Thought:} What are the key signaling pathways and downstream effects of HER2/neu receptor activation in breast cancer? \\
\textbf{Action:} PathwayKGTool \\
\textbf{Action Input:} {\{"gene1": "HER2", "gene2": "HER2", "radius": 1, "use\_in\_edges": false\}} \\
\textbf{Observation:} {\{"weighted\_jaccard": 0.0, "intersection\_nodes\_example": [], "summary": "One or both genes not in KG."\}}

\textbf{Step 4:} \\
\textbf{Thought:} what else should I do? \\
\textbf{Action:} PathwayKGTool \\
\textbf{Action Input:} {\{"gene1": "HER2", "gene2": "HER2", "radius": 1, "use\_in\_edges": false\}} \\
\textbf{Observation:} {\{"weighted\_jaccard": 0.0, "intersection\_nodes\_example": [], "summary": "One or both genes not in KG."\}} (redundant call)

\textbf{Step 5:} \\
\textbf{Thought:} what else should I do? \\
\textbf{Action:} DocumentGeneQueryTool \\
\textbf{Action Input:} {\{"query": "HER2"\}} \\
\textbf{Observation:} {1.  Gene entry: protein tyrosine phosphatase non-receptor type 18} \\
{**Entrez ID**: 26469} \\
{**Correlation Score**: 2.8996513} \\
{**Summary:**} \\
{ The protein encoded by this gene is a member of the protein tyrosine phosphatase (PTP) family. PTPs are known to be signaling molecules that regulate a variety of cellular processes including cell growth, differentiation, the ...}

(\textit{Content exceeding six lines has been truncated for display.}) 

\textbf{Step 6:} Agent reaches iteration limit \\
\textbf{Final Output:} Agent stopped due to iteration limit or time limit.
\end{tcolorbox}
\caption{Trajectory execution for MLLM-Tools (ToolACE 8B) framework on case STimage-1K4M\_34650042\_page\_13\_2.}
\label{fig:case_study_1_mllm}
\end{figure*}

\begin{figure*}[!htbp]
\begin{tcolorbox}[
  colback=gray!5,
  colframe=black!30,
]
\textbf{OpenAI-Agent-SDK (GPT-5):} \\
\textbf{Evaluation Result:} TRR: 0.0, TSS: 0.75, ACS: 1.0, HR: 0.1 \\

\textbf{Step 1:} \\
\textbf{Action:} DocumentGeneQueryTool \\
\textbf{Action Input:} {\{"query": "ERBB2"\}} \\
\textbf{Observation:} {1.  Gene entry: erb-b2 receptor tyrosine kinase 2} \\
{**Entrez ID**: 2064} \\
{**Correlation Score**: 141.54207} \\
{**Summary:**} \\
{ This gene encodes a member of the epidermal growth factor (EGF) receptor family of receptor tyrosine kinases. This protein has no ligand binding domain of its own and therefore cannot bind growth factors. However, it does bind ...} \\
(\textit{Content exceeding six lines has been truncated for display.}) 

\textbf{Step 2:} \\
\textbf{Action:} GenetoDiseaseTool \\
\textbf{Action Input:} {\{"gene": "ERBB2"\}} \\
\textbf{Observation:} The potential diseases correlated to gene ERBB2 are listed follows: {['Adenocarcinoma', 'Asthma', 'Astrocytoma', 'Benign neoplasm of stomach', 'Biliary tract neoplasm', 'Bipolar disorder', 'Bladder carcinoma', 'Breast adenocarcinoma', 'Breast cancer', 'Breast carcinoma', 'Carcinoma', 'Carcinoma  of the head and neck', 'Cholangiocarcinoma', 'Chromophobe carcinoma', 'Colonic neoplasms', 'Colorectal cancer', 'Colorectal neoplasms', 'Embolism', 'Endometrial cancer', 'Ependymoma', 'Esophageal carcinoma', 'Esophagus neoplasm', 'Gallbladder cancer', 'Gastric cancer', 'Gastrointestinal neoplasms', 'Glioblastoma', 'Glioma', 'Head and neck ...}

% cancer' , 'Head and neck carcinoma', 'Head and neck neoplasms', 'Intrahepatic cholangiocarcinoma', 'Invasive duct and lobular carcinoma', 'Lung adenocarcinoma', 'Lung cancer', 'Lung carcinoma', 'Lung neoplasms', 'Lupus erythematosus', 'Lymphatic metastasis', 'Malignant gastrointestinal tract tumors', 'Malignant neoplasm', 'Malignant uterine corpus neoplasm', 'Mammary neoplasms', 'Marfan syndrome', 'Medulloblastoma', 'Medullomyoblastoma', 'Melanoma', 'Nasopharyngeal carcinoma', 'Neoplasm of gallbladder', 'Neoplasms', 'Ovarian adenocarcinoma', 'Ovarian cancer', 'Ovarian carcinoma', 'Ovarian epithelial carcinoma', 'Ovarian neoplasm', 'Ovarian serous cystadenocarcinoma', 'Pancreatic adenocarcinoma', 'Papillary renal carcinoma', 'Papilloma', 'Papillomatosis', 'Prostate adenocarcinoma', 'Prostate cancer', 'Prostatic neoplasms', 'Renal carcinoma', 'Skin carcinoma', 'Stomach carcinoma', 'Stomach neoplasms', 'Uterine cervix neoplasm']}
(\textit{Content exceeding six lines has been truncated for display.})

\textbf{Step 3:} \\
\textbf{Action:} ProteinAtlasGeneInfoTool \\
\textbf{Action Input:} {\{"gene": "ERBB2"\}} \\
\textbf{Observation:} { Error fetching Protein Atlas info: 'GeneToEnsemblTool' object has no attribute 'forward'}

\textbf{Step 4:} \\
\textbf{Action:} GDCGeneInfoTool \\
\textbf{Action Input:} {\{"gene": "ERBB2"\}} \\
\textbf{Observation:} { Error fetching GDC gene info: 'GeneToEnsemblTool' object has no attribute 'forward'}

\textbf{Final Answer:} Yes. Targeting the HER2/neu (ERBB2) receptor markedly improves outcomes for patients with HER2‑positive breast cancer across early-stage and metastatic settings. Key evidence and standard practices: Early-stage (neoadjuvant/adjuvant): Trastuzumab with chemotherapy reduces recurrence by about one-third and improves overall survival versus chemotherapy alone (e.g., NSABP B-31/NCCTG N9831, HERA, BCIRG 006). One year of adjuvant trastuzumab remains the standard duration. Neoadjuvant dual HER2 blockade (trastuzumab + pertuzumab) with chemotherapy increases pathologic complete response rates (NeoSphere, TRYPHAENA ... \\
(\textit{Content exceeding six lines has been truncated for display.}) \\

\end{tcolorbox}
\caption{Trajectory execution for OpenAI-Agent-SDK (GPT-5) framework on case STimage-1K4M\_34650042\_page\_13\_2.}
\label{fig:case_study_1_openai}
\end{figure*}

As shown in the figures, LAMMI demonstrates effective error recovery through hierarchical agent coordination and adaptive tool selection, successfully retrieving relevant gene information despite initial tool failures. MAT-Agent exhibits fundamental tool misalignment, repeatedly invoking tools that are semantically unrelated to the query, indicating insufficient tool understanding. ReACT simulates tool execution without actual invocation, generating responses through internal reasoning rather than external tool integration. MLLM-Tools displays significant tool redundancy and prompt engineering overfitting, with repetitive similar calls and formulaic queries that reveal limited comprehension of tool invocation semantics. OpenAI-Agent-SDK produces comprehensive responses but demonstrates insufficient tool exploration, failing to employ alternative strategies when initial tool calls encounter errors.

\subsubsection{Case 2: Spatial Transcriptomic Signature in Oral Squamous Cell Carcinoma}

The sample information for case 3649\_8072\_10393 is shown in \cref{fig:case_study_2_info}. The complete execution trajectories of case 3649\_8072\_10393 for all frameworks are shown in \cref{fig:case_study_2_lammi,fig:case_study_2_mat,fig:case_study_2_react,fig:case_study_2_mllm,fig:case_study_2_openai}.

\begin{figure*}[!htbp]
\begin{tcolorbox}[
  colback=gray!5,
  colframe=black!30,
  fontupper=\fontsize{7.2pt}{8.64pt}\selectfont,
]
\textbf{LAMMI (Qwen3 VL 8B Instruct):} \\
\textbf{Evaluation Result:} TRR: 0.0, TSS: 0.83, ACS: 0.9, HR: 0.2 \\

\textbf{Step 1: Planner Agent Invokes ImageAgent} \\
\textbf{Thought:} The user's query involves analyzing a pathology image for signs of oral squamous cell carcinoma and interpreting whether the spatial transcriptomic signature can reliably distinguish aggressive disease across cancer types. This requires specialized image analysis to evaluate the tumor's invasive phenotype and potentially gene expression patterns. I need to delegate this task to the ImageAgent, which is designed to process pathology images and extract relevant biological and morphological features. \\
\textbf{Action:} ImageAgent \\
\textbf{Action Input:} "Analyze this pathology image of oral squamous cell carcinoma to assess the architectural complexity of the tumor leading edge and determine if the observed invasive phenotype can be linked to spatial transcriptomic signatures that reliably distinguish aggressive disease across cancer types."

\textbf{Step 2: ImageAgent Tool Call} \\
\textbf{Thought:} I need to analyze the pathology image to assess the architectural complexity of the tumor leading edge and determine if the invasive phenotype can be linked to spatial transcriptomic signatures. The QwenVLCaptionTool is suitable for providing a detailed textual analysis of the pathology image. \\
\textbf{Action:} QwenVLCaptionTool \\
\textbf{Action Input:} {\{"text": "The image shows a pathology slide of oral squamous cell carcinoma (OSCC) with a tumor leading edge...", "image\_path": "GSE208253\_GSM6339637\_7x14-43x50.png"\}} \\
\textbf{Observation:} The image shows a pathology slide of oral squamous cell carcinoma (OSCC) with a tumor leading edge. The tumor cells are arranged in a disorganized manner, with irregularly shaped nuclei and prominent nucleoli. The tumor cells are surrounded by a dense stroma of fibroblasts and inflammatory cells. The tumor cells are also seen invading into the surrounding normal tissue, with some areas of necrosis and hemorrhage. The overall architectural complexity of the tumor leading edge is high, with a mixture of solid and glandular structures. The invasive phenotype of the tumor is evident, with tumor cells invading into the surrounding normal tissue and forming microvascular channels. The spatial transcriptomic signatures of the tumor cells are not clearly visible in the image, but the overall pattern of gene expression is consistent with an aggressive phenotype. The image is consistent with the diagnosis of OSCC, and the observed invasive ...

(\textit{Content exceeding six lines has been truncated for display.})

% phenotype can be linked to the spatial transcriptomic signatures of the tumor cells.

\textbf{Step 3: Planner Agent Invokes GeneAgent} \\
\textbf{Thought:} The ImageAgent has provided a detailed analysis of the pathology image, confirming the high architectural complexity and invasive phenotype of the oral squamous cell carcinoma. It also noted that while spatial transcriptomic signatures are not clearly visible in the image, the overall gene expression pattern is consistent with an aggressive phenotype. However, the user's question specifically asks whether the spatial transcriptomic signature here reliably distinguishes aggressive disease across cancer types. The ImageAgent's output does not address this broader, comparative question across cancer types. To answer this, I need to consult a specialist in gene expression and spatial transcriptomics. \\
\textbf{Action:} GeneAgent \\
\textbf{Action Input:} "Can the spatial transcriptomic signature observed in this oral squamous cell carcinoma reliably distinguish aggressive disease across different cancer types?"

\textbf{Step 4: GeneAgent Tool Call 1} \\
\textbf{Thought:} The image provided is a histological section, but it does not contain any gene names or specific molecular data required to answer the question about spatial transcriptomic signatures distinguishing aggressive disease across cancer types. Since the question is gene-related, I need to determine if any biological entities or genes are implied or can be extracted from the context. However, the image alone does not provide textual or gene-related data. I should use the BiomedicalEntityExtractorTool to extract any potential biological entities from the question text. \\
\textbf{Action:} BiomedicalEntityExtractorTool \\
\textbf{Action Input:} {\{"text": "spatial transcriptomic signature, oral squamous cell carcinoma, aggressive disease, different cancer types"\}} \\
\textbf{Observation:} {\{"error": "1 validation error for BiomedicalEntityExtractorSchema.  Field required [type=missing], pydantic error.}\}

\textbf{Step 5: GeneAgent Tool Call 2} \\
\textbf{Thought:} The BiomedicalEntityExtractorTool failed due to an invalid input format. Since the question is gene-related and involves spatial transcriptomic signatures and cancer types, I should use the DocumentGeneQueryTool to retrieve gene-related information that might help answer the question. I will query for "spatial transcriptomic signature" and "oral squamous cell carcinoma" to gather relevant gene summaries. \\
\textbf{Action:} DocumentGeneQueryTool \\
\textbf{Action Input:} {\{"query": "spatial transcriptomic signature"\}} \\
\textbf{Observation:} No results found.

\textbf{Step 6: GeneAgent Tool Call 3} \\
\textbf{Thought:} The DocumentGeneQueryTool did not return any results for "spatial transcriptomic signature." I should now try querying for "oral squamous cell carcinoma" to retrieve relevant gene information. \\
\textbf{Action:} DocumentGeneQueryTool \\
\textbf{Action Input:} {\{"query": "oral squamous cell carcinoma"\}} \\
\textbf{Observation:} 1. Gene entry: BarH like homeobox 2 (Entrez ID: 343472, Correlation Score: 23.115345) \\
Summary: Enables sequence-specific double-stranded DNA binding activity. Predicted to be involved in regulation of transcription by RNA polymerase II. Implicated in oral squamous cell carcinoma. Biomarker of colorectal cancer; oral squamous cell carcinoma; and stomach cancer. \\
--- \\
2. Gene entry: calcium/calmodulin dependent protein kinase II inhibitor 1 (Entrez ID: 55450, Correlation Score: 22.122725) \\
Summary: Enables calcium-dependent protein kinase inhibitor activity and protein kinase binding activity. Implicated in ovarian cancer; ovarian carcinoma ...

(\textit{Content exceeding six lines has been truncated for display.}) 

% and prostate adenocarcinoma. Biomarker of hepatocellular carcinoma; oral squamous cell carcinoma; prostate cancer; and thyroid gland papillary carcinoma. \\
% --- \\
% 3. Gene entry: long intergenic non-protein coding RNA 667 (Entrez ID: 339290, Correlation Score: 19.287077) \\
% Summary: No summary available.

\textbf{GeneAgent Final Answer} \\
\textbf{Final Answer:} The provided image and query do not contain sufficient gene or pathway data to determine whether spatial transcriptomic signatures can reliably distinguish aggressive disease across different cancer types. Additional molecular or clinical data specific to spatial transcriptomics and comparative cancer biology would be required.

\textbf{Planner Agent Final Answer} \\
\textbf{Final Answer:} Based on the analysis, the architectural complexity and invasive phenotype of the tumor leading edge in this oral squamous cell carcinoma sample are evident, consistent with an aggressive phenotype. However, the spatial transcriptomic signature cannot be reliably assessed or validated for its ability to distinguish aggressive disease across different cancer types with the available data. Additional molecular or clinical data specific to spatial transcriptomics and comparative cancer biology would be required to address this broader question.
\end{tcolorbox}
\caption{Trajectory execution for LAMMI (Qwen3 VL 8B Instruct) framework on case 3649\_8072\_10393.}
\label{fig:case_study_2_lammi}
\end{figure*}

\begin{figure*}[!htbp]
\begin{tcolorbox}[
  colback=gray!5,
  colframe=black!30,
]
\textbf{MAT-Agent (Qwen3 VL 8B Instruct):} \\
\textbf{Evaluation Result:} TRR: 0.0, TSS: 0.5, ACS: 0.3, HR: 0.9 \\

\textbf{Step 1:} \\
\textbf{Thought:} I need to analyze the image and provide an answer based on the analysis. I will use the blip\_pathology\_analysis tool to get genetic information about the image. \\
\textbf{Action:} BLIPTool \\
\textbf{Action Input:} {\{"text": "Can we confidently assert that the spatial transcriptomic signature here reliably distinguishes aggressive disease across cancer types?", "image\_path": "GSE208253\_GSM6339637\_7x14-43x50.png"\}} \\
\textbf{Observation:} Homo sapiens breast tissue the oncotree code is idc and highly expressed genes are mt - co2, erbb2, scd, mt - co3, cyb.

\textbf{Step 2:} \\
\textbf{Thought:} I will now analyze the genetic information provided by the blip\_pathology\_analysis tool to answer the question. \\
\textbf{Final Answer:} Yes, the spatial transcriptomic signature in this oral squamous cell carcinoma sample can confidently distinguish aggressive disease across cancer types based on the genetic information provided.
\end{tcolorbox}
\caption{Trajectory execution for MAT-Agent (Qwen3 VL 8B Instruct) framework on case 3649\_8072\_10393.}
\label{fig:case_study_2_mat}
\end{figure*}

\begin{figure*}[!htbp]
\begin{tcolorbox}[
  colback=gray!5,
  colframe=black!30,
]
\textbf{ReACT (Qwen3 VL 8B Instruct):} \\
\textbf{Evaluation Result:} TRR: 0.0, TSS: 0.83, ACS: 0.3, HR: 0.9 \\

\textbf{Step 1:} \\
\textbf{Thought:} To address the question, I need to analyze the spatial transcriptomic signature of the oral squamous cell carcinoma sample and determine if it can reliably distinguish aggressive disease across cancer types. This requires understanding the genetic and molecular features of the tumor, as well as its spatial architecture. I should start by using a tool that can provide genetic information related to the pathology image. \\
\textbf{Action:} BLIPTool \\
\textbf{Action Input:} {\{"instruction": "What are the genes with high expression, species, and tissue correlated to this oral squamous cell carcinoma sample?", "image\_name": "GSE208253\_GSM6339637\_7x14-43x50.png"\}} \\
\textbf{Observation:} {\{"error": "2 validation errors for BLIPToolSchema.  Field required [type=missing], pydantic error.}\}

\textbf{Step 2:} \\
\textbf{Thought:} The BLIPTool failed due to invalid input format. I should use QwenVLCaptionTool to analyze the pathology image and extract relevant information about the tumor's architecture and potential gene expressions. \\
\textbf{Action:} QwenVLCaptionTool \\
\textbf{Action Input:} {\{"text": "The image shows a section of oral squamous cell carcinoma (OSCC) tissue...", "image\_path": "GSE208253\_GSM6339637\_7x14-43x50.png"\}} \\
\textbf{Observation:} The image shows a section of oral squamous cell carcinoma (OSCC) tissue. The tumor cells are arranged in a complex, irregular pattern, with some areas showing a more organized structure and others appearing more disorganized. The tumor cells are surrounded by a dense stroma, which is composed of fibroblasts, blood vessels, and other connective tissue components. The tumor cells are also seen to be invading into the surrounding tissue, with some areas showing a more aggressive phenotype, characterized by a higher degree of cellular atypia and increased mitotic activity. The tumor cells are also seen to be expressing a number of genes that are associated with the invasive phenotype, including matrix metalloproteinases (MMPs), integrins, and adhesion molecules. Overall, the architectural complexity of the tumor leading edge and the observable features of the tumor cells suggest that this is a highly aggressive form of OSCC that is likely to have a poor prognosis.

\textbf{Step 3:} \\
\textbf{Thought:} The QwenVLCaptionTool provided a detailed description of the oral squamous cell carcinoma (OSCC) tissue, highlighting its architectural complexity, invasive phenotype, and expression of genes associated with invasion (e.g., MMPs, integrins, adhesion molecules). However, to determine whether this spatial transcriptomic signature can reliably distinguish aggressive disease across cancer types, I need to explore the broader context of cancer biology and gene expression patterns across different cancers. I should use the OncoTreeTool to investigate upstream/downstream nodes and tissue mapping related to OSCC, which may help identify conserved pathways or genes across cancer types. \\
\textbf{Action:} OncoTreeTool \\
\textbf{Action Input:} {\{"query": "oral squamous cell carcinoma", "query\_type": "tissue"\}} \\
\textbf{Observation:} No results found for query 'oral squamous cell carcinoma'

\textbf{Step 4:} \\
\textbf{Final Answer:} Yes, we can confidently assert that the spatial transcriptomic signature of the oral squamous cell carcinoma sample, characterized by the expression of genes associated with invasion (e.g., MMPs, integrins, adhesion molecules), reliably distinguishes aggressive disease across cancer types. These genes are conserved across multiple cancers and are known biomarkers of tumor invasiveness and metastasis, supporting the notion that the observed architectural complexity and invasive phenotype are indicative of aggressive disease.
\end{tcolorbox}
\caption{Trajectory execution for ReACT (Qwen3 VL 8B Instruct) framework on case 3649\_8072\_10393.}
\label{fig:case_study_2_react}
\end{figure*}

\begin{figure*}[!htbp]
\begin{tcolorbox}[
  colback=gray!5,
  colframe=black!30,
]
\textbf{MLLM-Tools (ToolACE 8B):} \\
\textbf{Evaluation Result:} TRR: 0.0, TSS: 0.5, ACS: 0.2, HR: 0.9 \\

\textbf{Step 1:} \\
\textbf{Thought:} what should I do? \\
\textbf{Action:} QwenVLCaptionTool \\
\textbf{Action Input:} {\{"text": "Describe the spatial transcriptomic signature in this oral squamous cell carcinoma sample, focusing on its reliability in distinguishing aggressive disease across cancer types.", "image\_path": "GSE208253\_GSM6339637\_7x14-43x50.png"\}} \\
\textbf{Observation:} The spatial transcriptomic signature in this oral squamous cell carcinoma (OSCC) sample is characterized by a dense, uniform distribution of cells with high expression of MDM2. This uniformity and intensity are striking features, as MDM2 is a highly sensitive and specific marker for certain types of OSCC, particularly those exhibiting 12p loss and amplification, which often correlates with aggressive behavior. In contrast, other areas of the sample show less intense staining, suggesting heterogeneity within the tumor. While this pattern is highly suggestive of an aggressive disease, it's important to note that the absence of MDM2 expression in some regions does not rule out malignancy. The heterogeneity could represent different stages of tumor evolution or varying degrees of differentiation within the tumor itself. Therefore, while this sample is classified as having a high-risk profile based on MDM2 expression, it demonstrates the complexity of predicting tumor behavior and the need for comprehensive diagnostic evaluations.

\textbf{Step 2:} \\
\textbf{Thought:} what should I do? \\
\textbf{Action:} ProteinAtlasGeneInfoTool \\
\textbf{Action Input:} {\{"gene": "MDM2"\}} \\
\textbf{Observation:} Gene: MDM2, Gene name: HDM2, MGC5370, Gene description: MDM2 proto-oncogene, Protein class: Cancer-related genes, Disease related genes, Human disease related genes, Metabolic proteins, Potential drug targets... Protein function: E3 ubiquitin-protein ligase that mediates ubiquitination of p53/TP53, leading to its degradation by the proteasome. Inhibits p53/TP53- and p73/TP73-mediated cell cycle arrest and apoptosis by binding its transcriptional activation domain... Gene summary: This gene encodes a nuclear-localized E3 ubiquitin ligase. The encoded protein can promote tumor formation by targeting tumor suppressor proteins, such as p53, for proteasomal degradation. This gene is itself transcriptionally-regulated by p53. Overexpression or amplification of this locus is detected in a variety of different cancers...

\textbf{Step 3:} \\
\textbf{Thought:} what should I do? \\
\textbf{Action:} PathwayKGTool \\
\textbf{Action Input:} {\{"gene1": "MDM2", "gene2": "TP53", "radius": 1, "use\_in\_edges": false\}} \\
\textbf{Observation:} weighted\_jaccard: 0.022364217252396165, intersection\_nodes\_example: ["CREBBP", "TFIP11", "TP73", "YY1", "CACNG2"], summary: "Weighted Jaccard similarity between MDM2 and TP53 is 0.022 with 28 overlapping nodes (showing up to 5)."

\textbf{Step 4:} \\
\textbf{Output:} Agent stopped due to iteration limit or time limit.
\end{tcolorbox}
\caption{Trajectory execution for MLLM-Tools (ToolACE 8B) framework on case 3649\_8072\_10393.}
\label{fig:case_study_2_mllm}
\end{figure*}

\begin{figure*}[!htbp]
\begin{tcolorbox}[
  colback=gray!5,
  colframe=black!30,
]
\textbf{OpenAI-Agent-SDK (GPT-5):} \\
\textbf{Evaluation Result:} TRR: 0.0, TCF1: 0.5, TSS: 1.0, ACS: 0.7, HR: 0.0 \\

\textbf{Step 1:} \\
\textbf{Action:} QwenVLCaptionTool \\
\textbf{Action Input:} \{"text": "Describe the histopathologic features in this H\&E-stained oral squamous cell carcinoma leading edge, focusing on invasive front architecture (keratinization, budding nests, stromal reaction, necrosis, inflammatory infiltrate).", "image\_path": "GSE208253\_GSM6339637\_7x14-43x50.png"\} \\
\textbf{Observation:} The image shows an invasive squamous cell carcinoma at the leading edge. Key features include: 1) **Keratinization:** There is evidence of individual cell keratinization and potentially early keratin pearl formation, indicating squamous differentiation. 2) **Budding Nests \& Strands:** Small clusters and linear arrangements of malignant squamous cells infiltrate the stroma, characteristic of invasion. 3) **Desmoplastic Stromal Reaction:** The surrounding connective tissue (stroma) appears reactive, characterized by increased cellularity (fibroblasts, inflammatory cells) and collagen deposition, responding to the tumor invasion. 4) **Necrosis:** Areas of cell death are present, likely due to tumor growth outstripping its blood supply or direct cytotoxic effects. 5) **Inflammatory Infiltrate:** Numerous inflammatory cells (lymphocytes, plasma cells, neutrophils) are scattered within the stroma and amongst the tumor cells, representing a host immune response to the malignancy.

\textbf{Step 2:} \\
\textbf{Action:} BLIPTool \\
\textbf{Action Input:} \{"text": "Summarize the key morphologic patterns visible in this H\&E section of oral squamous cell carcinoma at the tumor leading edge (e.g., tumor nests, keratinization, stromal reaction, inflammation).", "image\_path": "GSE208253\_GSM6339637\_7x14-43x50.png"\} \\
\textbf{Observation:} Homo sapiens breast tissue the oncotree code is idc and highly expressed genes are igkc, ighg4, rpl41, mt - n and.

\textbf{Final Answer:} Short answer: no—without multi‑cancer external validation, you can't confidently claim that a spatial leading‑edge signature from oral SCC will reliably distinguish aggressive disease across cancer types.

What the image supports
- The invasive front shows budding nests/strands, squamous differentiation with keratinization, reactive stroma, necrosis, and prominent inflammatory infiltrate—features consistent with an invasion‑associated program.

Why cross‑cancer reliability is uncertain
- Biological context: Invasive‑edge programs share modules (partial EMT/TGF‑$\beta$, CAF/ECM remodeling, hypoxia, myeloid inflammation) but their gene composition and thresholds are tissue‑ and lineage‑specific.
- Cell‑mixing in ST: Leading‑edge spots often blend tumor, CAFs, and immune cells; signatures may reflect stromal content rather than tumor‑intrinsic aggressiveness.
- Etiology and site effects: HPV status, keratinization, oral mucosa vs lung/breast/colon microenvironments alter the transcriptomic pattern.
- Technical heterogeneity: Platform resolution, capture area, and batch effects affect transferability.
- Prognostic transfer: Pan‑cancer EMT/CAF scores correlate with worse outcomes in many carcinomas but fail or invert in some settings; calibration is required.

Recommended validation strategy
- Derive a tumor‑intrinsic signature by deconvolving ST spots (e.g., cell2location/RCTD/SPOTlight) and modeling spatial interaction with CAFs/immune cells.
- Train in HNSC/OSCC; test in independent OSCC and other carcinomas (lung/eso SCC, colorectal, breast) with nested cross‑validation and domain adaptation (Harmony/ComBat).
- Assess discrimination for metastasis/recurrence/survival (AUC, calibration) and perform site‑specific recalibration.
- Orthogonal validation with multiplex IHC for canonical markers (e.g., VIM, ZEB1, EPCAM loss, ITGA5/ITGB1, MMP9, COL1A1/POSTN/SPP1; CAF: ACTA2, FAP; hypoxia: CA9).
- Report robustness to stromal proportion and batch effects; include decision thresholds per cancer type.

Bottom line
Your spatial signature likely captures a conserved invasive‑edge phenotype, but its ability to reliably distinguish "aggressive" disease across cancer types must be demonstrated with multi‑cohort, cross‑cancer validation and careful control for stromal mixing and technical variation. If you share the gene list, I can check pan‑cancer consistency and disease associations to prioritize a transferable core module.
\end{tcolorbox}
\caption{Trajectory execution for OpenAI-Agent-SDK (GPT-5) framework on case 3649\_8072\_10393.}
\label{fig:case_study_2_openai}
\end{figure*}

This case examines whether spatial transcriptomic signatures at the tumor leading edge can reliably distinguish aggressive disease across different cancer types, requiring cross-cancer comparative analysis to assess molecular program conservation. MAT-Agent performs a single BLIPTool invocation that incorrectly returns breast cancer tissue information instead of oral squamous cell carcinoma data, then concludes affirmatively without any cross-cancer type validation, fundamentally missing the query's comparative requirement. ReACT demonstrates simulated tool invocation in \cref{fig:case_study_2_react}, where failed OncoTreeTool queries and invalid BLIPTool calls are followed by an affirmative conclusion that contradicts the reference answer's negative stance regarding cross-cancer type reliability. MLLM-Tools focuses solely on MDM2 gene expression without cross-cancer comparison and terminates prematurely due to execution time limits. OpenAI-Agent-SDK produces a correct negative conclusion with detailed cross-cancer validation reasoning, but this analysis derives primarily from pretrained knowledge rather than tool-derived comparative evidence, as the BLIPTool returns mismatched breast cancer data. LAMMI addresses the cross-modal nature of this query by coordinating ImageAgent for architectural analysis and GeneAgent for molecular investigation, recognizing the need for comparative cross-cancer validation even when direct tool evidence is unavailable.

\subsubsection{Case 3: Histological Pattern Analysis in Glioblastoma}

The sample information for case 1511\_6709\_6887\_8022 is shown in \cref{fig:case_study_3_info}. The complete execution trajectories of case 1511\_6709\_6887\_8022 for all frameworks are shown in \cref{fig:case_study_3_lammi,fig:case_study_3_mat,fig:case_study_3_react,fig:case_study_3_mllm,fig:case_study_3_openai}.

\begin{figure*}[!htbp]
\begin{tcolorbox}[
  colback=gray!5,
  colframe=black!30,
  fontupper=\fontsize{9.5pt}{11.4pt}\selectfont,
]
\textbf{LAMMI (MiniCPM-V 4.5):} \\
\textbf{Evaluation Result:} TRR: 0.0, TCF1: 0.33, TSS: 1.0, ACS: 0.8, HR: 0.2 \\

\textbf{Step 1: Planner Agent Invokes ImageAgent} \\
\textbf{Thought:} The user has provided an image of a glioblastoma sample and requested analysis for distinct histological patterns such as nuclear atypia, mitotic activity, microvascular proliferation, and their correlation with the OncoTree classification. This requires detailed visual inspection and expert knowledge in pathology. \\
\textbf{Action:} ImageAgent \\
\textbf{Action Input:} "Please analyze this glioblastoma sample to identify signs of nuclear atypia, mitotic activity, and microvascular proliferation, and explain how these features correlate with its assigned OncoTree classification."

\textbf{Step 2: ImageAgent Tool Call 1} \\
\textbf{Thought:} To analyze the glioblastoma sample, I should first use a tool to identify specific histopathological features such as nuclear atypia, mitotic activity, and microvascular proliferation. Then, using another tool, I can correlate these findings with its OncoTree classification. \\
\textbf{Action:} BLIPTool \\
\textbf{Action Input:} {\{"text": "Describe any notable histopathological features in this slide related to glioblastoma, including cellular morphology and tissue architecture.", "image\_path": "NCBI375\_16.8x22.11.png"\}} \\
\textbf{Observation:} Mus musculus brain shows healthy with oncotree code unknown key overexpressed genes are mbp, plp1, fth1, mobp, act.

\textbf{Step 3: ImageAgent Tool Call 2} \\
\textbf{Thought:} The BLIPTool provided information about the genetic expression and species, but did not directly address histopathological features such as nuclear atypia, mitotic activity, and microvascular proliferation. To get a more detailed description of these specific features in the context of glioblastoma, I should use another tool that can provide an image-based analysis. \\
\textbf{Action:} QwenVLCaptionTool \\
\textbf{Action Input:} {\{"text": "Analyze this glioblastoma sample to identify signs of nuclear atypia, mitotic activity, and microvascular proliferation.", "image\_path": "NCBI375\_16.8x22.11.png"\}} \\
\textbf{Observation:} The sample under examination is from the glioblastoma multiforme (GBM) cell line, which is known for its aggressive nature. The nuclei appear hyperchromatic and irregular, indicative of nuclear atypia. This irregularity in nuclear morphology is a hallmark of malignancy, suggesting abnormal cell growth and division. In addition to nuclear atypia, there is evidence of mitotic activity. The presence of cells in various stages of cell division, such as prophase, metaphase, anaphase, and telophase, further supports the diagnosis of a high-grade tumor like glioblastoma. Microvascular proliferation is also observed in this sample. The blood vessels appear dilated and tortuous, with multiple layers of  ...

(\textit{Content exceeding six lines has been truncated for display.})
% endothelial cells lining them. This is a characteristic feature of glioblastoma and contributes to its aggressive behavior. In summary, the histopathological features observed in this glioblastoma sample, including nuclear atypia, mitotic activity, and microvascular proliferation, are consistent with the diagnosis of a high-grade brain tumor.

\textbf{Step 4: ImageAgent Tool Call 3} \\
\textbf{Thought:} Now that I have identified the histopathological features of the glioblastoma sample, I should use the OncoTreeTool to correlate these findings with its assigned OncoTree classification. \\
\textbf{Action:} OncoTreeTool \\
\textbf{Action Input:} {\{"query": "glioblastoma", "query\_type": "tumor"\}} \\
\textbf{Observation:} Tumor/Disease: Glioblastoma, IDH-Wildtype (GB) \\
**Main Type**: Glioma \\
**Tissue/Organ**: CNS/Brain \\
**Upstream Nodes**: \{'parent': 'ADIFG', 'precursors': []\} \\
**Downstream Nodes**: None \\
--- \\
(\textit{Content exceeding six lines has been truncated for display.})

% % , Tumor/Disease: Small Cell Glioblastoma (SCGBM) \\
% **Main Type**: Glioma \\
% **Tissue/Organ**: CNS/Brain \\
% **Upstream Nodes**: \{'parent': 'GB', 'precursors': []\} \\
% **Downstream Nodes**: None \\
% ---
\textbf{ImageAgent Final Answer} \\
\textbf{Final Answer:} The analysis of the glioblastoma sample shows signs of nuclear atypia, mitotic activity, and microvascular proliferation. These features correlate with its assigned OncoTree classification as Glioma in the CNS/Brain tissue category.

\textbf{Planner Agent Final Answer} \\
\textbf{Final Answer:} The analysis of the glioblastoma sample shows signs of nuclear atypia, mitotic activity, and microvascular proliferation. These features correlate with its assigned OncoTree classification as Glioma in the CNS/Brain tissue category.
\end{tcolorbox}
\caption{Trajectory execution for LAMMI (MiniCPM-V 4.5) framework on case 1511\_6709\_6887\_8022.}
\label{fig:case_study_3_lammi}
\end{figure*}

\begin{figure*}[!htbp]
\begin{tcolorbox}[
  colback=gray!5,
  colframe=black!30,
]
\textbf{MAT-Agent (MiniCPM-V 4.5):} \\
\textbf{Evaluation Result:} TRR: 0.75, TCF1: 0.0, TSS: 0.5, ACS: 0.3, HR: 0.9 \\

\textbf{Step 1:} \\
\textbf{Thought:} To answer the question, I need to analyze the glioblastoma sample using blip\_pathology\_analysis and then use oncotree\_query to get information about its OncoTree classification. \\
\textbf{Code:} \\
{\# Analyze the glioblastoma sample} \\
{analysis\_result = blip\_pathology\_analysis(text="What distinct histological patterns reveal the aggressive nature of the tumor?", image\_path="NCBI375\_16.8x22.11.png")} \\
{\# Use oncotree\_query to get information about its OncoTree classification} \\
{oncotree\_classification = oncotree\_query(query="glioblastoma", query\_type="tumor")} \\
{print(oncotree\_classification)} \\
\textbf{Observation:} \\
{Tumor/Disease: Glioblastoma, IDH-Wildtype (GB)} \\
{**Main Type**: Glioma} \\
{**Tissue/Organ**: CNS/Brain} \\
{**Upstream Nodes**: \{'parent': 'ADIFG', 'precursors': []\}} \\
{**Downstream Nodes**: None} \\
{---} \\
(\textit{Content exceeding six lines has been truncated for display.})
% {Tumor/Disease: Small Cell Glioblastoma (SCGBM)} \\
% {**Main Type**: Glioma} \\
% {**Tissue/Organ**: CNS/Brain} \\
% {**Upstream Nodes**: \{'parent': 'GB', 'precursors': []\}} \\
% {**Downstream Nodes**: None} \\
% {---}

\textbf{Step 2:} \\
\textbf{Thought:} I will try to use qwen\_vl\_pathology\_qa to get an answer about the histological patterns in glioblastoma. \\
\textbf{Code:} \\
{\# Use qwen\_vl\_pathology\_qa to get an answer about the histological patterns in glioblastoma} \\
{pathology\_answer = qwen\_vl\_pathology\_qa(text="What distinct histological patterns reveal the aggressive nature of the tumor?", image\_path="NCBI375\_16.8x22.11.png")} \\
{print(pathology\_answer)} \\
\textbf{Observation:} \\
{Print outputs:} \\
{The tumor exhibits areas with a storiform pattern, characterized by a swirling arrangement of spindle cells, and regions showing a whorled pattern, where cells form concentric circles. These patterns are indicative of the tumor's aggressive behavior and are often associated with malignant neoplasms.}

\textbf{Step 3-8:} \\
\textbf{Thought:} I will try to use the provided tools to generate an image showcasing the distinct histological patterns in glioblastoma. \\
\textbf{Code:} \\
{\# Generate an image showcasing the distinct histological patterns in glioblastoma using blip\_pathology\_analysis and oncotree\_query results} \\
{image = generate\_image(title="Distinct histological patterns in glioblastoma", description=analysis\_result + " and OncoTree classification of glioblastoma")} \\
{print(image)} \\
\textbf{Observation:} \\
{Error: It is not permitted to evaluate other functions than the provided tools or functions defined in previous code (tried to execute generate\_image).} \\
(\textit{Steps 3-8 repeat the same error.})

\textbf{Final Answer:} I apologize for the errors. Let's try a different approach to generate an image showcasing the distinct histological patterns in glioblastoma. Please note that this solution may not provide the desired result due to the limitations of the provided tools.
\end{tcolorbox}
\caption{Trajectory execution for MAT-Agent (MiniCPM-V 4.5) framework on case 1511\_6709\_6887\_8022.}
\label{fig:case_study_3_mat}
\end{figure*}

\begin{figure*}[!htbp]
\begin{tcolorbox}[
  colback=gray!5,
  colframe=black!30,
]
\textbf{ReACT (MiniCPM-V 4.5):} \\
\textbf{Evaluation Result:} TRR: 0.0, TCF1: 0.4, TSS: 0.83, ACS: 0.8, HR: 0.2 \\

\textbf{Step 1:} \\
\textbf{Thought:} To address the question, I need to first analyze the histological features visible in the image using a tool that can interpret pathology images. Once the specific patterns (nuclear atypia, mitotic activity, microvascular proliferation) are identified, I will use another tool to map these findings onto the OncoTree classification system. \\
\textbf{Action:} QwenVLCaptionTool \\
\textbf{Action Input:} {\{"text": "glioblastoma", "image\_path": "glioblastoma.jpg"\}} \\
\textbf{Observation:} API call failed: Image file does not exist: glioblastoma.jpg

\textbf{Step 2:} \\
\textbf{Thought:} The image path provided is incorrect. The correct image path should be used to analyze the histological features. \\
\textbf{Action:} QwenVLCaptionTool \\
\textbf{Action Input:} {\{"text": "What distinct histological patterns in this glioblastoma sample—such as nuclear atypia, mitotic activity, or microvascular proliferation—reveal the aggressive nature of the tumor, and how do they correlate with its assigned OncoTree classification?", "image\_path": "NCBI375\_16.8x22.11.png"\}} \\
\textbf{Observation:} The glioblastoma sample exhibits several key histological patterns indicative of its aggressive nature. These include pronounced nuclear atypia, characterized by irregularly shaped and sized nuclei, and high mitotic activity, reflecting rapid cell division. Microvascular proliferation is also evident, with an increased number of small, abnormal blood vessels. These features collectively suggest high-grade malignancy, consistent with the OncoTree classification of glioblastoma.

\textbf{Step 3:} \\
\textbf{Thought:} Now that the histological features of the glioblastoma sample have been identified, I need to use a tool to map these findings onto the OncoTree classification system. \\
\textbf{Action:} OncoTreeTool \\
\textbf{Action Input:} {\{"query": "glioblastoma", "query\_type": "tumor"\}} \\
\textbf{Observation:} [ \\
"Tumor/Disease: Glioblastoma, IDH-Wildtype (GB) \\
**Main Type**: Glioma \\
**Tissue/Organ**: CNS/Brain \\
**Upstream Nodes**: \{'parent': 'ADIFG', 'precursors': []\} \\
**Downstream Nodes**: None \\
--- \\
(\textit{Content exceeding six lines has been truncated for display.})

% "Tumor/Disease: Small Cell Glioblastoma (SCGBM) \\
% **Main Type**: Glioma \\
% **Tissue/Organ**: CNS/Brain \\
% **Upstream Nodes**: \{'parent': 'GB', 'precursors': []\} \\
% **Downstream Nodes**: None \\
% --- \\

% ]

\textbf{Final Answer:} The glioblastoma sample exhibits pronounced nuclear atypia, high mitotic activity, and microvascular proliferation, which reveal its aggressive nature. These histological features correspond to the OncoTree classification of IDH-wildtype Glioblastoma (GB), highlighting the malignant potential associated with these characteristics.
\end{tcolorbox}
\caption{Trajectory execution for ReACT (MiniCPM-V 4.5) framework on case 1511\_6709\_6887\_8022.}
\label{fig:case_study_3_react}
\end{figure*}

\begin{figure*}[!htbp]
\begin{tcolorbox}[
  colback=gray!5,
  colframe=black!30,
]
\textbf{MLLM-Tools (Vicuna v1.5 7B + ImageBind):} \\
\textbf{Evaluation Result:} TRR: 0.0, TCF1: 0.0, TSS: 0.75, ACS: 0.8, HR: 0.7 \\

\textbf{Step 1:} \\
\textbf{Thought:} What distinct histological patterns in this glioblastoma sample—such as nuclear atypia, mitotic activity, or microvascular proliferation—reveal the aggressive nature of the tumor, and how do they correlate with its assigned OncoTree classification? \\
\textbf{Action:} BLIPTool \\
\textbf{Action Input:} {\{"text": "What distinct histological patterns in this glioblastoma sample—such as nuclear atypia, mitotic activity, or microvascular proliferation—reveal the aggressive nature of the tumor, and how do they correlate with its assigned OncoTree classification?", "image\_path": "NCBI375\_16.8x22.11.png"\}} \\
\textbf{Observation:} Mus musculus brain tissue the oncotree code is unknown and highly expressed genes are mbp, snap25, atp1b1, cpe, gnas itm2.

\textbf{Step 2:} \\
\textbf{Thought:} \\
\textbf{Action:} EchoTool \\
\textbf{Action Input:} {\{"error": "Could not parse Action / Action Input. Please follow the format: Thought/Action/Action Input or Final Answer."\}} \\
\textbf{Observation:} EchoTool is not a valid tool, try one of [BLIPTool, QwenVLCaptionTool, OncoTreeTool, PathwayKGTool, EnsemblToDatabaseTool, ProteinAtlasGeneInfoTool, DocumentGeneQueryTool, BiomedicalEntityExtractorTool].

\textbf{Final Answer:} The glioblastoma sample exhibits distinct histological patterns of nuclear atypia, mitotic activity, and microvascular proliferation, which are indicative of its aggressive nature. These patterns are highly correlated with the tumor's OncoTree classification, which reflects its aggressive behavior and poor prognosis.
\end{tcolorbox}
\caption{Trajectory execution for MLLM-Tools (Vicuna) framework on case 1511\_6709\_6887\_8022.}
\label{fig:case_study_3_mllm}
\end{figure*}

\begin{figure*}[!htbp]
\begin{tcolorbox}[
  colback=gray!5,
  colframe=black!30,
]
\textbf{OpenAI-Agent-SDK (GPT-5):} \\
\textbf{Evaluation Result:} TRR: 0.4, TCF1: 0.4, TSS: 0.6, ACS: 0.1, HR: 0.0 \\

\textbf{Step 1:} \\
\textbf{Action:} QwenVLCaptionTool \\
\textbf{Action Input:} \{"text": "You are a pathology assistant. Inspect this H\&E-like brain tumor image for glioblastoma histologic features. Identify if you can see any of the following: 1) nuclear atypia/pleomorphism, 2) mitotic figures, 3) microvascular proliferation (tufted/endothelial multilayering), 4) areas suggestive of necrosis or pseudopalisading, 5) high cellularity. Provide a concise description of visible patterns.", "image\_path": "NCBI375\_16.8x22.11.png"\} \\
\textbf{Observation:} The image displays a brain tissue sample with a purple hue, indicative of a standard hematoxylin and eosin (H\&E) stain. The tissue architecture appears disrupted, with irregularly shaped cells that exhibit nuclear atypia, suggesting possible malignancy. There are also areas of dense cellularity, hinting at a high proliferation rate. Some regions show a lack of staining, possibly indicating necrosis. The presence of endothelial multilayering suggests vascular proliferation. These features collectively suggest a high-grade glioma, such as glioblastoma.

\textbf{Step 2:} \\
\textbf{Action:} OncoTreeTool \\
\textbf{Action Input:} \{"query": "glioblastoma", "query\_type": "tumor"\} \\
\textbf{Observation:} An error occurred while running the tool. Please try again. Error: The type of input `query\_type` of OncoTreeTool.apply` is not supported. Supported types are ImageIO, bool, int, AudioIO, str, File, float

\textbf{Step 3-5:} Agent attempts OncoTreeTool with variations (query: "Glioblastoma Multiforme", query\_type: ""), all resulting in the same error.

\textbf{Final Answer:} Observed histologic patterns in the provided field:
- Nuclear atypia/pleomorphism: nuclei appear variably sized and irregular with hyperchromasia, consistent with malignant astrocytic cells.
- High cellularity: crowding of atypical cells suggests rapid growth.
- Mitotic activity: no unequivocal mitotic figures are identifiable in this small, low‑resolution field.
- Microvascular proliferation: classic glomeruloid/endothelial multilayering is not clearly seen in this view.
- Necrosis/pseudopalisading: not evident in this specific field.

Interpretation and correlation with OncoTree:
- Glioblastoma is defined histologically by marked atypia with brisk ...

(\textit{Content exceeding six lines has been truncated for display.})

% mitotic activity and, crucially, microvascular proliferation and/or necrosis (WHO CNS grade 4). While this image shows nuclear atypia and high cellularity, it does not definitively capture mitoses or microvascular proliferation—features typically found elsewhere on the slide in GBM.
% - OncoTree classification for this tumor is GBM (Glioblastoma), primary site: brain/CNS. This taxonomy corresponds to an aggressive, grade‑4 diffuse astrocytic tumor, most commonly IDH‑wildtype. The hallmark patterns (when present across the specimen)—microvascular proliferation and/or necrosis, together with atypia and mitoses—justify the GBM designation and reflect its poor prognosis.

% Note: A definitive assessment requires whole‑slide review; GBM criteria are often met in areas not represented in this single field.
\end{tcolorbox}
\caption{Trajectory execution for OpenAI-Agent-SDK (GPT-5) framework on case 1511\_6709\_6887\_8022.}
\label{fig:case_study_3_openai}
\end{figure*}

This case requires identifying specific histological patterns in glioblastoma samples, including nuclear atypia, mitotic activity, and microvascular proliferation, and correlating these morphological features with OncoTree classification to understand their diagnostic significance. LAMMI demonstrates robust multi-modal analysis capabilities by sequentially invoking BLIPTool and QwenVLCaptionTool through ImageAgent, effectively recovering from the initial BLIPTool error that returned incorrect species information by adaptively switching to the vision-language captioning tool. The framework then successfully queries OncoTreeTool to establish the correlation between identified histological features and the tumor classification, producing a comprehensive answer that addresses both morphological analysis and molecular taxonomy integration. MAT-Agent exhibits significant tool redundancy with repeated failed attempts to invoke non-existent functions, resulting in an incomplete response that fails to establish the OncoTree correlation. ReACT initially uses an incorrect image path but recovers to identify histological features, though it lacks the systematic multi-modal verification that LAMMI provides. MLLM-Tools performs only a single BLIPTool invocation that returns incorrect species data, then terminates without OncoTree integration, leaving the query partially unanswered. OpenAI-Agent-SDK successfully identifies histological patterns through vision-language analysis but encounters repeated OncoTreeTool parameter errors, preventing the critical correlation between morphological features and tumor classification that the query requires.

\subsubsection{Case 4: Tumor Subtyping in the Absence of OncoTree Classification}

The sample information for case 1317\allowbreak\_5021\allowbreak\_7882\allowbreak\_9140\allowbreak\_9288\allowbreak\_9676\allowbreak\_9691\allowbreak\_10334 is shown in \cref{fig:case_study_4_info}. The complete execution trajectories of case 1317\allowbreak\_5021\allowbreak\_7882\allowbreak\_9140\allowbreak\_9288\allowbreak\_9676\allowbreak\_9691\allowbreak\_10334 for all frameworks are shown in \cref{fig:case_study_4_lammi,fig:case_study_4_mat,fig:case_study_4_react,fig:case_study_4_mllm,fig:case_study_4_openai}.

\begin{figure*}[!htbp]
\centering
\resizebox{1\textwidth}{!}{\begin{tcolorbox}[
  colback=gray!5,
  colframe=black!30,
  fontupper=\fontsize{6.9pt}{8.28pt}\selectfont,
]
\textbf{LAMMI (Qwen-VL 8B Instruct):} \\
\textbf{Evaluation Result:} TRR: 0.0, TCF1: 0.6, TSS: 0.78, ACS: 0.8, HR: 0.8 \\

\textbf{Step 1: Planner Agent Invokes ImageAgent} \\
\textbf{Thought:} The user is asking about tumor subtyping in the absence of an OncoTree classification, specifically focusing on the cautious approach required when relying solely on histomorphology and what ancillary studies might be needed. This requires expert analysis of histopathological features and guidance on appropriate ancillary diagnostic tools. Since I am not equipped to directly analyze pathology images or provide nuanced clinical guidance on tumor classification, I need to delegate this to the appropriate specialist. \\
\textbf{Action:} ImageAgent \\
\textbf{Action Input:} "Analyze this pathology image for histomorphological features suggestive of tumor subtypes, and identify any ambiguous or atypical features that might necessitate ancillary studies for accurate classification."

\textbf{Step 2: ImageAgent Tool Call 1} \\
\textbf{Thought:} I need to analyze the histopathology image for histomorphological features suggestive of tumor subtypes and identify any ambiguous or atypical features. The QwenVLCaptionTool is suitable for providing a detailed textual analysis of the pathology image based on visual features. \\
\textbf{Action:} QwenVLCaptionTool \\
\textbf{Action Input:} \{"text": "The image shows a well-differentiated squamous cell carcinoma with keratin pearls and intercellular bridges. There are no atypical features that would necessitate ancillary studies for accurate classification.", "image\_path": "MEND160\_ATTCGTGCTATCTCTT-1.png"\} \\
\textbf{Observation:} Your observation about the well-differentiated squamous cell carcinoma is correct, and your reasoning about the absence of atypical features is also valid. However, the diagnosis might not be limited to squamous cell carcinoma. Hint: Consider the possibility of other pathologies that could coexist with squamous cell carcinoma. For instance, what if there were small nests of cells completely surrounded by stroma within the squamous mucosa? What could this suggest?

\textbf{Step 3: ImageAgent Tool Call 2} \\
\textbf{Thought:} The observation suggests considering other pathologies that might coexist with squamous cell carcinoma, such as small nests of cells surrounded by stroma. To further analyze this, I should use the OncoTreeTool to explore potential tissue mappings and related tumor subtypes that could be relevant. \\
\textbf{Action:} OncoTreeTool \\
\textbf{Action Input:} \{"query": "squamous cell carcinoma", "query\_type": "tumor"\} \\
\textbf{Observation:} Multiple subtypes of squamous cell carcinoma across various tissues, including head and neck, skin, lung, and others.

\textbf{Step 4: ImageAgent Tool Call 3} \\
\textbf{Thought:} The BLIPTool failed due to an incorrect image path. I should retry with the correct image path provided in the initial input. \\
\textbf{Action:} BLIPTool \\
\textbf{Action Input:} \{"text": "Analyze this pathology image for histomorphological features suggestive of tumor subtypes, and identify any ambiguous or atypical features that might necessitate ancillary studies for accurate classification.", "image\_path": "MEND160\_ATTCGTGCTATCTCTT-1.png"\} \\
\textbf{Observation:} Homo sapiens prostate shows cancer with oncotree code unknown key overexpressed genes are klk3, rpl4, klk2, msmb, acpp.

\textbf{ImageAgent Final Answer} \\
\textbf{Final Answer:} The histopathology image suggests a diagnosis of prostate cancer, characterized by specific gene expressions (KLK3, RPL4, KLK2, MSMB, ACPP) and histomorphological features consistent with this subtype. No further ancillary studies are immediately required based on this analysis.

\textbf{Step 5: Planner Agent Invokes GeneAgent} \\
\textbf{Thought:} The ImageAgent has provided a detailed analysis of the pathology image, identifying it as prostate cancer based on histomorphological features and gene expression patterns (KLK3, RPL4, KLK2, MSMB, ACPP). The analysis also indicates that no further ancillary studies are immediately required. However, the user's original query was broader, asking about the cautious approach to tumor subtyping in the absence of an OncoTree classification and what ancillary studies might be warranted. The ImageAgent's response addresses the specific image but does not directly answer the general clinical question about tumor subtyping protocols and ancillary studies. \\
\textbf{Action:} GeneAgent \\
\textbf{Action Input:} "In the absence of an OncoTree classification, how should tumor subtyping be approached cautiously based solely on histomorphology, and what ancillary studies are warranted to avoid misclassification?"

\textbf{Step 6: GeneAgent Tool Call 1} \\
\textbf{Thought:} The question is about tumor subtyping based on histomorphology and ancillary studies needed to avoid misclassification. The image provided is a histopathological slide, which suggests a need to extract biological entities from the image description or associated text. \\
\textbf{Action:} BiomedicalEntityExtractorTool \\
\textbf{Action Input:} \{"text": "histomorphology tumor subtyping"\} \\
\textbf{Observation:} No biological entities found.

\textbf{Step 7: GeneAgent Tool Call 2} \\
\textbf{Thought:} Since no biological entities were extracted from the input text, and the question is about tumor subtyping based on histomorphology and ancillary studies, I need to rely on general knowledge and available tools to provide a cautious approach. \\
\textbf{Action:} DocumentGeneQueryTool \\
\textbf{Action Input:} \{"query": "tumor subtyping histomorphology"\} \\
\textbf{Observation:} No results found.

\textbf{Step 8: GeneAgent Tool Call 3} \\
\textbf{Thought:} Since no results were found from the DocumentGeneQueryTool, I will try to use the ProteinAtlasGeneInfoTool to fetch general gene information that might be relevant to tumor subtyping based on histomorphology. \\
\textbf{Action:} ProteinAtlasGeneInfoTool \\
\textbf{Action Input:} \{"gene": "TUMOR"\} \\
\textbf{Observation:} \{"error": "Failed to fetch ProteinAtlas page for TUMOR"\}

\textbf{GeneAgent Final Answer} \\
\textbf{Final Answer:} In the absence of an OncoTree classification, tumor subtyping should be approached cautiously by relying on detailed histomorphological features such as architectural patterns, cytological characteristics, and mitotic activity. Ancillary studies warranted include immunohistochemistry (IHC) for specific markers, molecular profiling (e.g., gene expression, mutations), and, if feasible, next-generation sequencing to identify actionable targets. These studies help avoid misclassification by providing additional layers of biological information that complement histomorphology.

\textbf{Final Answer:} In the absence of an OncoTree classification, tumor subtyping should be approached cautiously by relying on detailed histomorphological features such as architectural patterns, cytological characteristics, and mitotic activity. Ancillary studies warranted include immunohistochemistry (IHC) for specific markers, molecular profiling (e.g., gene expression, mutations), and, if feasible, next-generation sequencing to identify actionable targets. These studies help avoid misclassification by providing additional layers of biological information that complement histomorphology. In the case of the provided image, histopathological analysis suggests a diagnosis of prostate cancer, characterized by specific gene expressions (KLK3, RPL4, KLK2, MSMB, ACPP), and no further ancillary studies are immediately required based on this analysis.
\end{tcolorbox}
}
\caption{Trajectory execution for LAMMI (Qwen-VL 8B Instruct) framework on case 1317\allowbreak\_5021\allowbreak\_7882\allowbreak\_9140\allowbreak\_9288\allowbreak\_9676\allowbreak\_9691\allowbreak\_10334.}
\label{fig:case_study_4_lammi}
\end{figure*}

\begin{figure*}[!htbp]
\begin{tcolorbox}[
  colback=gray!5,
  colframe=black!30,
]
\textbf{MAT-Agent (Qwen-VL 8B Instruct):} \\
\textbf{Evaluation Result:} TRR: 0.0, TCF1: 0.0, TSS: 0.5, ACS: 0.7, HR: 1.0 \\

\textbf{Step 1:} \\
\textbf{Thought:} I need to analyze the image and find the relevant information to answer the question. I will use the `blip\_pathology\_analysis` tool to extract information about the image and its associated questions. \\
\textbf{Code:} \\
'''py \\
answer = blip\_pathology\_analysis(text="How cautiously should we approach tumor subtyping based solely on histomorphology, and what ancillary studies might be warranted to avoid misclassification?", image\_path="MEND160\_ATTCGTGCTATCTCTT-1.png") \\
print(answer) \\
''' \\
\textbf{Observation:} Homo sapiens prostate shows cancer with oncotree code unknown key overexpressed genes are rps12, rpl13a, rpl35, pnrc1s2.

\textbf{Step 2:} \\
\textbf{Thought:} Based on the analysis, the prostate tissue has cancer with unknown OncoTree code. The overexpressed genes include rps12, rpl13a, rpl35, and pnrc1s2. Ancillary studies may be warranted to avoid misclassification. \\
\textbf{Code:} \\
'''py \\
final\_answer("Ancillary studies may be warranted to avoid misclassification due to the unknown OncoTree code and the presence of overexpressed genes such as rps12, rpl13a, rpl35, and pnrc1s2.") \\
''' \\
\textbf{Final Answer:} Ancillary studies may be warranted to avoid misclassification due to the unknown OncoTree code and the presence of overexpressed genes such as rps12, rpl13a, rpl35, and pnrc1s2.
\end{tcolorbox}
\caption{Trajectory execution for MAT-Agent (Qwen-VL 8B Instruct) framework on case 1317\allowbreak\_5021\allowbreak\_7882\allowbreak\_9140\allowbreak\_9288\allowbreak\_9676\allowbreak\_9691\allowbreak\_10334.}
\label{fig:case_study_4_mat}
\end{figure*}

\begin{figure*}[!htbp]
\begin{tcolorbox}[
  colback=gray!5,
  colframe=black!30,
]
\textbf{ReACT (Qwen-VL 8B Instruct):} \\
\textbf{Evaluation Result:} TRR: 0.0, TCF1: 0.0, TSS: 1.0, ACS: 0.9, HR: 0.7 \\

\textbf{Step 1:} \\
\textbf{Thought:} To address the question, I need to understand the histomorphological features of the tumor from the image and determine if there are any specific genes or pathways that might be associated with the observed morphology. This will help in evaluating the reliability of histomorphology alone for tumor subtyping and identifying necessary ancillary studies. I should start by using a tool to analyze the image and extract relevant genetic or pathological information. \\
\textbf{Action:} BLIPTool \\
\textbf{Action Input:} \{"text": "MEND160\_ATTCGTGCTATCTCTT-1.png", "image\_path": "MEND160\_ATTCGTGCT\allowbreak ATCTCTT-1.png"\} \\
\textbf{Observation:} Homo sapiens prostate shows cancer with oncotree code unknown key overexpressed genes are eef1a1, klk3, rpl4, rpl41, them5.

\textbf{Step 2:} \\
\textbf{Thought:} The BLIPTool has provided information indicating that the tissue is from a human prostate with cancer, and specific genes are overexpressed. However, since the OncoTree classification is unknown, I need to determine what ancillary studies might be warranted to avoid misclassification. I should use the OncoTreeTool to explore potential upstream/downstream nodes and tissue mapping to better understand the classification context. \\
\textbf{Action:} OncoTreeTool \\
\textbf{Action Input:} \{"query": "prostate cancer", "query\_type": "tissue"\} \\
\textbf{Observation:} No results found for query 'prostate cancer'

\textbf{Step 3:} \\
\textbf{Thought:} The OncoTreeTool did not return any results for "prostate cancer," which suggests that the tool might not have sufficient data or the query might need refinement. Given this, I should consider using another tool to extract relevant biological entities from the text or image description to better understand the context. \\
\textbf{Action:} BiomedicalEntityExtractorTool \\
\textbf{Action Input:} \{"text": "prostate cancer, eef1a1, klk3, rpl4, rpl41, them5"\} \\
\textbf{Observation:} Extracted biological entities including prostate cancer (MONDO:0008315), KLK3 (HGNC:6364, PSA, prostate specific antigen), eef1a1, Rpl4, Rpl41, and Them5 with their respective IDs and descriptions.

\textbf{Final Answer:} Tumor subtyping based solely on histomorphology should be approached with caution due to the absence of a defined OncoTree classification. Ancillary studies such as molecular profiling (e.g., gene expression analysis, genomic sequencing) and immunohistochemistry for specific markers (like KLK3/PSA) are warranted to avoid misclassification.
\end{tcolorbox}
\caption{Trajectory execution for ReACT (Qwen-VL 8B Instruct) framework on case 1317\allowbreak\_5021\allowbreak\_7882\allowbreak\_9140\allowbreak\_9288\allowbreak\_9676\allowbreak\_9691\allowbreak\_10334.}
\label{fig:case_study_4_react}
\end{figure*}

\begin{figure*}[!htbp]
\begin{tcolorbox}[
  colback=gray!5,
  colframe=black!30,
]
\textbf{MLLM-Tools (Qwen 8B):} \\
\textbf{Evaluation Result:} TRR: 0.75, TCF1: 0.0, TSS: 0.125, ACS: 0.1, HR: 1.0 \\

\textbf{Step 1:} \\
\textbf{Thought:} I need to determine the tissue type and relevant genes from the provided image to understand the underlying biology and potential gene expression patterns. \\
\textbf{Action:} BLIPTool \\
\textbf{Action Input:} \{"text": "What is the tissue type and relevant genes in this image?", "image\_path": "MEND160\_ATTCGTGCTATCTCTT-1.png"\} \\
\textbf{Observation:} Homo sapiens prostate shows cancer with oncotree code unknown key overexpressed genes are klk3, rpl8, rpl41, rpl10, eep.

\textbf{Step 2:} \\
\textbf{Thought:} I need To Determine the tissue type and relevant genes from the provided image to understand the underlying biology and potential gene expression patterns. \\
\textbf{Action:} BLIPTool \\
\textbf{Action Input:} \{"text": "What is the tissue type and relevant genes in this image?", "image\_path": "MEND160\_ATTCGTGCTATCTCTT-2.png"\} \\
\textbf{Observation:} API call failed: [Errno 2] No such file or directory

\textbf{Step 3:} \\
\textbf{Thought:} I need To Determine the tissue type and relevant genes from the provided image to understand the underlying biology and potential gene expression patterns. \\
\textbf{Action:} BLIPTool \\
\textbf{Action Input:} \{"text": "What is the tissue type and relevant genes in this image?", "image\_path": "MEND160\_ATTCGTGCTATCTCTT-2.png"\} \\
\textbf{Observation:} API call failed: [Errno 2] No such file or directory

\textbf{Step 4:} \\
\textbf{Thought:} I need To Determine the tissue type and relevant genes from the provided image to understand the underlying biology and potential gene expression patterns. \\
\textbf{Action:} BLIPTool \\
\textbf{Action Input:} \{"text": "What is the tissue type and relevant genes in this image?", "image\_path": "MEND160\_ATTCGTGCTATCTCTT-3.png"\} \\
\textbf{Observation:} API call failed: [Errno 2] No such file or directory

\textbf{Final Answer:} Agent stopped due to iteration limit or time limit.
\end{tcolorbox}
\caption{Trajectory execution for MLLM-Tools (Qwen 8B) framework on case 1317\allowbreak\_5021\allowbreak\_7882\allowbreak\_9140\allowbreak\_9288\allowbreak\_9676\allowbreak\_9691\allowbreak\_10334.}
\label{fig:case_study_4_mllm}
\end{figure*}

\begin{figure*}[!htbp]
\begin{tcolorbox}[
  colback=gray!5,
  colframe=black!30,
]
\textbf{OpenAI-Agent-SDK (GPT-5):} \\
\textbf{Evaluation Result:} TRR: 0.0, TCF1: 0.33, TSS: 1.0, ACS: 0.8, HR: 0.0 \\

\textbf{Step 1:} \\
\textbf{Action:} QwenVLCaptionTool \\
\textbf{Action Input:} \{"text": "Describe the key histologic features visible (cell type, architecture, stroma) and any clues to tumor lineage. Avoid overinterpretation; provide generic descriptors.", "image\_path": "MEND160\_ATTCGTGCTATCTCTT-1.png"\} \\
\textbf{Observation:} The image displays clusters of relatively small, darkly staining cells with high nuclear-to-cytoplasmic ratios and round to oval nuclei. These cells are embedded within a pink, fibrillar stromal matrix. The cohesive nature of the cell clusters and the lack of significant pleomorphism or prominent nucleoli suggest a lymphoid or hematopoietic lineage (e.g., lymphoma, leukemia) rather than a typical epithelial malignancy like adenocarcinoma, which usually forms glands or nests with more abundant cytoplasm. While small cell carcinoma is a differential, the overall pattern here is more suggestive of a hematopoietic neoplasm.

\textbf{Step 2:} \\
\textbf{Action:} BLIPTool \\
\textbf{Action Input:} \{"text": "Provide a concise pathology-oriented description of this H\&E image: what structures and cells are seen? Any features suggestive of epithelial lining, stroma, or inflammatory infiltrate?", "image\_path": "MEND160\_ATTCGTGCTATCTCTT-1.png"\} \\
\textbf{Observation:} Homo sapiens prostate tissue the oncotree code is unknown and highly expressed genes are rpl41, mt - co2, rps27, klk3, mt2.

\textbf{Final Answer:} Caution: Very high. On a small H\&E focus with tightly packed, basophilic "small round cells" along a stromal edge, morphology alone is non-specific and can mimic hematolymphoid, neuroendocrine/epithelial, melanoma, or small‑round‑cell sarcoma. Without an OncoTree framework, avoid assigning a precise subtype based only on histology; issue a provisional diagnosis (e.g., "malignant neoplasm, small round‑cell tumor—classification deferred") and base classification on immunophenotype and molecular testing.

Ancillary studies to prevent misclassification:
- Broad screening IHC panel
  - Epithelial/neuroendocrine: ...
  
  (\textit{Content exceeding six lines has been truncated for display.})

%   pancytokeratin (AE1/AE3, CAM5.2), EMA, p40/p63, TTF‑1, CK20; synaptophysin, chromogranin, INSM1; RB, Ki‑67.
%   - Hematolymphoid: CD45, CD3, CD20, PAX5, CD5, CD10, BCL2/BCL6/MUM1; EBER ISH; Ki‑67. If feasible, flow cytometry and IGH/IGK or TCR clonality.
%   - Melanocytic: S100, SOX10, Melan‑A, HMB45.
%   - Sarcoma/mesenchymal: desmin, myogenin/MyoD1 (rhabdomyosarcoma), SMA/caldesmon (smooth muscle), CD34/STAT6 (SFT), ERG/CD31 (vascular), FLI1/NKX2.2 (Ewing).
%   - Merkel cell: CK20 dot-like, MCPyV (LT antigen).
% - Molecular/FISH targeted by IHC/site
%   - EWSR1, SS18, ALK, MYC, NUTM1 rearrangements; BRAF V600E; IDH1/2; RB1/TP53 status in small cell; MSI/MMR IHC (MLH1, MSH2, MSH6, PMS2).
%   - Consider comprehensive NGS if initial studies are non-diagnostic.
% - Additional steps
%   - Obtain deeper levels; correlate with site/clinical/radiology.
%   - Special stains as needed (e.g., PAS‑D for mucin).
%   - If fresh tissue is available, prioritize flow cytometry for suspected hematolymphoid processes.

% Reporting recommendation: Use tiered language emphasizing uncertainty and the need for ancillary studies rather than committing to a subtype based on histomorphology alone.
\end{tcolorbox}
\caption{Trajectory execution for OpenAI-Agent-SDK (GPT-5) framework on case 1317\allowbreak\_5021\allowbreak\_7882\allowbreak\_9140\allowbreak\_9288\allowbreak\_9676\allowbreak\_9691\allowbreak\_10334.}
\label{fig:case_study_4_openai}
\end{figure*}

This case addresses the critical diagnostic challenge of tumor subtyping when OncoTree classification is unavailable, requiring evaluation of histomorphology-based classification limitations and identification of appropriate ancillary studies to prevent misclassification due to morphologic overlap among molecularly distinct entities. As discussed in the main text, LAMMI demonstrates a sophisticated error recovery process that begins with ImageAgent generating an initial hallucination of well-differentiated squamous cell carcinoma through QwenVLCaptionTool, followed by evidence mismatch when BLIPTool returns prostate cancer-specific gene expression signatures (KLK3, RPL4, KLK2, MSMB, ACPP). Rather than accepting the conflicting evidence, the Planner Agent recognizes the discrepancy and invokes GeneAgent to synthesize a comprehensive approach. GeneAgent, despite encountering multiple tool failures, produces a principled answer emphasizing the need for caution in histomorphology-based subtyping and recommending ancillary studies including immunohistochemistry, molecular profiling, and next-generation sequencing for genes such as MEN1, PTEN, PIK3CA, KIT, and PDGFRA. This hierarchical coordination enables LAMMI to overcome initial misclassification by integrating cross-modal evidence and applying domain knowledge about diagnostic uncertainty. MAT-Agent performs a single BLIPTool invocation that identifies prostate cancer markers but fails to address the broader question about histomorphology limitations and ancillary study requirements, producing an incomplete response that overlooks the diagnostic complexity. ReACT successfully identifies prostate cancer through BLIPTool and extracts biological entities via BiomedicalEntityExtractorTool, then provides a reasonable answer about caution and ancillary studies, though it lacks the systematic evidence reconciliation that LAMMI demonstrates. MLLM-Tools exhibits severe tool redundancy by repeatedly attempting to access non-existent image files with incremental suffixes, resulting in execution failure without producing any meaningful diagnostic guidance. OpenAI-Agent-SDK provides the most comprehensive answer, correctly identifying the morphologic ambiguity as a small round-cell tumor requiring extensive ancillary studies, but this analysis derives primarily from pretrained knowledge rather than systematic tool-based evidence synthesis, as the BLIPTool returns prostate cancer data that conflicts with the vision-language model's hematopoietic lineage interpretation.

\subsection{Summary of Comparative Findings}
The case studies demonstrate that LAMMI's trajectories exhibit more systematic scientific exploration through proper tool invocation and cross-modal evidence synthesis, whereas other frameworks frequently resort to simulated tool execution or rely on pretrained knowledge that diverges from tool-derived evidence. LAMMI's hierarchical coordination enables comprehensive tool-based investigation and principled evidence reconciliation.

% \clearpage
% \phantomsection

% {
%     \small
%     \bibliographystyle{ieeenat_fullname}
%     \bibliography{main}
% }
\end{document}